\documentclass[11pt, a4paper, logo, onecolumn, nonumbering]{paperstyle}

\pdftrailerid{redacted}


\usepackage{kantlipsum, lipsum}
\usepackage{dsfont}
\usepackage{multirow}
\usepackage{array}
\usepackage{amssymb} %
\usepackage{pifont} %
\usepackage{makecell}
\usepackage{algorithm}
\usepackage{algpseudocode}
\usepackage{amsmath}
\usepackage{multicol}
\usepackage{float}
\usepackage{placeins}
\usepackage{siunitx}
\usepackage{tabularx}
\usepackage{booktabs}
\usepackage{listings}

\usepackage{xcolor}
\definecolor{GoogleBlue}{HTML}{1A73E8}
\definecolor{GoogleBlueLight}{HTML}{e8f0fe}
\definecolor{GoogleBlueDark}{HTML}{174ea6}
\definecolor{GoogleRed}{HTML}{D93025}
\definecolor{GoogleGray}{HTML}{767676}
\definecolor{figtextcolor}{HTML}{333333}

\lstdefinestyle{qcprompt}{
    basicstyle=\ttfamily\fontsize{8pt}{9.4pt}\selectfont,
    breaklines=true,
    breakatwhitespace=false,
    columns=fullflexible,
    keepspaces=true,
    showstringspaces=false,
    upquote=true,
    frame=single,
    framerule=0.4pt,
    rulecolor=\color{GoogleGray!55},
    backgroundcolor=\color{GoogleBlueLight!28!white},
    framexleftmargin=4pt,
    framexrightmargin=4pt,
    framesep=4pt,
    aboveskip=4pt,
    belowskip=7pt
}
\usepackage{CJKutf8}
\makeatletter
\let\catalogAddCmap\CJK@addcmap
\def\CJK@addcmap#1{\catalogAddCmap{#1}\expandafter\pdfnobuiltintounicode\font@name}
\makeatother
\newsavebox{\catalogMeituan}
\newsavebox{\catalogHonorOfKings}
\AtBeginDocument{%
    \sbox{\catalogMeituan}{\fontsize{8pt}{9.4pt}\selectfont\begin{CJK*}{UTF8}{gbsn}美团\end{CJK*}}%
    \sbox{\catalogHonorOfKings}{\fontsize{8pt}{9.4pt}\selectfont\begin{CJK*}{UTF8}{gbsn}王者荣耀\end{CJK*}}%
}
\lstdefinestyle{catalogprompt}{
    style=qcprompt,
    basicstyle=\ttfamily\fontsize{8pt}{9.2pt}\selectfont,
    aboveskip=3pt,
    belowskip=6pt,
    literate={—}{{---}}3 {…}{{...}}3 {→}{{$\rightarrow$}}2
      {✅}{{\ensuremath{\checkmark}}}1 {❌}{{\ding{55}}}1
      {美团}{{\usebox{\catalogMeituan}}}4
      {王者荣耀}{{\usebox{\catalogHonorOfKings}}}8
}
\definecolor{perceptioncol}{HTML}{1A73E8}
\definecolor{modelingcol}{HTML}{925BCF}
\definecolor{manipulationcol}{HTML}{C440A7}
\definecolor{reasoningcol}{HTML}{DA2F77}

\usepackage{subcaption}
\usepackage{opensans}
\DeclareCaptionFont{subfigcapsize}{\fontsize{8pt}{8pt}\selectfont}  %
\usepackage{xurl}
\usepackage[colorlinks=true, allcolors=GoogleBlue]{hyperref}
\usepackage[capitalise]{cleveref}
\crefname{section}{Sec.}{Secs.}  %
\crefname{subsection}{Sec.}{Secs.}  %
\crefname{subsubsection}{Sec.}{Secs.}  %
\crefname{appendix}{App.}{Apps.}  %

\usepackage{graphicx}
\graphicspath{{figures/}}

\usepackage[breakable]{tcolorbox}
\renewcommand{\abscontent}{%
    \begin{tcolorbox}[colframe=HunyuanBlueDark, colback=HunyuanBlueDark!3!white,
        coltext=black, width=\linewidth, boxrule=0.5pt, arc=3mm,
        left=4mm, right=4mm, top=3mm, bottom=3mm,
        before skip=0pt, after skip=5mm]
        {\centering\sffamily\bfseries\large\color{HunyuanBlueDark}Abstract\par}
        \vspace{2mm}
        {\absfont\mdseries\theabstract\par}
        \vspace{2.5mm}
        {\centering\normalfont\small
        \renewcommand{\arraystretch}{1.18}
        \begin{tabular}{@{}c@{\hspace{0.65em}}l@{\hspace{0.9em}}l@{}}
          \raisebox{-0.18em}{\includegraphics[height=1.12em]{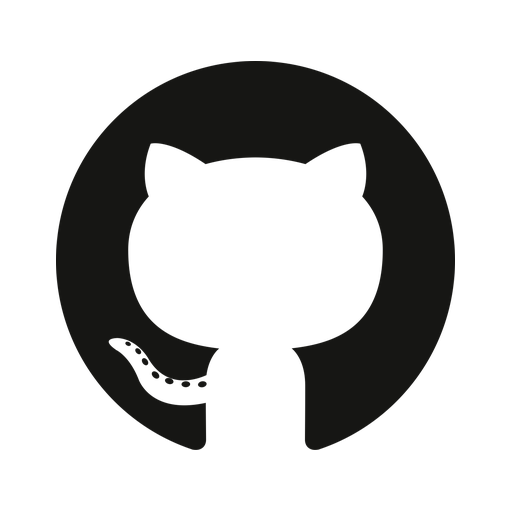}}
          & \textbf{GitHub:}
          & \href{https://github.com/ImYangC7/AutoGUIWorld}{\urlstyle{same}\url{https://github.com/ImYangC7/AutoGUIWorld}} \\
          \raisebox{-0.18em}{\includegraphics[height=1.12em]{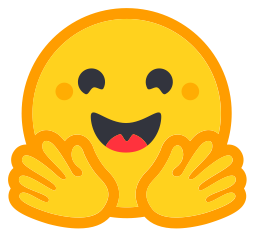}}
          & \textbf{Hugging Face:}
          & \href{https://huggingface.co/YangC777/AGW-35B}{\urlstyle{same}\url{https://huggingface.co/YangC777/AGW-35B}}
        \end{tabular}\par}
    \end{tcolorbox}%
    \begin{minipage}{\linewidth}
        \centering
        \includegraphics[width=\linewidth]{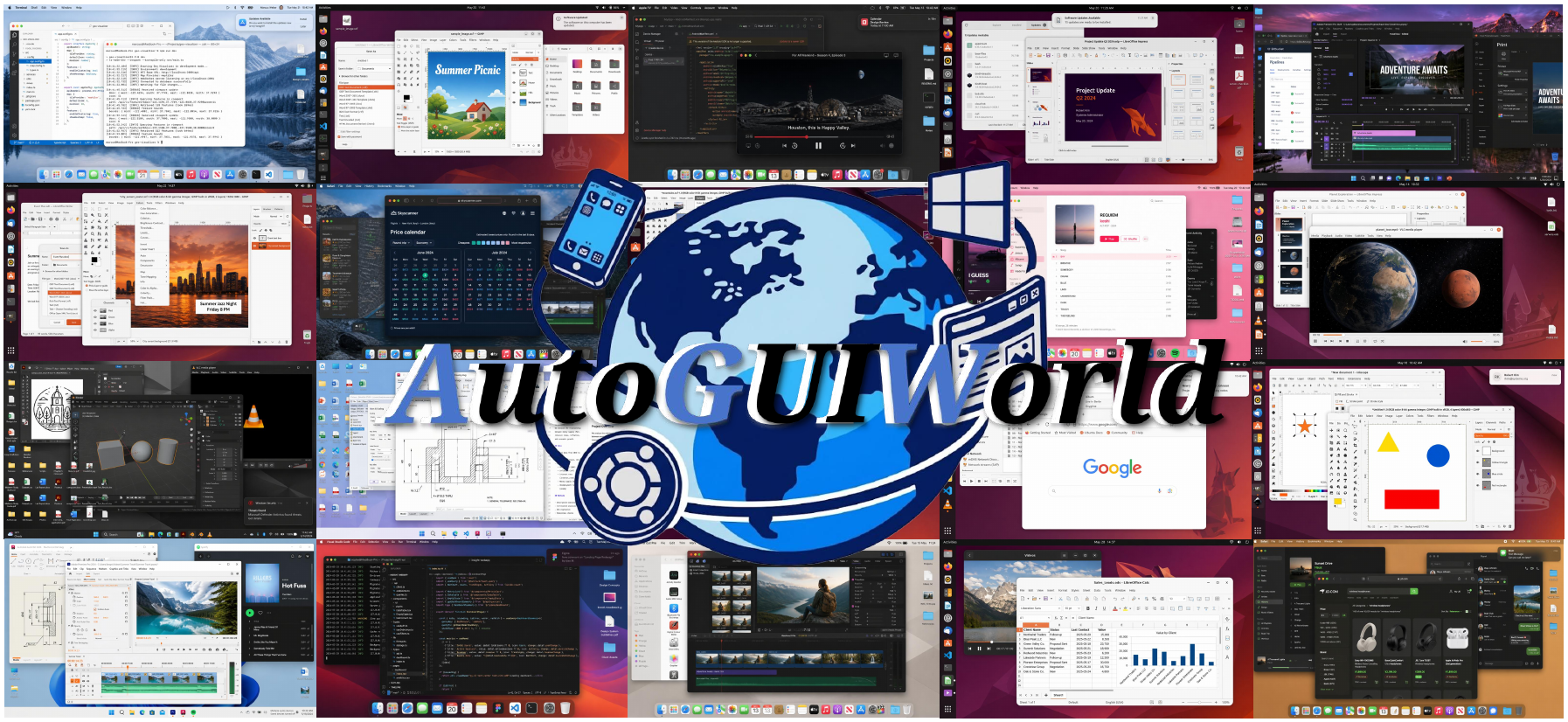}
        \captionsetup{hypcap=false}
        \captionof{figure}{Overview of AutoGUIWorld as an environment-free GUI data engine. It
        samples a structured GUI world, realizes a seed screenshot, plans
        action-conditioned transitions, and uses Image2 as a visual world model to
        synthesize coherent screenshot-action-screenshot trajectories.}
    \end{minipage}%
}
\newcommand{\callout}[1]{%
    \begin{center}
    \begin{tcolorbox}[colframe=GoogleBlue, colback=GoogleBlueLight, coltext=GoogleBlue!30!black, width=\columnwidth, left=2mm, right=2mm, top=1mm, bottom=1mm, boxrule=.5mm]
        #1
    \end{tcolorbox}
    \end{center}
}
\newcounter{takeaway}
\newcommand{\takeaway}[1]{%
    \stepcounter{takeaway}
    \callout{\setlength{\parfillskip}{0pt}%
        \textbf{Takeaway\;\thetakeaway}\hspace{1em}#1\par}
}

\usepackage{tikz}
\usetikzlibrary{calc}
\newlength{\tikzwidth}
\newlength{\boxpad}
\newlength{\boxgap}
\newlength{\boxtextheight}
\title{\textcolor{HunyuanBlueDark}{A}uto\textcolor{HunyuanBlueDark}{G}UI\textcolor{HunyuanBlueDark}{W}orld:\\[-0.15em]
{\fontsize{17}{19.2}\selectfont Image Generators as Visual World Models for GUI Agent}}
\usepackage[numbers, sort&compress]{natbib}
\correspondingauthor{}

\reportnumber{} %

\author{\makebox[\linewidth][c]{\large\bfseries Hunyuan~AI~Data~Team}}

\begin{abstract}
GUI agents require high-quality interaction trajectories to learn how software
environments respond to actions, maintain state, and support multi-step
workflows. However, the diversity of available trajectories is constrained by
the applications, interface states, and workflows accessible in the underlying
environments. Expanding this coverage requires deploying increasingly diverse
and complex software, with specialized applications imposing additional
installation, configuration, and runtime costs. We introduce
\textbf{AutoGUIWorld}, a data generation framework that combines the visual
priors of image generators with the task knowledge of a planner to synthesize
GUI interaction trajectories without deploying or running the corresponding
software environments. AutoGUIWorld samples initial GUI scenes from structured
specifications of operating-system context, visual appearance, and interface
state, and generates tasks conditioned on those scenes. A planner
then specifies atomic actions and their intended visual consequences, while an
image generator iteratively edits the current screenshot to produce subsequent
observations. Action grounding and transition-level quality filtering yield
79,266 spatially annotated step-level training samples across Ubuntu, Windows,
macOS, and Chrome.
Fine-tuning Qwen3.5-35B-A3B on AutoGUIWorld trajectories improves
the mean task score on OSWorld from 33.0\% to 40.8\% and the task success rate
on ScienceBoard from 14.0\% to 32.2\%. These results show that generated trajectories improve
GUI-agent performance on real desktop and scientific tasks.
\end{abstract}

\begin{document}

\newpage
\maketitle
\clearpage
\tableofcontents
\clearpage

\section{Introduction}
\label{sec:introduction}

GUI agents automate software tasks through screenshots, mouse actions, and
keyboard inputs. Large-scale interaction data improves GUI perception, grounding,
and task execution \citep{qin2025uitars,wang2025opencua,liu2025scalecua}.
Effective interaction also requires environment knowledge: how actions change
observable states, what persists, and how earlier operations constrain later
ones \citep{lu2024guiodyssey,chae2024wma}. An agent must relate the current
screenshot to previous actions and the remaining task, distinguishing content
that should persist from changes needed to make progress. High-quality
trajectories provide supervision for learning these dependencies by connecting
instructions and actions to observable outcomes. Their successive observations
show how intermediate decisions shape later states across multi-step workflows.

Trajectory diversity depends on the breadth and complexity of the environments
available for collection: their reachable interface states, action semantics,
and workflow structures determine which interactions can be observed.
Scientific, CAD, and electronic-design software expose specialized states and
operations \citep{sun2025scienceboard,dong2026cadworld,li2025guieda}, while creative
and professional workflows require content preservation and dependent editing
steps \citep{wang2026deskcraft,pang2026visculpt,ai2025prosoftarena,zhu2026workflowgym}.
Collecting more trajectories within an existing environment can increase task
coverage, but cannot supply interactions specific to software absent from the
collection. Expanding the environment pool requires deploying and configuring
software, preparing task states, and maintaining reproducible execution.
For specialized software, this entails accommodating different dependencies,
runtime requirements, and initialization procedures, alongside possible
licensing or account constraints
\citep{sun2025scienceboard,dong2026cadworld,aggarwal2026gymanything}.
These costs motivate generating training experience without deploying or running
each corresponding software environment.

Existing acquisition methods use human demonstrations
\citep{deng2023mind2web,rawles2023androidwild}, automated exploration and filtering
\citep{sun2024osgenesis,xiao2025uigenie}, or extraction from tutorials and recorded
videos \citep{zhang2025tongui,lu2025videoagenttrek}. Their coverage remains tied
to accessible executable environments or the interfaces and workflows present
in existing records. Generation offers a complementary way to expand the
available interaction experience.

Generative approaches construct executable environments with tasks and verifiers
\citep{zhang2026infiniteweb,wu2026autowebworld,wang2026cuagym}, or simulate
observations through structured states
\citep{wang2025uisimulator,xiao2026webworld,ding2026dynaweb}, visual prediction
\citep{luo2025vimo,rivard2025neuralos,xiang2025uisim,guan2026cuwm}, and renderable
code \citep{zheng2026code2world,koh2026gworld}. Pretrained image generators offer
another source of visual priors \citep{gabeur2026imagegenerators}, but turning
these priors into training trajectories requires observations that follow
intended actions and preserve context across steps \citep{fu2026guicc}.
Visual plausibility alone does not establish these properties: opening a dialog
should preserve the document behind it, and editing one field should leave
unrelated content intact. Generated screenshots also need corresponding action
coordinates to provide spatial supervision. These requirements motivate a
process that connects task-level intent to individual visual changes, grounds
actions in screenshots, and checks the resulting transitions. We investigate
whether task planning and transition-level quality control can make pretrained
image generators a practical source of training data for GUI agents operating
in real environments.

We introduce \textbf{AutoGUIWorld}, a framework that combines structured
environment sampling, scene-conditioned tasks, and visual trajectory generation.
It constructs training experiences without deploying or running each sampled
environment. A Meta Planner specifies the overall interaction plan; Voyager uses that plan and the
current screenshot to describe the next scene, which Image2 generates by editing
the screenshot. Grounding supplies action coordinates, and quality control
repairs annotations and removes defective samples. Our contributions are:

\begingroup
\setlength{\leftmargini}{1em}
\begin{itemize}
  \item \textbf{GUI trajectory generation.}
  We combine task planning and image generation to synthesize trajectories
  without deploying or running the corresponding software environments.

  \item \textbf{Curated training data.}
  We construct 79,266 grounded and filtered step-level samples across Ubuntu,
  Windows, macOS, and Chrome, and analyze their coverage and transition defects.

  \item \textbf{Transfer to real environments.}
  Fine-tuning Qwen3.5-35B-A3B improves all four interactive benchmarks, including
  OSWorld from \textbf{33.0\% to 40.8\%} mean task score and
  ScienceBoard from \textbf{14.0\% to 32.2\%} task success.
\end{itemize}
\endgroup

\section{Preliminaries}

\paragraph{GUI trajectory.}
We cast GUI interaction as a partially observable Markov decision
process (POMDP) $\mathcal{M} = (\mathcal{S}, \mathcal{A}, T,
\tilde{\mathcal{S}}, \Omega)$ together with a task instruction $\tau$.
The latent state $s_t \in \mathcal{S}$ is the full computer
configuration at step $t$, including operating-system state,
application internals, file-system contents, and any hidden controller
state, and the transition $T(s_{t+1} \mid s_t, a_t)$ is governed by the
host system. The agent never reads $s_t$ directly; it only observes the
rendered screenshot $\tilde{s}_t = \Omega(s_t) \in \tilde{\mathcal{S}}$
and emits an atomic action $a_t \in \mathcal{A}$ from the OS-dependent
action space listed in the appendix. Writing $h_t = (\tilde{s}_{\le t},
a_{<t})$ for the observation--action history, a GUI policy takes the
form
\begin{equation}
    a_t \sim \pi_\theta(\cdot \mid h_t, \tau).
\end{equation}
We condition on the full history because single screenshots are
generally non-Markov: scroll positions, dialog stacks, in-progress text
edits, and dynamic page content all carry information that is not
visible in $\tilde{s}_t$ alone.

\paragraph{Interaction fidelity.}
A screenshot-based agent never consumes $s_t$ and never invokes $T$,
so its training data is fully determined by the observation-level
transition $P(\tilde{s}_{t+1} \mid h_t, a_t)$ induced by marginalising
the latent dynamics. We refer to faithfulness with respect to $P$ as
\emph{interaction fidelity}: a synthetic trajectory has high
interaction fidelity if its post-action screenshot $\tilde{s}_{t+1}$
matches the distribution a real system would produce under the same
history, even when the underlying $s_{t+1}$ is never reconstructed.
AutoGUIWorld is organised around this relaxation: we model $P$ directly
with a visual world model and never instantiate $\mathcal{S}$ or $T$.

\paragraph{Data-generation factorization.}
The joint distribution of a length-$T$ GUI trajectory factorizes as
\begin{equation}
    p(\tau, \tilde{s}_{0:T}, a_{0:T-1})
    \;=\;
    p(\tau, \tilde{s}_0)
    \prod_{t=0}^{T-1}
    \pi(a_t \mid h_t, \tau)\,
    P(\tilde{s}_{t+1} \mid h_t, a_t).
\end{equation}
For data synthesis, AutoGUIWorld samples the task and initial visual state
from $p_\xi(\tau, \tilde{s}_0)$. The meta planner $\Pi_\psi$ uses the task
and fixed seed context to generate an action sequence and intended visual
transition descriptions $\delta_t \in \mathcal{D}$. During rollout, Voyager uses the
current screenshot, planned action, and rollout context to expand
$\delta_t$ into a rendering prompt $r_t$. Image2 generates the next
screenshot conditioned on $\tilde{s}_t$ and $r_t$.
\Cref{sec:method} details this construction.

\section{Method}
\label{sec:method}

\subsection{Overview and formulation}
We formulate AutoGUIWorld as a planner-guided visual world-model data engine for synthesizing GUI-agent training trajectories without executing actions in a real computer environment. AutoGUIWorld first samples and realizes an initial GUI seed, then generates a task instruction $\tau$ conditioned on that seed. After the seed-conditioned task generator produces $\tau$, the rollout process constructs a sequence of rendered GUI visual states $\tilde{s}_0,\ldots,\tilde{s}_T$. Here $\tilde{s}_t$ denotes a screen-level visual state proxy: it captures the visible window layout, page content, foreground application, interface controls, visual style, and interaction context at step $t$, but it is not equivalent to the full underlying system state such as file-system contents, application internals, or browser DOM state.

The generation process is decomposed into two layers. The \emph{planning layer} uses a meta planner $\Pi_{\psi}$ to generate an ordered sequence of atomic GUI actions and intended visual changes before rollout:
\begin{equation}
    (a_{0:T-1}, \delta_{0:T-1}) \sim \Pi_{\psi}(\tau, c_0),
\end{equation}
where $c_0$ is the fixed seed context, including the platform, visual style, and initial GUI description. Each $a_t \in \mathcal{A}$ is an action such as clicking, typing, scrolling, or dragging, and $\delta_t \in \mathcal{D}$ describes its expected visual consequence. The planner establishes the task logic, action order, and dependencies across steps.

The \emph{visual world-model layer} is instantiated by Image2, denoted as $M_{\phi}^{V}$. At each step, Voyager uses the current screenshot, planned action, and rollout context to expand $\delta_t$ into a rendering prompt $r_t$. Image2 generates the next visual state from the current screenshot and this prompt:
\begin{equation}
    \tilde{s}_{t+1} \sim M_{\phi}^{V}(\tilde{s}_t, r_t).
\end{equation}
Image2 functions as an action-conditioned visual state transition model. It preserves layout, style, background context, and user-visible content while realizing the changes specified by $\delta_t$ through $r_t$.

The resulting synthetic trajectory is therefore
\begin{equation}
    \tilde{s}_0
    \xrightarrow{(a_0,\delta_0)}
    \tilde{s}_1
    \xrightarrow{(a_1,\delta_1)}
    \cdots
    \xrightarrow{(a_{T-1},\delta_{T-1})}
    \tilde{s}_T .
\end{equation}
This formulation separates semantic planning from visual state transition: the meta planner specifies what should happen next, while the Image2 visual world model determines how the next GUI state should appear. AutoGUIWorld produces temporally coherent, action-grounded GUI experience for training agents.

\takeaway{Pretrained image generators can serve as GUI visual world models:
conditioned on the current screenshot and planned action effects, they predict
the next visual state.}

\begin{figure}[t]
  \centering
  \includegraphics[width=1\linewidth]{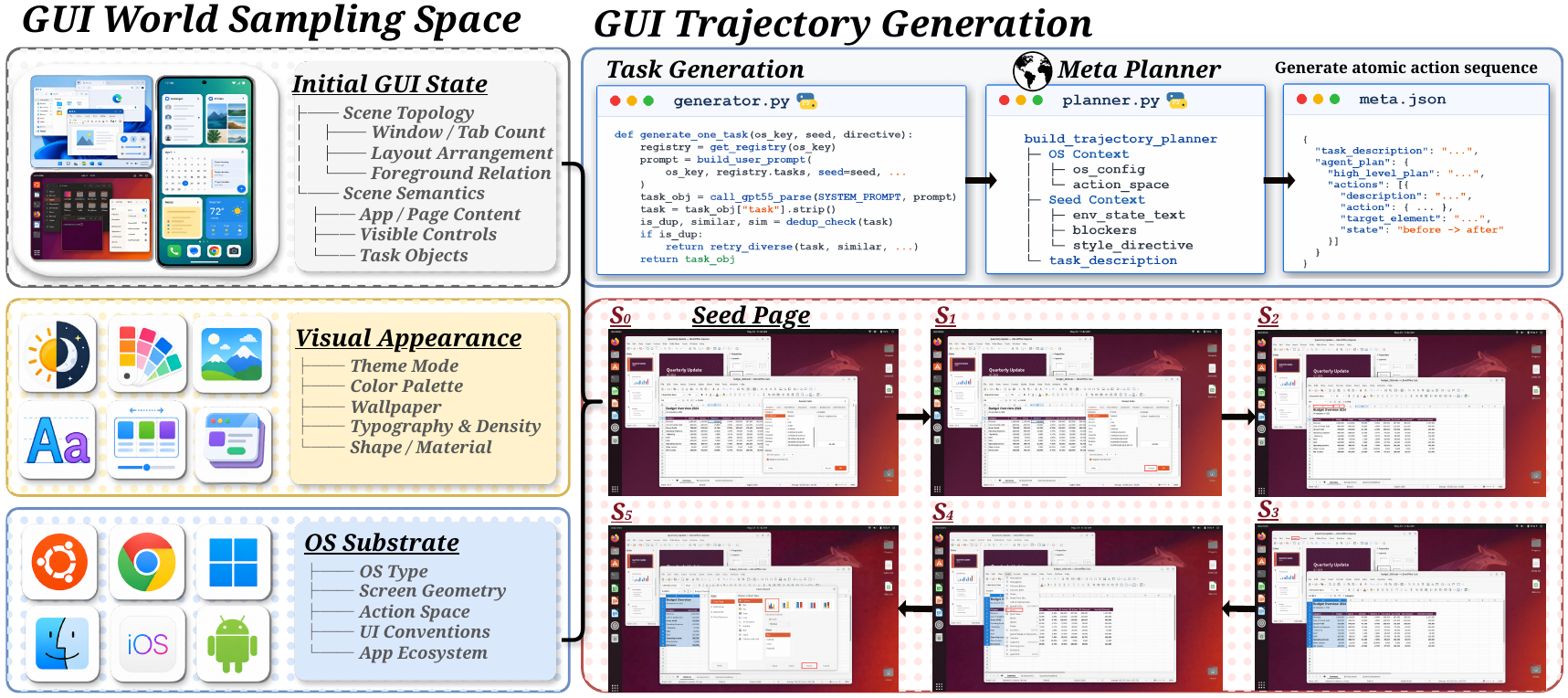}
  \caption{Overview of AutoGUIWorld. The method first samples a structured GUI world from an OS substrate, visual appearance space, and initial GUI state space. It then realizes a seed screenshot, generates seed-conditioned tasks, plans atomic actions, and rolls out a clean sequence of GUI visual states using Image2 as a visual world model.}
  \label{fig:method_overview}
\end{figure}

\begin{figure}[t]
  \centering
  \includegraphics[width=1\linewidth]{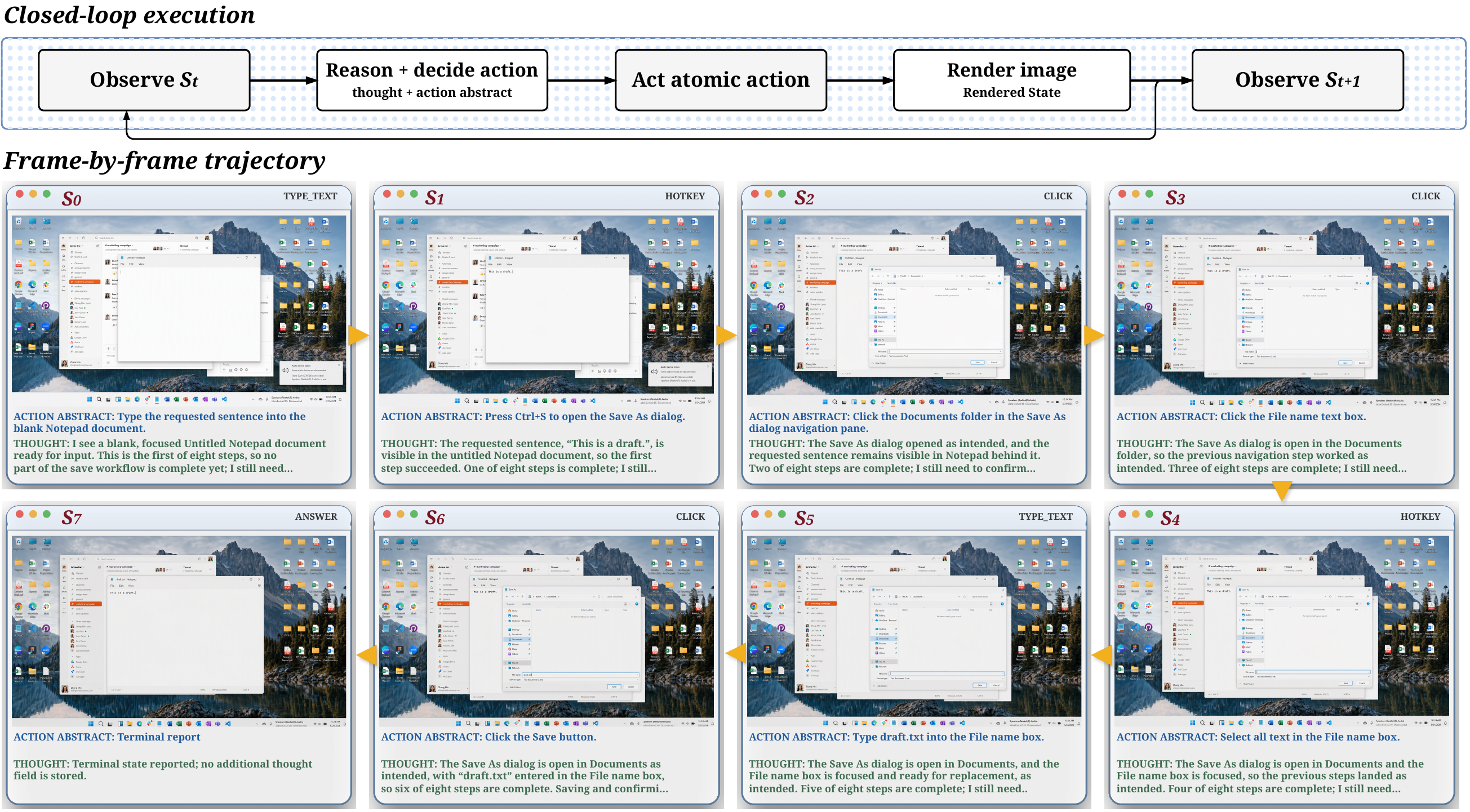}
  \caption{Voyager-guided closed-loop rendering. At each step, Voyager observes the current GUI image and receives the planned action. It produces a first-person thought, an action abstract, and an after-action rendering prompt. Image2 uses the current image as an image-to-image reference and renders the next GUI state, which is returned to Voyager for the next iteration. The bottom trajectory shows how one planned action sequence becomes a temporally consistent chain of rendered GUI states.}
  \label{fig:voyager_render}
\end{figure}

The closed-loop rollout is illustrated in Fig.~\ref{fig:voyager_render}. Voyager observes the current clean GUI state and receives the next action from the planned sequence. It generates a first-person thought, an action summary, and an after-action rendering prompt. Image2 applies this prompt to the current frame, producing the next visual state for the following Voyager step. Repeating this process converts the planned action sequence into a temporally linked screenshot trajectory.

\subsection{GUI World Sampling Space}

The initial state is sampled from a structured GUI world space rather than from an unconstrained text prompt. This space contains three complementary factors. The \emph{OS substrate} specifies platform-level constraints, including the operating system type, screen geometry, action space, interface conventions, and application ecosystem. The \emph{visual appearance} specifies the rendering style, including theme mode, color palette, wallpaper, typography, density, and material treatment. The \emph{initial GUI state} specifies the visible scene, including window or tab count, layout arrangement, foreground relation, application or page content, visible controls, and task-relevant objects.

This structured design serves two purposes. It provides controllable diversity across devices, platforms, applications, and visual styles. It also creates a persistent seed specification that can be reused by the task generator and the trajectory planner, ensuring that task instructions and planned actions are consistent with the generated initial screen.

\subsection{Seed Realization}

Given a sampled GUI world, AutoGUIWorld compiles the structured state into a detailed visual description. The description enumerates platform conventions, foreground and background surfaces, visible UI elements, layout relations, and appearance constraints. Image2 then renders the description into the initial screenshot $\tilde{s}_0$. The rendered image is stored together with its seed metadata, including the sampled state, visible elements, target surface, and visual style. Subsequent task generation and trajectory rollout condition on this same seed context.

The seed image provides the root of the trajectory. It is not treated as an isolated sample; instead, it becomes the reference state from which all subsequent visual transitions are generated.

\subsection{Seed-Conditioned Task Generation}

AutoGUIWorld generates tasks after the seed has been fixed. The task generator receives the platform context and the seed-visible elements, then proposes a concrete instruction that can be grounded in the current GUI world. This ordering is important: an instruction alone does not determine the screen on which it should be performed. By conditioning on the seed, the generator avoids tasks that refer to absent applications, hidden windows, or unsupported page contents.

The same interface supports both free-form task synthesis and benchmark-driven task adaptation. For synthetic tasks, a task registry discourages near-duplicate instructions and encourages coverage across applications and interaction types. For benchmark-derived tasks, AutoGUIWorld maps the benchmark metadata to the seed sampler, for example by pinning the required foreground application and adding related background windows when the task spans multiple applications.

\subsection{Planner-Guided Trajectory Rollout}

Before rollout, the meta planner generates the action sequence from the task and fixed seed context. Each step specifies an action $a_t$, its expected visual change $\delta_t$, and any target element $e_t$. The sequence respects dependencies and resolves preconditions first.

At step $t$, Voyager uses the current screenshot $\tilde{s}_t$, planned action $a_t$, and rollout context to expand $\delta_t$ into the rendering prompt $r_t$. The rollout context includes the overall plan, current step, and completed action summaries. Image2 edits $\tilde{s}_t$ according to $r_t$ to produce the next clean visual state $\tilde{s}_{t+1}$. Each new screenshot serves as the reference for the next step, carrying layout, background context, visual style, and user-visible content through the trajectory.

This division of labor is central to AutoGUIWorld. The planner determines what should happen next, while Image2 determines how the resulting GUI state should look. The output is a temporally coherent screenshot-action-screenshot trajectory rather than a collection of unrelated screenshots.

\subsection{Action Target Annotation}

Pointing actions require spatial supervision. For an action whose target is a visible element, AutoGUIWorld maps the semantic element description $e_t$ to a target point $p_t$ on the pre-action screenshot. LocateAnything~\citep{wang2026locateanything} locates the target region, whose center provides the point used as the action label.

AutoGUIWorld separates clean observations from action annotations. The clean state $\tilde{s}_t$ is used as the policy input and as the reference for the next visual transition. The annotated action frame marks the target region on a copy of $\tilde{s}_t$ for quality inspection and visualization; $p_t$ provides the spatial label for training. Non-pointing actions such as typing, scrolling, hotkeys, waiting, or answering do not require a target point.

\subsection{Training Instance Construction}

The generated trajectory can be converted into single-step training instances. Each instance uses the clean pre-action screenshot, the task instruction, and the prior interaction history as input. The target output is the planned action, together with a point when the action is spatially grounded. This conversion supports supervised fine-tuning, trajectory replay, and reinforcement-learning data construction without requiring the synthetic generator to be present at training time.

\takeaway{AutoGUIWorld frees GUI training-data generation from the need to
deploy and run each application. It expands coverage across software
environments, interface states, and workflows. Clean observations and grounded
actions provide supervision for GUI agents.}

\section{Data Feature}
\label{sec:data_feature}

AutoGUIWorld generates GUI scenes with broader interface coverage than ScaleCUA
and Qwen feature distances comparable to real cross-source variation.
We measure this combination of visual alignment and coverage expansion through
Qwen embeddings, low-level image statistics, and interface structure.
All open-source corpora are treated as real data because their screenshots were
collected from real GUI systems. The corpus pool contains ScaleCUA, AgentNet,
aria\_ui, WebSTAR, Mind2Web, and OS-Atlas. ScaleCUA serves as the matched real
reference because its native full-screen resolution is closest to the
corresponding synthetic domain.

\subsection{Training Set Overview}

\begin{figure}[!htbp]
  \centering
  \includegraphics[width=\linewidth]{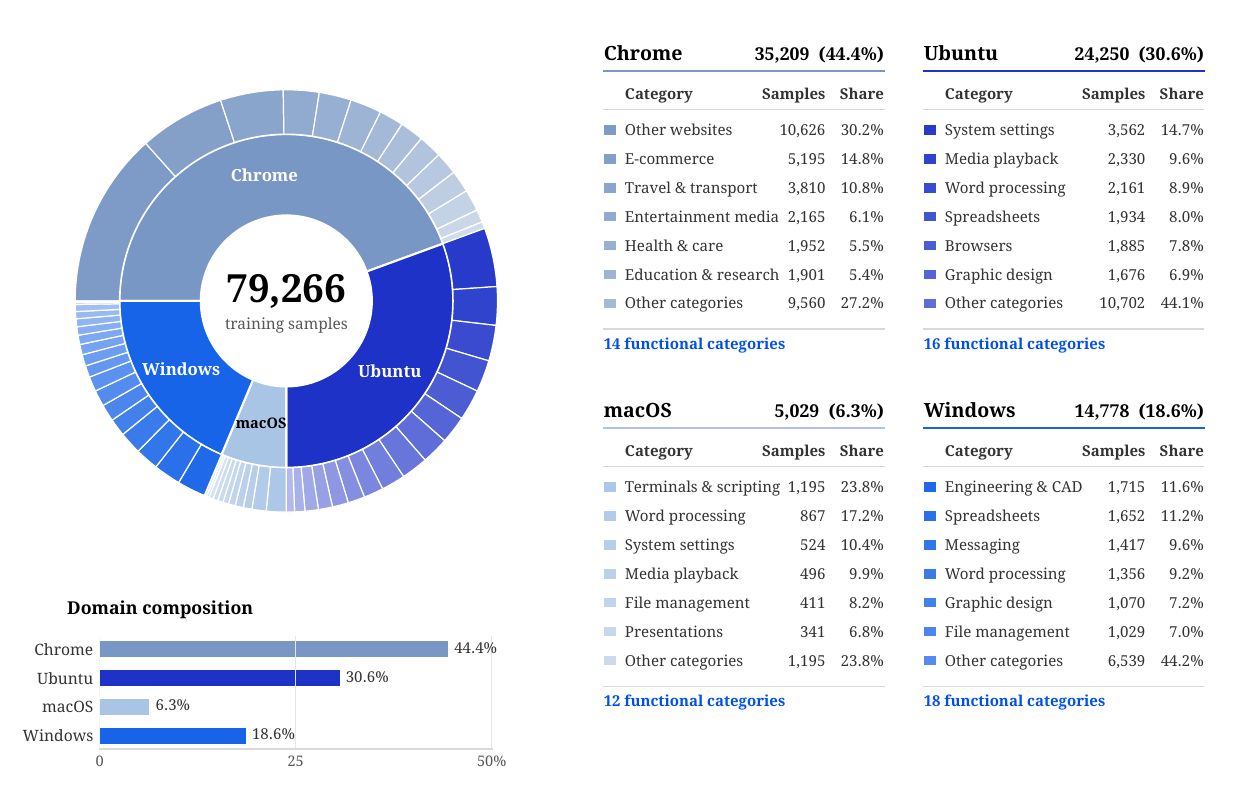}
  \caption{Distribution of the AutoGUIWorld training set. The inner ring shows
  79,266 training samples across Chrome, Ubuntu, Windows, and macOS. The outer
  ring and accompanying tables summarize the functional-category composition
  within each environment.}
  \label{fig:training_distribution}
\end{figure}

AutoGUIWorld contains 79,266 training samples across four GUI environments.
Chrome contributes 35,209 samples, followed by Ubuntu with 24,250, Windows
with 14,778, and macOS with 5,029. Figure~\ref{fig:training_distribution}
summarizes both the domain-level composition and the major functional
categories within each environment. The application- and website-level
coverage is detailed in Appendix~\ref{app:qualitative_trajectory_examples}.

\subsection{Experimental Setting}

The analysis compares gpt-image-2 synthetic screenshots with real data from the
same OS domain. Ubuntu pairs AutoGUI with ScaleCUA, aria\_ui, and AgentNet;
Windows uses ScaleCUA and AgentNet; Web pairs Image2GUI with ScaleCUA, WebSTAR,
and Mind2Web; macOS uses ScaleCUA, AgentNet, and OS-Atlas. Each metric uses the
sources for which that measurement is available. All train--test splits preserve
trajectory, session, or site groups. Adjacent frames from one interaction group
never cross a split.

\begin{table}[H]
  \centering
  \small
  \setlength{\tabcolsep}{5pt}
  \caption{Experimental configuration. Qwen features follow the visual path
  used to construct the model input. Auxiliary encoders and pixel statistics
  test whether the result depends on that representation.}
  \label{tab:data_feature_setting}
  \begin{tabularx}{\linewidth}{@{}l>{\raggedright\arraybackslash}X@{}}
    \toprule
    Component & Configuration \\
    \midrule
    Qwen representation & Qwen3.5-9B PatchMerger output; merge factor 2;
      4096-dimensional merged tokens; image-level mean pooling \\
    Image preprocessing & \texttt{smart\_resize} with
      \texttt{max\_pixels}$=2048\times32\times32$, matching the data pipeline \\
    MMD & Per-domain joint z-score; RBF kernel; one bandwidth selected from the
      median heuristic over the union of sources in that domain \\
    C2ST & GroupKFold logistic-regression probe; AUC 0.5 denotes chance-level
      discrimination \\
    Sampling & About 1,500 images per source for Qwen embeddings; 500 for
      DINOv2; 80 for OCR; 200--500 for element detection \\
    Pixel diagnostics & 300 images/source $\times$ 20 random $64^2$ patches;
      250 images/source $\times$ 12 artifact-probe $128^2$ patches \\
    Uncertainty & Group bootstrap with 400--1,000 resamples; joint resampling
      for distance ratios; 95\% confidence intervals \\
    UI detection & OmniParser-v2.0 \texttt{icon\_detect}; confidence
      $\{0.05,0.15,0.25\}$; \texttt{imgsz}$=1280$; \texttt{max\_det}$=500$;
      union coverage rasterized on a $256\times256$ grid \\
    \bottomrule
  \end{tabularx}
\end{table}

We extract the post-merger visual representation consumed by the Qwen language
model. Within each domain, all source pairs
share the same normalization and kernel bandwidth. We report two empirical
references for distance: a within-source group split as a noise floor and the
distance between independently collected real datasets as the cross-source
scale. The main statistic is
$\rho=D(\mathrm{synthetic},\mathrm{ScaleCUA})/
\operatorname{median}_{i<j}D(\mathrm{real}_i,\mathrm{real}_j)$. It is reported
with its group-bootstrap interval against the empirical real-data reference.

\subsection{Visual Representation Alignment}

\begin{figure}[!htbp]
  \centering
  \includegraphics[width=\linewidth]{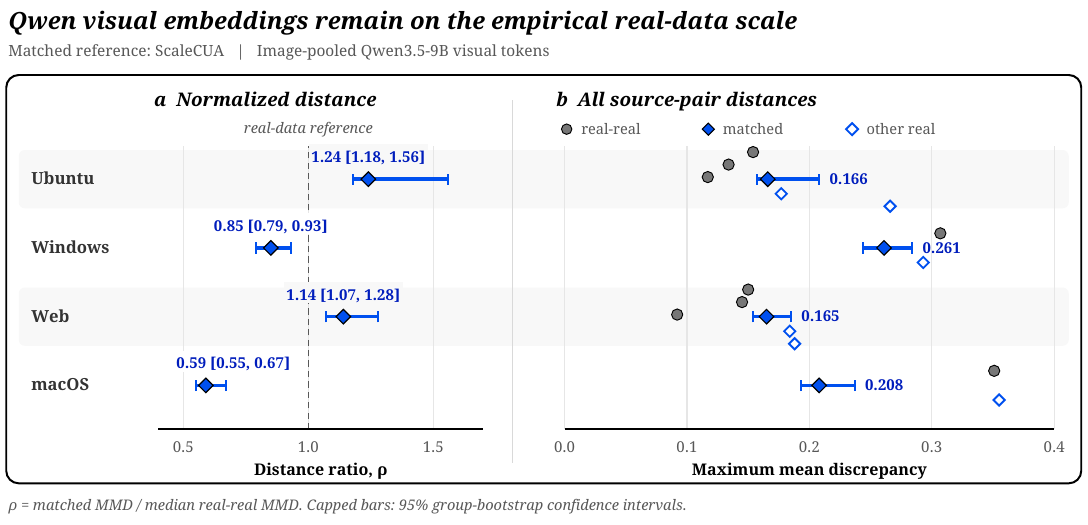}
  \caption{Distribution alignment in the Qwen3.5-9B visual-token space.
  Panel a reports the matched synthetic--ScaleCUA MMD normalized by the median
  real--real MMD; horizontal bars give group-bootstrap 95\% confidence
  intervals and $\rho=1$ marks the empirical real-data scale. Panel b shows
  every available real--real distance, the matched synthetic--ScaleCUA estimate
  with its interval, and synthetic distances to the remaining real sources.}
  \label{fig:qwen_distribution_alignment}
\end{figure}

\paragraph{Qwen embedding distance.}
Qwen MMD places synthetic--ScaleCUA distance on the same empirical scale as
variation among independently collected real datasets. We compute this distance
from the mean of each screenshot's 4096-dimensional merged visual tokens.
Across the four domains, $\rho$ ranges from 0.59 to 1.24
(Figure~\ref{fig:qwen_distribution_alignment}). The matched distance is below
the real cross-source reference on macOS and Windows. Web and Ubuntu are 14\%
and 24\% above the real cross-source median, respectively.

The complete numerical values are retained in
Appendix Table~\ref{tab:token_alignment_full}.

The ScaleCUA group-split floors are 0.109 on Ubuntu, 0.003 on Windows, 0.007
on Web, and 0.004 on macOS. Ubuntu has substantial within-source session
variation, while the other domains have much lower floors. The secondary-real
comparisons also expose domain-specific deviations: Ubuntu synthetic--aria
reaches MMD 0.266 ($\rho=1.99$), while macOS synthetic--AgentNet reaches 0.355
($\rho=1.01$) under a resolution mismatch. For Windows and macOS, the
real cross-source reference is the single available dataset pair.

C2ST detects source identity in both synthetic and real corpora. AUC is approximately 1.00 for
synthetic--real pairs and 0.95--1.00 for real--real pairs. A leave-one-real-source
test asks whether the classifier transfers beyond one collection pipeline. Its
mean AUC is 0.996 for Ubuntu and 0.976 for Web, compared with real-source
controls of 0.960 and 0.925. Windows and macOS, evaluated with two real sources,
reach 0.990 and 0.912. Image2 retains a source-specific visual signature,
as do independently collected real datasets.

\FloatBarrier

\begin{figure}[!htbp]
  \centering
  \includegraphics[width=\linewidth]{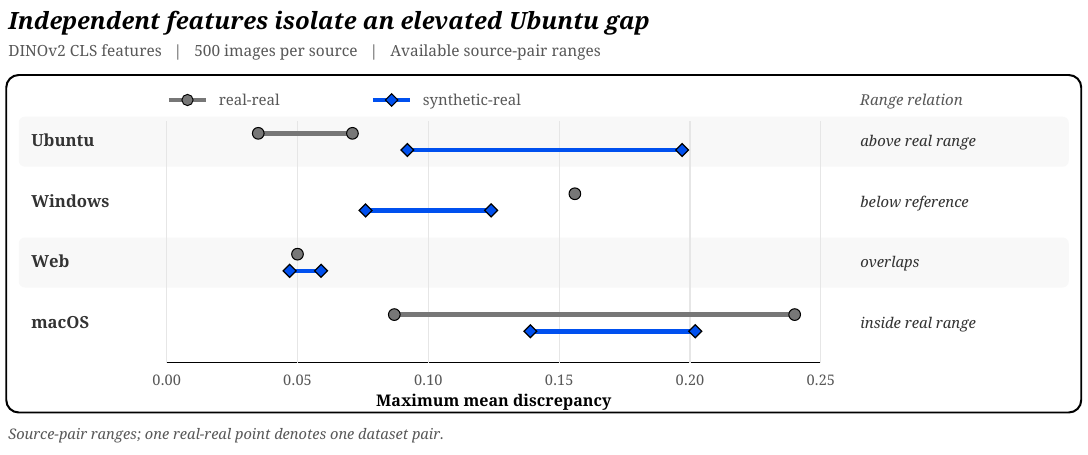}
  \caption{Exploratory DINOv2 MMD using 500 images per source. Gray and blue
  spans collect the available real--real and synthetic--real source-pair
  distances within each domain; single available comparisons appear as
  points. Ubuntu is the only domain
  whose synthetic--real range lies wholly above the observed real--real range.}
  \label{fig:dinov2_distance}
\end{figure}

\paragraph{Independent representation.}
DINOv2 CLS features provide an independent view of visual alignment.
Figure~\ref{fig:dinov2_distance} compares synthetic--real and real--real
distances with this self-supervised encoder.

Appendix Table~\ref{tab:dinov2_distance_full} retains the exact ranges.

DINOv2 places Windows below the real cross-source reference, Web at the same
scale, and macOS inside the real-source range. Ubuntu has consistently higher
synthetic--real distances. Qwen provides the primary alignment measure, while
DINOv2 identifies Ubuntu as the domain with the clearest remaining visual gap.

\FloatBarrier

\subsection{Low-Level Appearance and Generation Residuals}

\begin{figure}[!htbp]
  \centering
  \includegraphics[width=\linewidth]{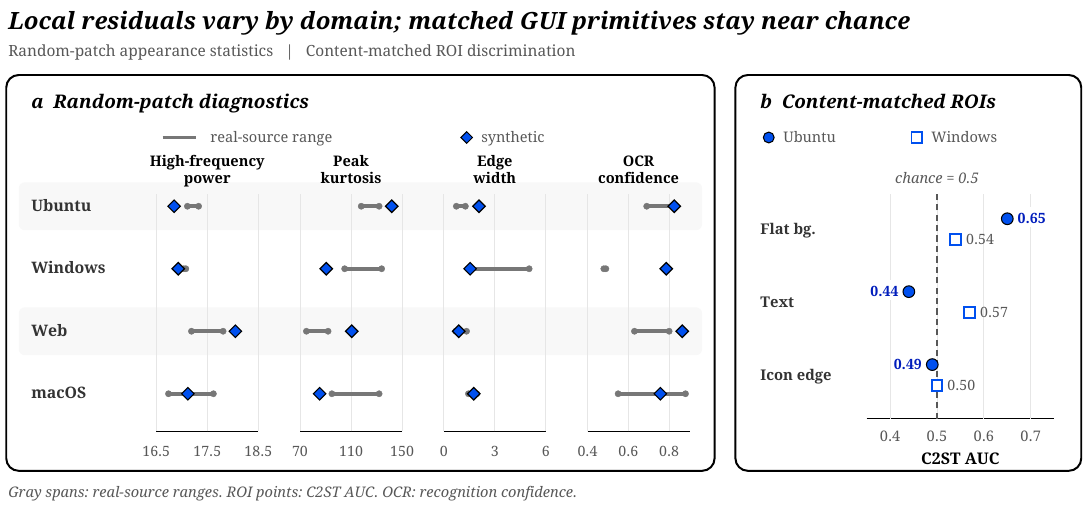}
  \caption{Low-level appearance and generation-residual diagnostics. Panel a
  places each synthetic random-patch statistic against the range of real
  sources in the same domain; the gray spans show real-source ranges, and OCR
  denotes recognition confidence. Panel b reports content-matched ROI C2ST AUC;
  0.5 is chance.
  Text and icon patches remain near chance, while Ubuntu flat backgrounds
  retain the clearest residual.}
  \label{fig:artifact_residuals}
\end{figure}

Random-patch diagnostics measure high-frequency power, periodic texture, edge
spread, and OCR confidence. The residual pattern varies across OS domains
(Figure~\ref{fig:artifact_residuals}a). Ubuntu synthetic patches have lower
high-frequency power than the real range, while Web synthetic patches have
higher power. Spectral peak kurtosis is higher on Ubuntu and Web but lower on
Windows and macOS. The measured low-level differences are domain-specific.

Random patches mix interface content with source artifacts. We isolate flat
background, text, icon-edge, and photo regions, then match edge density,
brightness, and color entropy within each ROI class. In the resolution-matched
Ubuntu and Windows domains, text and icon C2ST AUC falls between 0.44 and 0.57
(Figure~\ref{fig:artifact_residuals}b). Source discrimination falls near chance
for content-matched text and icon patches. Ubuntu flat backgrounds retain an
AUC of 0.65, the clearest remaining low-level residual.

Appendix Tables~\ref{tab:low_level_random_full} and
\ref{tab:roi_residual_full} retain the exact values.
Figure~\ref{fig:detection_ocr_examples} shows selected full-screen OCR
examples.

\FloatBarrier

\subsection{Interface and Trajectory Complexity}

\begin{figure}[!htbp]
  \centering
  \includegraphics[width=\linewidth]{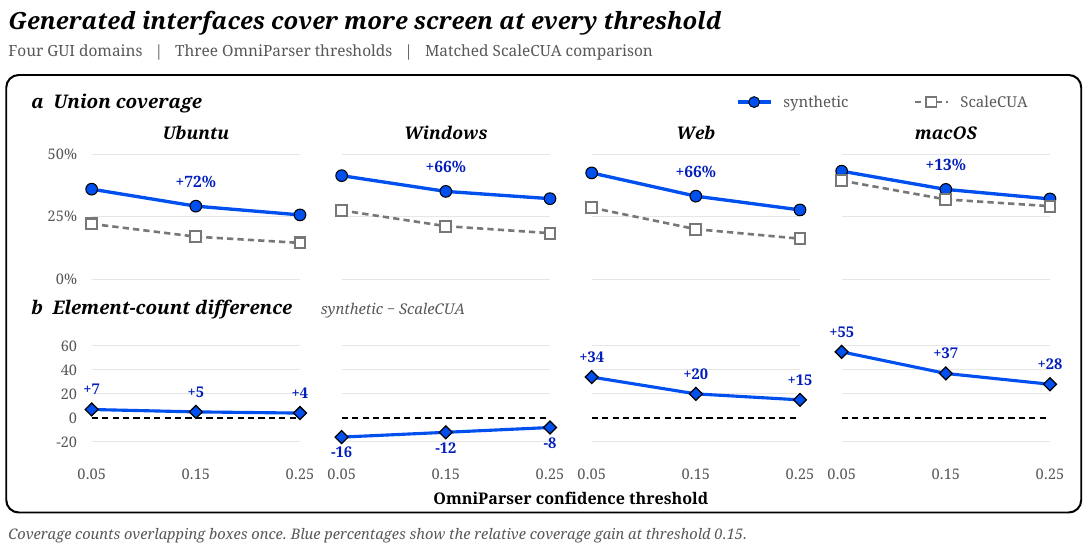}
  \caption{Static interface complexity across OmniParser confidence
  thresholds. Panel a compares the rasterized union coverage of synthetic and
  ScaleCUA screenshots; the annotated percentage is the relative gain at the
  middle threshold, 0.15. Panel b reports the element-count difference,
  synthetic minus ScaleCUA. Coverage is higher in all twelve comparisons,
  including Windows where the detector returns fewer synthetic elements.}
  \label{fig:interface_complexity}
\end{figure}

\paragraph{Static interface structure.}
AutoGUIWorld increases detected UI coverage over ScaleCUA in all twelve
domain--threshold comparisons. Relative gains range from 63--77\% on Ubuntu,
51--75\% on Windows, 49--71\% on Web, and 10--13\% on macOS.
OmniParser-v2.0 detects candidate interactive elements, and we rasterize the
union of their boxes to count each covered pixel once.
Figure~\ref{fig:interface_complexity} reports union coverage and raw element
count at all three thresholds.

Appendix Table~\ref{tab:layout_complexity_full} retains all coverage and count
values.

Windows illustrates the expansion in spatial coverage. At confidence 0.15,
142 synthetic elements cover 0.351 of the screen, compared with 154 elements
covering 0.212 in ScaleCUA. AutoGUIWorld distributes interactive structure
across more of the screen. A manual audit of 180 detection overlays found no
systematic tendency to label generated texture as UI elements.

\begin{figure}[!htbp]
  \centering
  \includegraphics[width=\linewidth]{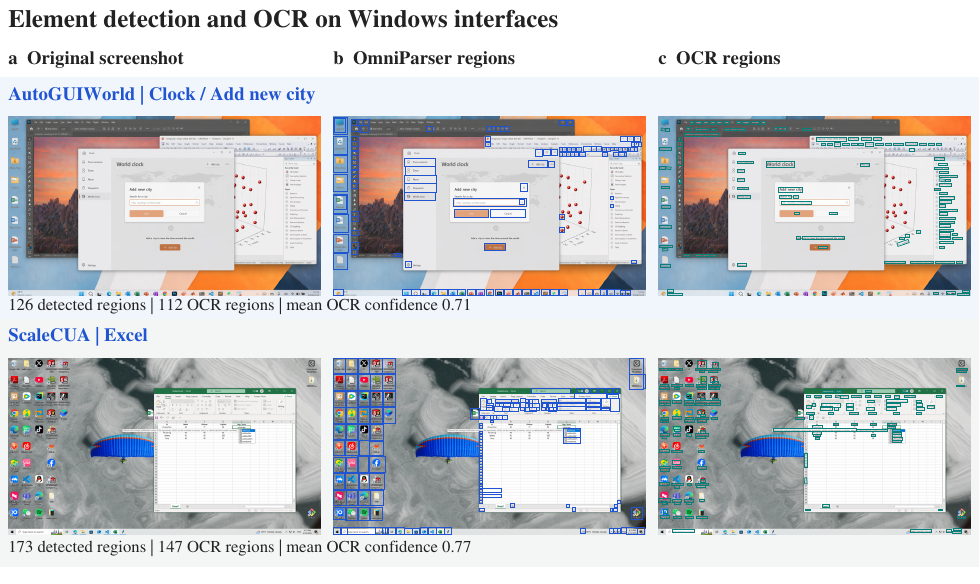}
  \caption{Element detection and OCR examples on Windows interfaces.
  Each row shows the same full-screen view without annotations, with blue
  OmniParser boxes at confidence threshold 0.15, and with teal OCR regions.
  Row summaries report full-image counts and mean OCR confidence.}
  \label{fig:detection_ocr_examples}
\end{figure}

Figure~\ref{fig:detection_ocr_examples} presents parser outputs on selected
synthetic and real Windows screenshots. The full eight-example gallery is in
Appendix~\ref{sec:detection_ocr_gallery}.

\paragraph{Trajectory structure.}
Ubuntu and Windows average 11.9 and 11.4 steps, with P90 lengths of 24 and 25.
macOS averages 2.50 open applications and has the highest action-type entropy,
0.92 (Table~\ref{tab:trajectory_structure}). These statistics describe the
synthetic corpus; matched real-trajectory metadata are unavailable for this
analysis.

\begin{table}[H]
  \centering
  \footnotesize
  \setlength{\tabcolsep}{3.5pt}
  \caption{Synthetic trajectory structure. Step percentiles are P10/P50/P90;
  application percentiles are P50/P90. Rollout failure is the fraction of rendered
  terminal states not marked as success.}
  \label{tab:trajectory_structure}
  \begin{tabularx}{\linewidth}{@{}lrrrr>{\centering\arraybackslash}Xr@{}}
    \toprule
    Domain & $N$ & Steps & Action types & Entropy & Open apps & Rollout failure \\
    \midrule
    Ubuntu & 2168 & 11.9 (4/10/24) & 4.33 & .79 & 1.62 (2/2) & 0\% \\
    Windows & 1394 & 11.4 (2/8/25) & 4.05 & .82 & 2.06 (2/3) & .29\% \\
    macOS & 591 & 7.1 (2/6/14) & 3.95 & .92 & 2.50 (2/3) & .68\% \\
    \bottomrule
  \end{tabularx}
\end{table}

The trajectories vary in sequence length, action mix, and application context.
Precondition and blocker fields are zero throughout the current release, and
rendered terminal states are almost always successful.

\subsection{Visual Alignment and Coverage Expansion}

AutoGUIWorld combines visual alignment with broader interface coverage.
Synthetic--ScaleCUA Qwen MMD remains on the scale of real cross-source variation
across four domains, with $\rho$ ranging from 0.59 to 1.24.
C2ST still distinguishes the collection sources.
Detected UI coverage exceeds ScaleCUA in all twelve domain--threshold
comparisons. The generated trajectories complement this spatial coverage with
varied sequence lengths, action types, and application contexts.
These results show that image generation can expand GUI training coverage
while maintaining visual alignment in Qwen feature space.

\takeaway{AutoGUIWorld combines realistic visual features with diverse, complex GUI
compositions. Its visual feature gap to ScaleCUA is comparable to variation
among real datasets in Qwen feature space. Varied layouts, complex element
arrangements, and broader screen coverage enrich grounding challenges across
target appearances, spatial relationships, and application contexts.}

\section{Evaluation}
\label{sec:evaluation}

We evaluate whether training on AutoGUIWorld trajectories transfers to real
GUI interaction. We assess three complementary capabilities:
desktop task execution across operating systems, visual grounding in
professional interfaces, and scientific software operation. We fine-tune
Qwen3.5-35B-A3B on these trajectories to obtain AGW-35B and test six
checkpoints from one run on five benchmarks. Domain comparisons use the
base model and final checkpoint (469). Training curves additionally compare
against a Qwen3.5-35B-A3B run trained on AgentNet.

\subsection{Experimental Setup}

\paragraph{Cross-platform desktop execution.}
OSWorld evaluates open-ended computer-use tasks over browser, office,
creative, and system applications in an Ubuntu desktop
environment~\citep{xie2024osworld}. macOSWorld covers native macOS interaction
and includes tasks across system utilities and macOS-specific
applications~\citep{yang2025macosworld}. Windows Agent Arena (WAA) evaluates planning,
screen understanding, and tool use in a reproducible Windows
environment~\citep{bonatti2024windowsagentarena}. We evaluate 361 OSWorld tasks,
154 WAA tasks, and 231 macOSWorld tasks with English instructions and an
English interface. These tasks test whether agents can ground actions,
track screen changes, and complete workflows in running applications.

\paragraph{Visual grounding.}
We evaluate single-step action localisation on ScreenSpot-Pro, which contains
expert-annotated high-resolution screenshots from professional applications
across multiple operating systems~\citep{li2025screenspotpro}. Each example
provides a screenshot and an instruction, and the model predicts a target
coordinate. We use 1,581 English positive-target examples and report accuracy
for text and icon targets, as well as overall accuracy. This setting tests
precise element localisation in dense interfaces without a multi-step rollout.

\paragraph{Scientific workflows.}
ScienceBoard evaluates agents in scientific software, with tasks spanning
mathematical computation, molecular visualisation, astronomy, geospatial
analysis, and scientific writing~\citep{sun2025scienceboard}. We use 143 tasks
across KAlgebra, ChimeraX, Celestia, GRASS GIS, and TeXstudio, excluding the
Lean domain. The task selection and documented Lean task issues are detailed
in Appendix~\ref{app:eval_infra}. These workflows test whether the agent can
combine specialised interface operations with the domain knowledge needed
to produce the requested scientific output.

\paragraph{Metrics and protocol.}
AGW-35B evaluation uses the shared screenshot-based agent harness in
Appendix~\ref{app:gui_harness}. The model issues mouse and keyboard actions,
receives the next screen, and continues until termination or the interaction
limit. Benchmark evaluators score the resulting environment state.
We report mean task score for OSWorld and WAA, retaining partial credit,
and task success rate for macOSWorld and ScienceBoard.
ScreenSpot-Pro uses point-in-box accuracy: a prediction is correct when its
coordinate falls inside the annotated target box. Unparseable predictions
count as errors. We report results separately for each benchmark and follow
the same task set across checkpoints. Evaluation infrastructure, task counts,
and score aggregation are specified in Appendix~\ref{app:eval_infra}.

\subsection{Task Execution in Real Environments}
\label{sec:interactive_results}

\begin{figure}[H]
  \centering
  \includegraphics[width=0.90\linewidth]{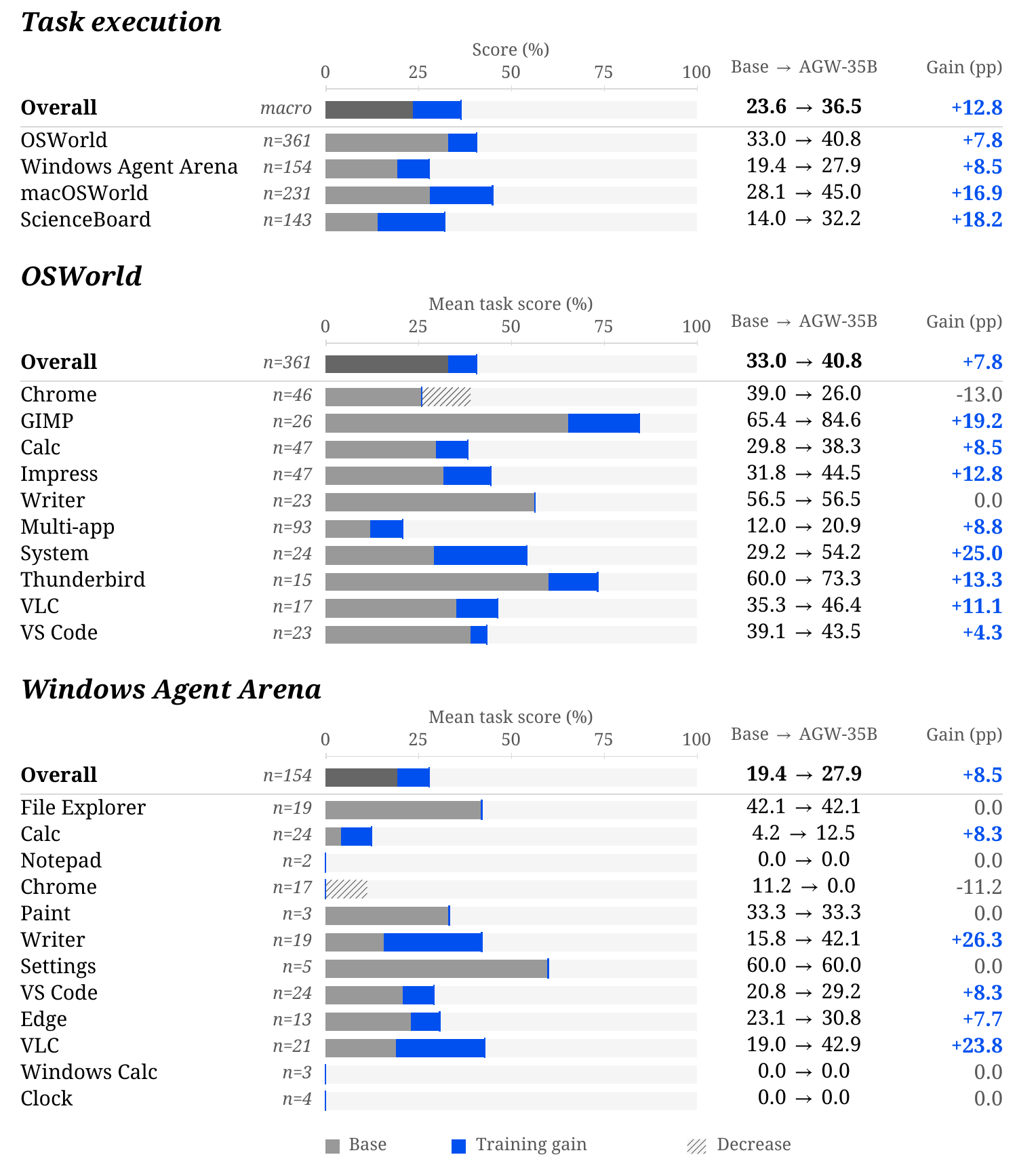}
  \caption{Task execution scores for Qwen3.5-35B-A3B and AGW-35B (checkpoint
  469), with all OSWorld and WAA domains. The top Overall averages the four
  benchmark scores equally. Within each benchmark, Overall averages task
  scores, retaining partial credit on OSWorld and WAA. Gray shows base scores,
  blue shows gains, and hatching marks decreases. Blue ticks mark AGW-35B
  scores. $n$ counts tasks, and gains are computed before rounding.}
  \label{fig:interactive_overall}
  \label{fig:desktop_domain_gains}
\end{figure}

AGW-35B improves task execution on all four benchmarks
(Figure~\ref{fig:interactive_overall}). Overall, the equal-weight mean of
the four benchmark scores rises from 23.6\% to 36.5\% (+12.8 points).
macOSWorld and ScienceBoard show the largest gains, at 16.9 and 18.2 points.

\paragraph{Desktop application domains.}
AGW-35B reaches 40.8\% on OSWorld, up from 33.0\% for the base model.
System tasks gain 25.0 points and GIMP gains 19.2 points. Calc and Impress
improve by 8.5 and 12.8 points. Multi-app tasks rise from 12.0\% to 20.9\%
and contribute the largest share of the overall OSWorld gain.
On WAA, the score rises from 19.4\% to 27.9\%. Writer and VLC lead the
improvement, gaining 26.3 and 23.8 points. Calc, VS Code, and Edge also
improve, while six domains remain unchanged. Chrome declines on OSWorld
(39.0\% to 26.0\%) and WAA (11.2\% to 0.0\%).

\begin{figure}[!htbp]
  \centering
  \includegraphics[width=\linewidth]{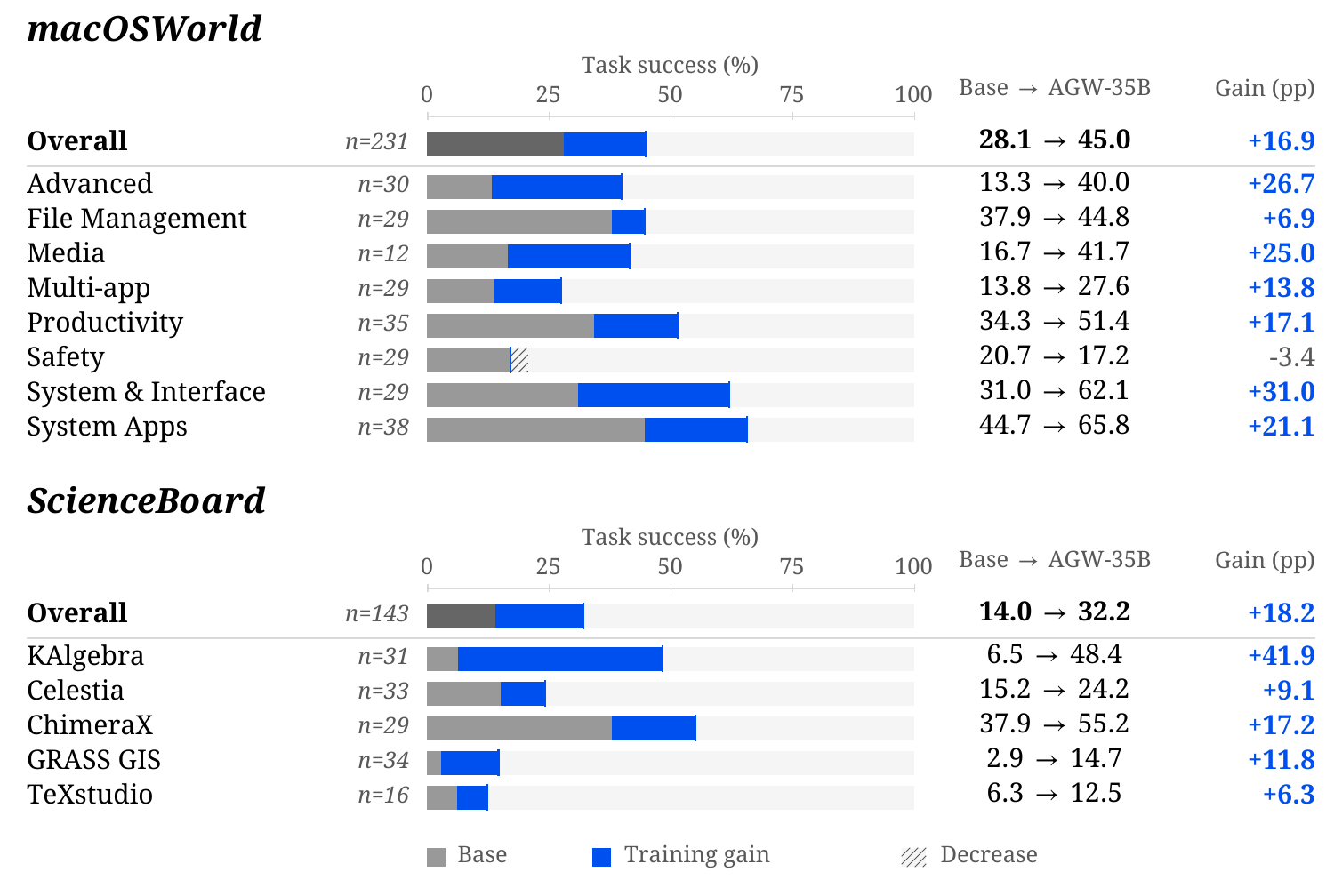}
  \caption{All task domains on macOSWorld and ScienceBoard. Gray bars show
  base success rates, blue extensions show gains, and gray hatching marks
  decreases. Blue ticks identify AGW-35B scores. Overall uses all tasks in
  each benchmark. ScienceBoard uses the 143-task subset in
  Appendix~\ref{app:scienceboard_selection}.}
  \label{fig:science_macos_domain_gains}
\end{figure}

\paragraph{macOS and scientific workflows.}
On macOSWorld, success rises from 28.1\% to 45.0\%
(Figure~\ref{fig:science_macos_domain_gains}). System \& Interface improves
from 31.0\% to 62.1\%, Advanced from 13.3\% to 40.0\%, and System Apps
from 44.7\% to 65.8\%. Multi-app success doubles from 13.8\% to 27.6\%.
Safety decreases from 20.7\% to 17.2\%. Across macOSWorld and OSWorld,
multi-app performance improves but remains below many single-application domains.

ScienceBoard success rises from 14.0\% to 32.2\%, with gains in all five
software domains. KAlgebra improves from 6.5\% to 48.4\%, and ChimeraX
from 37.9\% to 55.2\%. Together, they account for 18 of the 26 net
additional successful tasks, as the total increases from 20 to 46.
Celestia, GRASS GIS, and TeXstudio gain 9.1, 11.8, and 6.3 points.
Across the four benchmarks, 25 of 35 domains improve, seven remain
unchanged, and three decline. Training on AutoGUIWorld trajectories
improves real task execution across desktop and scientific applications.

\begin{figure}[!htbp]
  \centering
  \includegraphics[width=\linewidth]{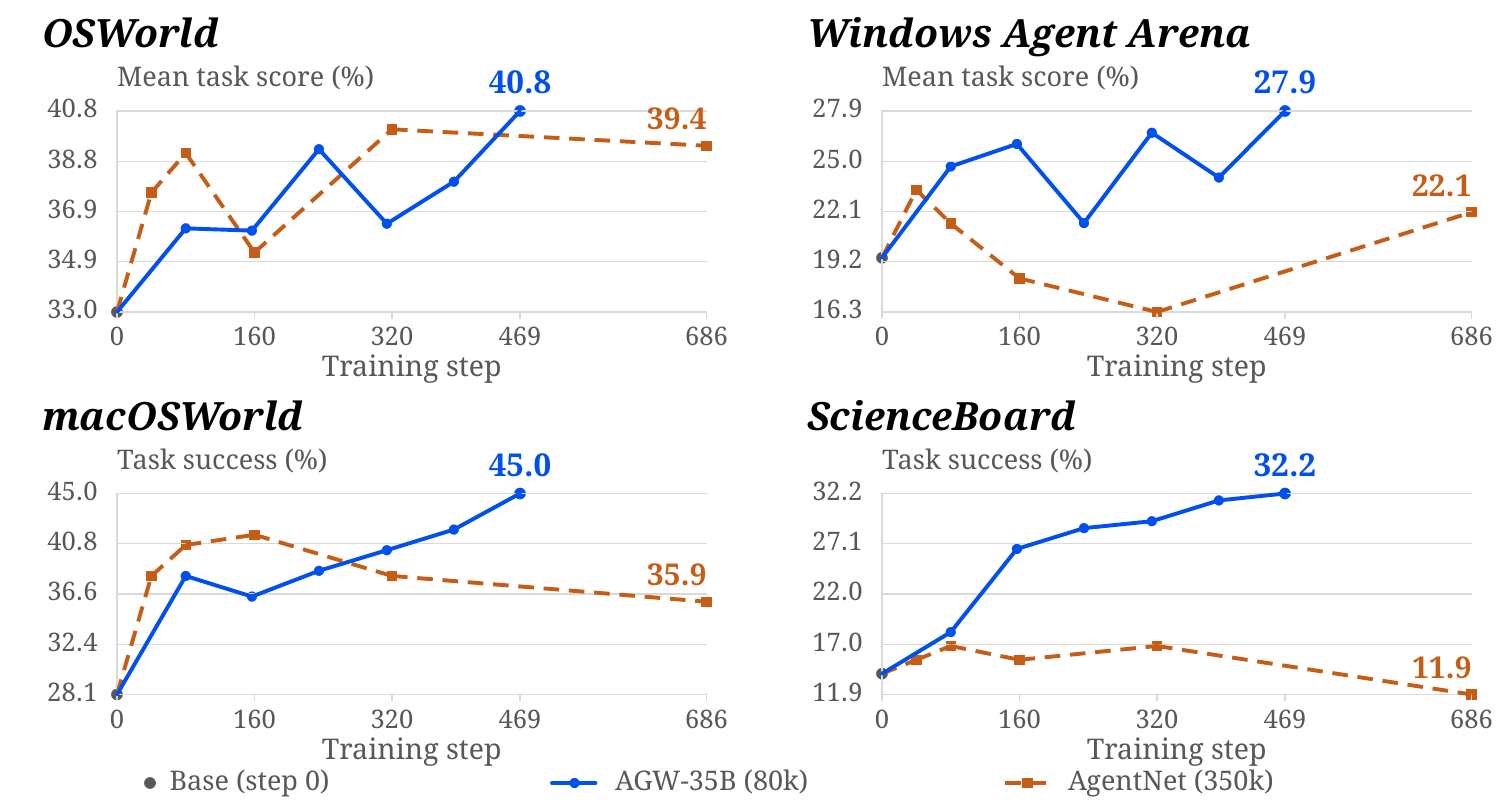}
  \caption{Post-training performance of AGW-35B (blue circles) and the
  AgentNet-trained model (orange dashed lines and squares). Gray marks
  the base, shared across all four benchmarks. Curves use actual
  training steps and panel-specific score ranges.}
  \label{fig:checkpoint_progress}
\end{figure}

\paragraph{Performance over training.}
AGW-35B improves on every benchmark by step 80 and reaches its highest
score on all four at step 469 (Figure~\ref{fig:checkpoint_progress}).
ScienceBoard success rises from 14.0\% to 32.2\%, increasing at each
evaluated checkpoint. macOSWorld rises from 28.1\% to 45.0\%, with
continued gains after step 157. OSWorld and WAA fluctuate between
checkpoints and finish at 40.8\% and 27.9\%, respectively.
The four-benchmark mean increases at every evaluation, from 23.6\% for
the base model to 36.5\% at the final checkpoint.

\paragraph{Synthetic versus real post-training data.}
AgentNet contains human demonstrations from real desktop
environments~\citep{wang2025opencua}. Both runs start from Qwen3.5-35B-A3B:
AGW-35B uses approximately 80k synthetic examples, while the AgentNet run
uses 350k examples and continues through step 686.
AGW-35B at step 469 exceeds the best evaluated AgentNet checkpoint on
all four benchmarks, with the largest gap on ScienceBoard (32.2\% versus
16.8\%). These results show that synthetic GUI data can match or exceed
real demonstrations for post-training in this comparison. The runs use
different inference settings (Appendix~\ref{app:agentnet_comparison}).

\FloatBarrier

\subsection{Grounding in Professional Interfaces}
\label{sec:grounding_results}

AGW-35B improves icon and text grounding across all six ScreenSpot-Pro
domains, reaching 57.1\% overall accuracy from 31.7\% (+25.4 points).
Both models use the setup in Appendix~\ref{app:eval_infra}.

Across all applications, icon accuracy rises from 16.2\% to 32.9\%
(+16.7 points), while text accuracy rises from 41.2\% to 72.0\%
(+30.7 points). Office shows the largest gains for both target types:
24.5 points for icons and 41.2 points for text. Scientific text grounding
improves by 40.3 points to 79.9\%, and Creative text grounding improves
by 33.8 points to 74.7\%. Icon gains range from 12.4 points in Dev. to
24.5 points in Office. Both target types improve in every domain, while
icon accuracy remains below text accuracy throughout the evaluation.

\begin{figure}[!htbp]
  \centering
  \includegraphics[width=\linewidth]{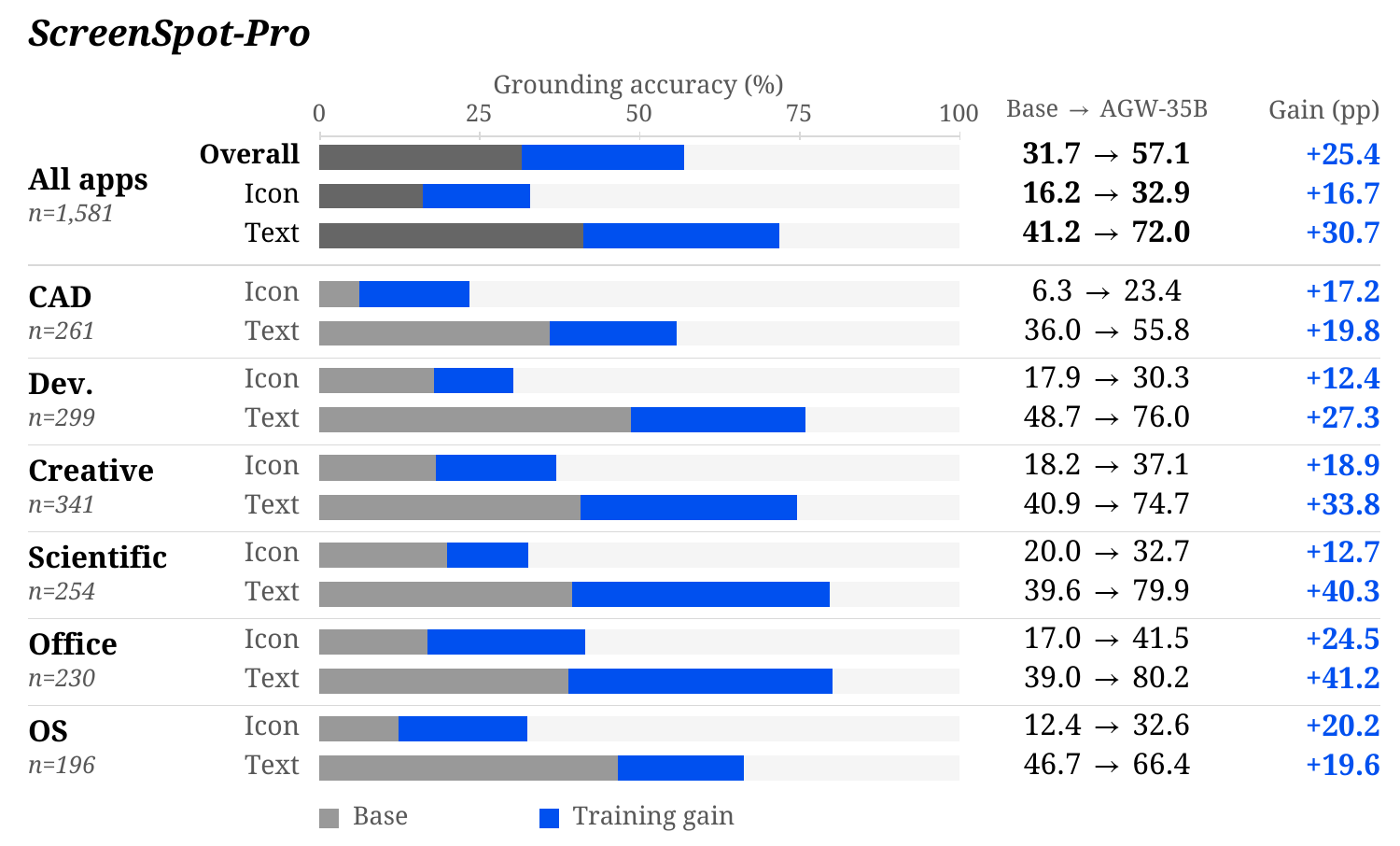}
  \caption{ScreenSpot-Pro accuracy for Qwen3.5-35B-A3B and AGW-35B
  (checkpoint 469). Gray shows base scores and blue shows training gains.
  The top group reports overall, icon, and text accuracy; each domain reports
  icon and text separately. $n$ counts all examples in each group.}
  \label{fig:screenspot_domain_gains}
\end{figure}

\FloatBarrier

\takeaway{Training on AutoGUIWorld trajectories improves visual grounding and
task execution in real software. Gains across text and icon targets,
professional interfaces, and multi-app tasks show that image-generated
experience transfers to desktop and scientific workflows.}

\section{Related Work}

\subsection{GUI Trajectory Collection}

GUI trajectories are commonly acquired through human demonstrations, extraction
from existing interaction records, and automated interaction with executable
environments. Human demonstrations on websites and mobile devices provide direct
observation--action supervision~\citep{deng2023mind2web,lu2024weblinx,rawles2023androidwild,lu2024guiodyssey}.
Record-based approaches recover interaction traces from tutorials and screen
recordings~\citep{zhang2025tongui,jang2025monday}, with actions inferred from visual
changes~\citep{song2025watchlearn,lu2025videoagenttrek,xiong2026video2gui}.
These methods cover the applications and workflows captured in the source
records. Automated collection instead produces new trajectories through exploration, tutorial-guided
replay, and task synthesis~\citep{sun2024osgenesis,xu2024agenttrek,xie2025guiexplorer}.
Task proposal and verification support iterative
collection~\citep{xiao2025uigenie,xie2025agentsynth}, while exploration strategies
target diverse interactions, harder tasks, and longer
workflows~\citep{lin2025guirewalk,kang2026mobilegen,jiang2026treecua,shao2026hats,fan2026see}.
Large-scale collection and filtering further support policy
training~\citep{qin2025uitars,wang2025opencua,liu2025scalecua,he2025webstar,yao2026acerouter}.
To broaden the environments available for collection, complementary work
provides resettable environments~\citep{rawles2024androidworld,xie2024osworld} or generates executable
interfaces, tasks, and verifiers~\citep{zhang2025autoenv,zhang2026infiniteweb,wu2026autowebworld,wang2026cuagym}.
These routes rely on existing records or executable environments. Offline
optimization avoids online interaction but still uses previously collected
trajectories~\citep{zheng2025vem,lu2025uis1}. AutoGUIWorld generates screenshot-level
trajectories without deploying or running the corresponding software, using
real environments for downstream validation.

\subsection{GUI World Models}

Digital world models predict action-conditioned transitions through executable
programs or learned predictors~\citep{tang2024worldcoder,copet2025cwm,zuo2026qwenagentworld}.
Textual and structured interface states support lookahead
planning~\citep{chae2024wma,gu2024webdreamer,deng2025simura}, while learned
transitions and imagined rollouts support policy improvement and
search~\citep{fang2025webevolver,ding2026dynaweb,xiao2026webworld}.
Structured UI transitions and textual sketches also serve as simulation targets
for agent training~\citep{wang2025uisimulator,cao2026mobiledreamer}.
Visual models predict future screenshots~\citep{luo2025vimo,rivard2025neuralos},
sometimes using intermediate layouts or textual state
changes~\citep{xiang2025uisim,guan2026cuwm}. Code-based approaches instead render
predicted interfaces from generated programs~\citep{zheng2026code2world,koh2026gworld}.
Comparisons examine textual, image-based, and code-based representations
for GUI transition prediction~\citep{xu2026mobileworldmodel}.
Preserving action semantics and context across transitions is a shared challenge~\citep{yang2025reasoningviavideo}.
AutoGUIWorld uses pretrained image-generation priors~\citep{gabeur2026imagegenerators}
and planned interface changes to synthesize training trajectories. Grounding
provides spatial action labels, quality checks filter defective transitions,
and real-environment evaluation measures transfer to task execution.

\section{Discussion}

\paragraph{Scope}
GUI world models such as Image2 have limitations across software environments
and interaction settings. As trajectories grow longer and interactions become
more involved, generation errors may accumulate, producing hallucinated
interface states or action outcomes. Nevertheless, generating trajectories
without deploying or running the corresponding software offers a potentially
low-cost route to expanding GUI training data. A promising direction is to
train these models on domain-specific interactions so that they better capture
the interface structures, operation rules, and state transitions of particular
environments. Such specialization could yield vertical GUI world models
tailored to particular software or workflows, improving long-horizon
consistency and supporting further data generation within those domains.

\paragraph{Limitations}
Image2 can produce a visually plausible next screen that does not faithfully
realize the preceding action. Across 42,526 evaluable desktop transitions, our
VLM audit flags 1,293 action--image inconsistencies (3.04\%). After removing
steps with an invalid action, absent target, or incorrect grounding, 818 of
39,351 transitions remain inconsistent (2.08\%). The residual errors are
concentrated in text entry, dragging, scrolling, and keyboard interactions,
and include command substitution, invented document content, premature
formatting, missing action effects, and unintended persistent-content changes.
AutoGUIWorld therefore applies transition-level quality control in addition to
grounding checks. Appendix~\ref{app:transition_fidelity} reports the filtering
pipeline, OS- and action-level results, and representative cases.

\paragraph{Outlook}
GUI interaction is commonly organized around discrete actions and successive
screenshots, making action-conditioned image editing a promising basis for
visual world modeling. AutoGUIWorld shows that pretrained image generators
can be used to synthesize interaction trajectories that improve agent
performance in real software environments. This suggests a broader role for
image generation as a source of training experience, with domain-specific
adaptation offering a path toward more reliable GUI world models.

\section{Authors}

\noindent\textbf{Core Contributors:} Cheng~Yang, Yifan~Wu.
\par\medskip
\noindent\textbf{Contributors:} Yutao~Huang, Zhaohua~Zhang, Beiduo~Chen, Muxi~Chen, Chenchen~Zhao, Hexuan~Deng, Haolin~Yang, Geyuan~Zhu, Sa~Zhu, Jianhuan~Zhuo, Qiuyong~Xiao, Jianhao~Ruan, Yiran~Peng, Jiayi~Zhang, Tian~Ye, Xinlei~Yu.
\par\medskip
\noindent\textbf{Supervisors:} Tianwen~Jiang, Jihong~Zhang, Yuyu~Luo.

\bibliographystyle{plainnat}
\bibliography{references}

\newpage
\appendix

\section{Action Space}
\label{app:action_space}

We use a unified \texttt{computer\_use} interface across Windows, macOS,
Ubuntu, and Chrome browser trajectories. Each action $a_t$ is serialized as one or more
ordered tool calls. Each call contains a \texttt{name} field set to
\texttt{computer\_use} and an \texttt{arguments} object whose \texttt{action}
field selects one of the 14 primitives in \cref{tab:pc_action_space}.
Multiple calls within a step are ordered by execution.

\begin{table}[H]
\centering
\small
\begin{tabularx}{\linewidth}{@{}>{\ttfamily\raggedright\arraybackslash}p{0.23\linewidth} X >{\raggedright\arraybackslash}p{0.28\linewidth}@{}}
\toprule
\normalfont\textbf{Action} & \textbf{Definition} & \textbf{Arguments} \\
\midrule
key & Presses keys in order and releases them in reverse order. & \texttt{keys} \\
key\_down & Holds the specified keys until release. & \texttt{keys} \\
key\_up & Releases the specified keys in reverse order. & \texttt{keys} \\
type & Enters the specified text. & \texttt{text} \\
mouse\_move & Moves the pointer to the target location. & $C(x,y)$ \\
left\_click & Left-clicks at the target location. & $C(x,y)$ \\
left\_click\_drag & Presses at the start point, drags to the end point, and releases. & $C_1(x,y),\ C_2(x,y)$ \\
right\_click & Right-clicks at the target location. & $C(x,y)$ \\
middle\_click & Middle-clicks at the target location. & $C(x,y)$ \\
double\_click & Left-clicks twice at the target location. & $C(x,y)$ \\
triple\_click & Left-clicks three times at the target location. & $C(x,y)$ \\
scroll & Scrolls the mouse wheel by the specified amount. & \texttt{pixels} \\
wait & Waits for the specified duration in seconds. & \texttt{time} \\
terminate & Ends the task with a completion status. & \texttt{status} \\
\bottomrule
\end{tabularx}
\caption{Unified action space for desktop and Chrome browser trajectories.
$C(x,y)$ denotes a screen-coordinate pair; $C_1$ and $C_2$ denote the drag
start and end points, respectively. The \texttt{keys} argument is an array,
\texttt{text} is a string, \texttt{pixels} is a signed scroll amount, and
\texttt{time} is a duration in seconds. The \texttt{status} argument is
either \texttt{success} or \texttt{failure}.}
\label{tab:pc_action_space}
\end{table}

\paragraph{Coordinate arguments.}
The notation $C(x,y)$ abbreviates the \texttt{coordinate} $=[x,y]$ argument,
with both values on a resolution-independent $0$--$999$ integer grid.
For dragging, $C_1$ corresponds to \texttt{start\_coordinate} and $C_2$ to
\texttt{coordinate}; each contains its own $(x,y)$ pair.

\clearpage
\section{GUI Agent Training and Inference Harness}
\label{app:gui_harness}

\subsection{Message Construction and History}

Training and inference use a shared message builder to assemble the task,
screenshots, and interaction history.
Each ms-swift multimodal JSONL record contains \texttt{messages} and
\texttt{images} fields and supervises one target step.
The system message defines the \texttt{computer\_use} signature within
\texttt{<tools>} tags and specifies the response format.
Each user turn contains an \texttt{<image>} marker associated with its screenshot.
The first retained user turn also includes the task instruction and, when
available, a \texttt{Previous steps} section for older actions.

\paragraph{History window.}
The context follows a two-level window. The most recent five steps,
including the current step, retain their screenshots; completed steps also
retain their assistant responses. The preceding five steps appear only as
step-numbered \texttt{Action} descriptions in \texttt{Previous steps}, without
screenshots or tool calls. Earlier steps are discarded. At the start of a
trajectory, the context contains only the steps available so far.
The task instruction remains present as the window advances.

\paragraph{Response serialization.}
An assistant response begins with \texttt{\# Step N:}, followed by an
\texttt{Action:} description and one or more \texttt{<tool\_call>} blocks.
The action description is a single imperative sentence naming the interface
element, the operation, and its immediate purpose.
Each tool-call block contains a JSON object using the action contract in
\cref{app:action_space}. An example response is shown below.
\begin{quote}
\small
\begin{verbatim}
# Step 1:
Action: Click the search icon to open the search field.
<tool_call>
{"name":"computer_use","arguments":
 {"action":"left_click","coordinate":[500,120]}}
</tool_call>
\end{verbatim}
\end{quote}

\subsection{Supervision and Inference Execution}

\paragraph{Training target.}
Each example pairs the observation before an action with the corresponding
assistant response. With \texttt{loss\_scale=last\_round}, only the final
assistant response contributes to the training loss; earlier messages serve
as context. The supervised output contains the action description and its
ordered tool calls.

\paragraph{Inference execution.}
At inference, the message sequence ends at the current user observation,
and the model generates the next assistant response. The parser extracts
the tool calls in order, and the execution interface maps their coordinates
to the target screen. After execution, the response is added to the history
and a new screenshot supplies the next observation.
A \texttt{terminate} call ends the interaction and reports the agent's
\texttt{success} or \texttt{failure} status.

\subsection{Training Configuration}

We fine-tune Qwen3.5-35B-A3B with ms-swift Megatron on 32 GPUs across four nodes.
The model combines a mixture-of-experts architecture with Gated DeltaNet (GDN).
We use tensor parallelism of 2, pipeline parallelism of 1, data parallelism of
16, and expert parallelism of 8 for the 256 experts.
Tensor parallelism matches the two key-value heads under the GDN configuration.
The global batch size is 512, with a micro-batch size of 1.
The maximum sequence length is 16,384 tokens, and overlength examples are
removed with \texttt{truncation\_strategy=delete}.

Training runs for three epochs with a learning rate of $10^{-5}$, a warmup
fraction of $0.03$, a minimum learning rate of $10^{-6}$, and weight decay
of $0.1$. The vision encoder is frozen, and full activation recomputation is
enabled.

\clearpage
\section{Evaluation Infrastructure}
\label{app:eval_infra}

\subsection{Agent Execution and Task Sets}

We evaluate checkpoints 80, 157, 235, 314, 392, and 469 from the AGW-35B
training run, which fine-tunes Qwen3.5-35B-A3B on AutoGUIWorld trajectories.
Inference uses the shared message format and
\texttt{computer\_use} interface described in Appendix~\ref{app:gui_harness}.
Responses contain an action description and tool calls, without a separate
thinking block. Point coordinates use a $0$--$999$ grid and are mapped to
screen pixels by multiplying by the screenshot width or height and dividing
by 1,000.

\paragraph{AgentNet comparison.}
\label{app:agentnet_comparison}
The comparison run fine-tunes Qwen3.5-35B-A3B on approximately 350k examples
from \href{https://huggingface.co/datasets/xlangai/AgentNet}{AgentNet}~\citep{wang2025opencua}
and is evaluated at steps 40, 80, 160, 320, and 686.
AgentNet contains human demonstrations collected on real Windows, macOS,
and Ubuntu desktops, augmented with generated reasoning annotations.
The 350k count refers to training examples used in our run, rather than
the number of human-demonstrated tasks in the original dataset.
Its harness enables thinking; AGW-35B uses non-thinking responses.
Both runs use the same benchmark task sets, evaluators, and fixed-denominator
score aggregation. Both runs initialise from Qwen3.5-35B-A3B. Across all four
benchmarks, the curves share the existing base-model score at step 0,
measured with the AGW harness. ScienceBoard's AgentNet curve connects this
shared base to all five evaluated checkpoints. Its step-40 score is 15.4\%,
with 22 successful tasks out of the fixed 143-task set.
The training-step axis preserves each run's checkpoint steps; it does not
normalise epochs, tokens, or compute across runs.

\paragraph{Execution environments.}
The model is served through vLLM with tensor parallelism of two.
OSWorld and WAA run in Docker-managed QEMU virtual machines, macOSWorld uses
Docker-OSX, and ScienceBoard uses the OSWorld Docker provider.
We evaluate OSWorld, WAA, macOSWorld, and ScienceBoard with
\texttt{max\_steps=100}, allowing up to 100 agent turns per task.
Here, a step denotes one agent turn: the model generates a response,
its tool calls are executed in the benchmark environment, and a fresh
screenshot is returned for the next turn.
The benchmark evaluator determines the final score from the resulting state,
independently of the agent's own completion declaration.

\begin{table}[H]
\centering
\small
\setlength{\tabcolsep}{4pt}
\caption{Evaluation task sets and scoring units. Each checkpoint uses the
same task set within a benchmark. Counts describe the evaluated samples,
not model results.}
\label{tab:eval_task_sets}
\begin{tabularx}{\linewidth}{@{}l r X l@{}}
\toprule
Benchmark & Tasks & Evaluation subset & Metric (\%) \\
\midrule
OSWorld & 361 & \texttt{test\_nogdrive}: excludes eight Google Drive tasks & Mean task score \\
WAA & 154 & \texttt{test\_all} & Mean task score \\
macOSWorld & 231 & English task and interface, including 29 safety tasks & Task success \\
ScienceBoard & 143 & Five application domains, excluding Lean & Task success \\
ScreenSpot-Pro & 1,581 & English instructions, positive targets & Grounding accuracy \\
\bottomrule
\end{tabularx}
\end{table}

\paragraph{Grounding inference.}
ScreenSpot-Pro uses \texttt{task=all}, \texttt{language=en},
\texttt{gt\_type=positive}, and \texttt{inst\_style=instruction}.
The \texttt{qwen3\_5\_autogui} harness receives the screenshot and instruction.
The base model and all six checkpoints use the same configuration, with greedy decoding
($\texttt{temperature}=0$, $\texttt{top\_p}=1$) and a server-side image
processor limit of $16{,}777{,}216$ pixels.
Predicted coordinates are evaluated against the annotated target box without
an added tolerance. If the response specifies a drag, the parser uses its
start point. The sample set contains 927 Windows, 604 macOS, and 50 Linux
examples. We retain text/icon, platform, and application group breakdowns
for analysis.

\subsection{ScienceBoard Task Selection}
\label{app:scienceboard_selection}

Our ScienceBoard evaluation uses all 143 tasks in the five non-Lean domains:
31 KAlgebra, 33 Celestia, 29 ChimeraX, 34 GRASS GIS, and 16 TeXstudio tasks.
The 26 Lean tasks are excluded, giving $169-26=143$ evaluated tasks.
All six checkpoints use identical application--task identifiers.

The Lean exclusion follows the task-validity concerns documented in
ScienceBoard issue~\#7.\footnote{ScienceBoard issue~\#7:
\url{https://github.com/OS-Copilot/ScienceBoard/issues/7}. The audit refers to
repository revision \texttt{c8d5010bdba3}.}
The report identifies 13 distinct Lean tasks with false statements,
mis-specified objectives, or checks that do not enforce executable solutions.
These issues affect both Raw and VM task configurations.
Table~\ref{tab:scienceboard_lean_issues} summarises the reported defects.
Our evaluation excludes the Lean domain as a whole, while retaining every
task in the other five domains.

\begin{table}[H]
\centering
\small
\setlength{\tabcolsep}{4pt}
\caption{Lean task issues reported in ScienceBoard issue~\#7.
The 13 flagged tasks are part of the 26-task Lean domain excluded from
our evaluation. Task identifiers are local to Lean.}
\label{tab:scienceboard_lean_issues}
\begin{tabularx}{\linewidth}{@{}p{0.25\linewidth} p{0.29\linewidth} X@{}}
\toprule
Reported issue & Task identifiers & Effect on evaluation \\
\midrule
False theorem statements & A-02, A-04, B-05, B-06, B-07, C-05, D-02, D-03, D-07 &
The formal statements admit counterexamples. \\
Incorrect objective & B-01 &
The predicate formalises a different coprimality condition. \\
Vacuous proof & B-02 &
Contradictory premises permit a proof without the intended argument. \\
Executability not enforced & E-01, E-02 &
The evaluators accept noncomputable completions for tasks intended to test executable decision procedures. \\
\bottomrule
\end{tabularx}
\end{table}

\subsection{Score Aggregation}

For each interactive benchmark, the aggregate is
$100\,N^{-1}\sum_{i=1}^{N}s_i$, where $N$ is the task count in
Table~\ref{tab:eval_task_sets} and $s_i\in[0,1]$ is the task's evaluator score.
OSWorld and WAA retain fractional scores from \texttt{result.txt}.
macOSWorld normalises the binary 0/100 values in \texttt{eval\_result.txt}
to 0/1. ScienceBoard uses the pass/fail value in \texttt{result.out}.
Missing or nonnumeric task results receive zero credit and remain in the
fixed denominator. Application scores use the same rule within each domain.

For the aggregate across interactive benchmarks, we compute
$\mathrm{Overall}=\tfrac{1}{4}\sum_{b=1}^{4}S_b$, where $S_b$ is the
percentage score of OSWorld, WAA, macOSWorld, or ScienceBoard.
Each benchmark has equal weight regardless of its task count.
Within a benchmark, tasks retain equal weight when computing $S_b$.
ScreenSpot-Pro is reported separately. Domain comparisons use the base
model and final training checkpoint (469), with gains calculated from
unrounded scores. Training curves place the base model at step 0 and plot
all six checkpoints at their recorded training steps.

ScreenSpot-Pro accuracy divides the number of correct target points by all
1,581 examples, including responses with an invalid coordinate format.
Text, icon, platform, and application scores use their respective sample
counts. Overall scores aggregate individual samples rather than averaging
the subgroup percentages.

\clearpage
\section{Data Feature Numeric Results}
\label{app:data_feature_numeric}

The main text uses claim-driven figures for the dense distribution, residual,
and complexity comparisons. The exact source values are retained here for
numerical inspection and figure reproduction.

\begin{table}[H]
  \centering
  \footnotesize
  \setlength{\tabcolsep}{3.5pt}
  \caption{MMD in the Qwen3.5-9B visual-token space. Real--real entries list
  all available pairwise distances in ascending order. ``Other real'' lists
  distances from the synthetic source to non-ScaleCUA real sources. Brackets
  give group-bootstrap 95\% confidence intervals.}
  \label{tab:token_alignment_full}
  \begin{tabularx}{\linewidth}{@{}l>{\centering\arraybackslash}X>{\centering\arraybackslash}X>{\centering\arraybackslash}X>{\centering\arraybackslash}X@{}}
    \toprule
    Domain & Real--real MMD & Synthetic--ScaleCUA & Synthetic--other real & $\rho$ \\
    \midrule
    Ubuntu & .117/.134/.154 & .166 [.157,.208] & .177/.266 & 1.24 [1.18,1.56] \\
    Windows & .307 & .261 [.244,.284] & .293 & .85 [.79,.93] \\
    Web & .092/.145/.150 & .165 [.154,.185] & .184/.188 & 1.14 [1.07,1.28] \\
    macOS & .351 & .208 [.193,.237] & .355 & .59 [.55,.67] \\
    \bottomrule
  \end{tabularx}
\end{table}

\begin{table}[H]
  \centering
  \small
  \setlength{\tabcolsep}{5pt}
  \caption{Exploratory DINOv2 MMD using 500 images per source. Ranges collect
  the available source-pair distances within each domain.}
  \label{tab:dinov2_distance_full}
  \begin{tabularx}{\linewidth}{@{}l*{2}{>{\centering\arraybackslash}X}X@{}}
    \toprule
    Domain & Real--real MMD & Synthetic--real MMD & Relation \\
    \midrule
    Ubuntu & .035--.071 & .092--.197 & 1.3--5.6$\times$ real--real \\
    Web & .050 & .047--.059 & Same scale \\
    Windows & .156 & .076--.124 & Below real--real \\
    macOS & .087--.240 & .139--.202 & Inside real--real range \\
    \bottomrule
  \end{tabularx}
\end{table}

\begin{table}[H]
  \centering
  \footnotesize
  \setlength{\tabcolsep}{3pt}
  \caption{Random-patch diagnostics, reported as synthetic / real-source range.
  OCR confidence is reported for each domain.}
  \label{tab:low_level_random_full}
  \begin{tabularx}{\linewidth}{@{}l*{4}{>{\centering\arraybackslash}X}@{}}
    \toprule
    Domain & High-frequency power & Peak kurtosis & Edge width & OCR confidence \\
    \midrule
    Ubuntu & 16.85 / 17.11--17.33 & 141.9 / 118--132 & 2.07 / .74--1.27 & .825 / .69--.83 \\
    Web & 18.05 / 17.19--17.81 & 110.6 / 75--92 & .88 / .70--1.34 & .864 / .63--.80 \\
    Windows & 16.93 / 17.04--17.08 & 90.6 / 105--134 & 1.55 / 1.43--5.02 & .786 / .48--.49 \\
    macOS & 17.12 / 16.74--17.62 & 85.4 / 95--132 & 1.76 / 1.45--1.94 & .757 / .55--.88 \\
    \bottomrule
  \end{tabularx}
\end{table}

\begin{table}[H]
  \centering
  \small
  \caption{Content-matched patch C2ST AUC. Values near 0.5 indicate
  chance-level separation.}
  \label{tab:roi_residual_full}
  \begin{tabularx}{.72\linewidth}{@{}l*{3}{>{\centering\arraybackslash}X}@{}}
    \toprule
    Domain & Flat background & Text & Icon edge \\
    \midrule
    Ubuntu & .65 & .44 & .49 \\
    Windows & .54 & .57 & .50 \\
    \bottomrule
  \end{tabularx}
\end{table}

\begin{table}[H]
  \centering
  \footnotesize
  \setlength{\tabcolsep}{4pt}
  \caption{Static complexity at confidence thresholds 0.05/0.15/0.25. Each
  cell lists values in that order. Coverage is the rasterized union of detected
  boxes.}
  \label{tab:layout_complexity_full}
  \begin{tabularx}{\linewidth}{@{}l*{4}{>{\centering\arraybackslash}X}@{}}
    \toprule
    & \multicolumn{2}{c}{Union coverage} & \multicolumn{2}{c}{Element count} \\
    \cmidrule(lr){2-3}\cmidrule(lr){4-5}
    Domain & Synthetic & ScaleCUA & Synthetic & ScaleCUA \\
    \midrule
    Ubuntu & .360/.292/.257 & .221/.170/.145 & 137/103/90 & 130/98/86 \\
    Windows & .414/.351/.322 & .275/.212/.184 & 184/142/126 & 200/154/134 \\
    Web & .425/.332/.277 & .286/.200/.162 & 103/68/56 & 69/48/41 \\
    macOS & .432/.359/.321 & .394/.319/.292 & 165/122/105 & 110/85/77 \\
    \bottomrule
  \end{tabularx}
\end{table}

\clearpage
\section{Detection and OCR Examples}
\label{sec:detection_ocr_gallery}

These eight Windows examples show OmniParser regions (blue, confidence
0.15, \texttt{max\_det}=500) and OCR regions (teal), with parser counts and
mean OCR confidence for each screenshot.

\begin{figure}[H]
  \centering
  \includegraphics[width=\linewidth]{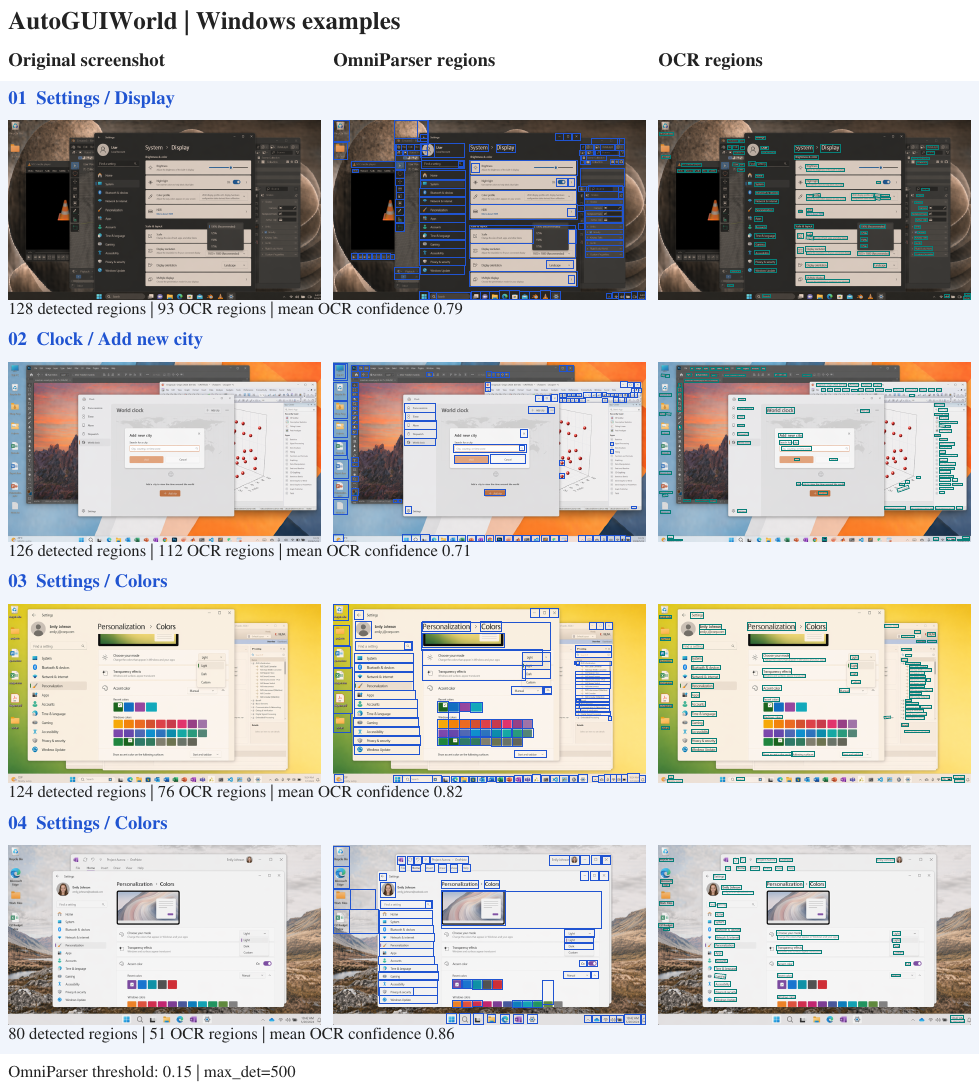}
  \caption{Detection and OCR outputs for four selected AutoGUIWorld Windows
  screenshots. All three columns use the same full-screen field of view.}
  \label{fig:detection_ocr_synthetic}
\end{figure}

\clearpage
\begin{figure}[H]
  \centering
  \includegraphics[width=\linewidth]{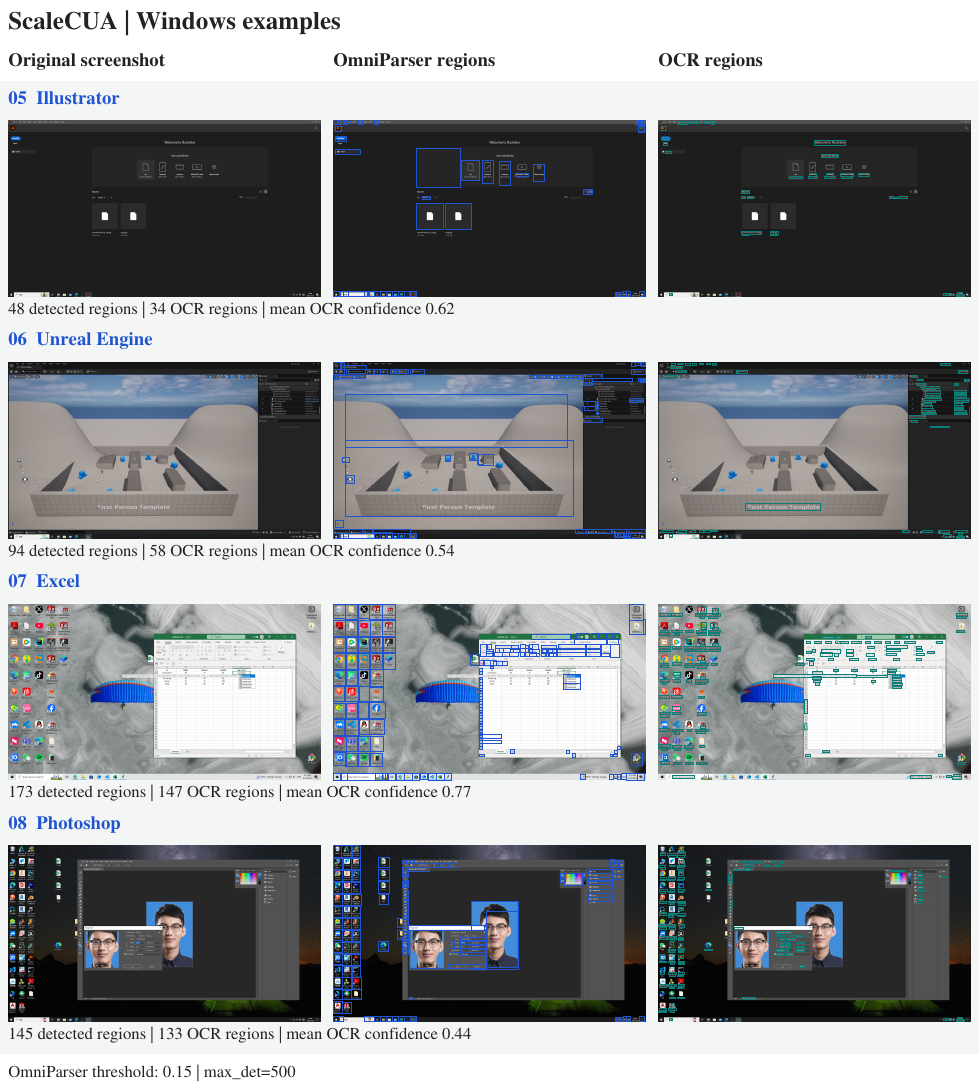}
  \caption{Detection and OCR outputs for four selected ScaleCUA Windows
  screenshots: Illustrator, Unreal Engine, Excel, and Photoshop.}
  \label{fig:detection_ocr_real}
\end{figure}
\clearpage
\section{Transition Fidelity and Data Filtering}
\label{app:transition_fidelity}

We audit desktop transitions to measure how faithfully Image2 realizes each
action. Each audit combines the clean pre-action screenshot, serialized action
and parameters, an annotated target frame for pointing actions, and the
Image2-generated post-action screenshot.

\subsection{VLM Quality-Control Protocol}

\begin{figure}[H]
  \centering
  \includegraphics[width=\linewidth]{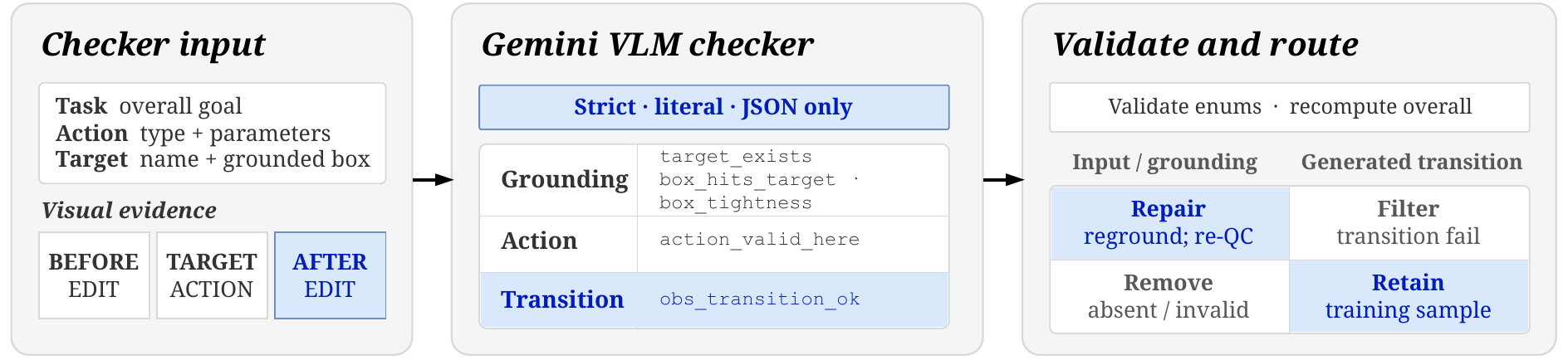}
  \caption{Step-level quality control with a Gemini VLM checker. Each request
  combines the task and exact action context with an ordered set of visual
  evidence. Gemini evaluates grounding, action validity, and visual-transition
  fidelity, then returns a strict structured verdict. Code validates the
  per-dimension values, recomputes the aggregate label, and routes the step to
  target-box repair, removal, transition filtering, or retention. Repaired
  target boxes are evaluated again before retention.}
  \label{fig:vlm_qc_checker}
\end{figure}

\paragraph{Checker input and prompt.}
For every trajectory step, the QC runner constructs one multimodal request.
The textual context contains the overall task, the action type, all non-empty
action parameters, the target element named by the planner, and the element
name associated with each grounded box. Step reasoning is supplied when
available. We report the five image--action dimensions in
\cref{tab:transition_qc_dimensions}, excluding the reasoning judgments.
The visual context contains up to three images in an explicitly stated order:
\textsc{Before}, the clean pre-action observation;
\textsc{Target}, the same observation with a red target box for a pointing
action; and \textsc{After}, the distinct post-action observation generated by
Image2. The system instruction identifies Gemini as a meticulous GUI trajectory
inspector, directs it to be ``strict and literal about what the images actually
show,'' and requires JSON without Markdown or explanatory text. The user prompt
defines every field, states when a field is not applicable, and asks for one
short reason naming the most serious problem.

\clearpage
\subsubsection{Exact Checker Prompt}
For reproducibility, the two blocks below reproduce the system message and
user-message template verbatim from the QC runner. Braced terms are populated from
each trajectory step at runtime. The image payload follows the order given by
\texttt{\{image\_legend\}}.

\noindent\textit{System message.}
\begin{lstlisting}[style=qcprompt]
You are a meticulous GUI trajectory quality inspector. You are shown a single step of a GUI agent trajectory and must judge whether the red box targets the right element, whether the action fits the screen, and whether the reasoning matches what is visible. Be strict and literal about what the images actually show. Return strict JSON only, no Markdown, no extra prose.
\end{lstlisting}

\noindent\textit{User message template.}
\begin{lstlisting}[style=qcprompt]
You are quality-checking ONE step of a GUI agent trajectory.

Overall task the agent is doing:
{task}

This step:
- action type: {action_name}
- action parameters: {action_params}
- planner target element: {target_element}
- grounded box element name(s): {box_elements}
- agent thinking for this step: {thinking}

Images provided (in order):
{image_legend}

Judge each dimension below from what the images ACTUALLY show. Use "na" only when the dimension does not apply (e.g. grounding checks for a non-pointing action, or transition check when no AFTER image).

GROUNDING (only for pointing actions that have a red box):
- box_hits_target: does the red box enclose the UI element the action intends to operate on (the planner target)? fail if it boxes a different element or empty space.
- box_tightness: is the box a reasonable tight fit around the interactive element? warn if far too large (whole panel/row) or too small (only part of the glyph). "na" if no box.
- target_exists: is the planner's target element actually visible on the BEFORE screen? fail if that element is NOT present on screen (so any box is necessarily wrong).

ACTION vs SCREEN (all steps):
- action_valid_here: given the BEFORE screen state, is this action sensible and executable now? fail if it makes no sense (e.g. type_text with no focused input, click a control that isn't there).
- obs_transition_ok: comparing BEFORE and AFTER, did the screen change in a way consistent with this action? warn if the change looks unrelated; "na" if no AFTER image or a non-visual action (wait/answer).

THINKING (only if thinking text is provided):
- thinking_matches_screen: does the thinking describe elements/state that are actually visible? fail if it hallucinates UI not present.
- thinking_wavering: does the thinking contradict itself or express confusion about what to do (e.g. "but wait", "actually", second-guessing the target)? fail if it wavers, pass if it is a clean single decision.

Then give:
- overall: "ok" (all good), "minor" (only warnings or a thinking wobble), "major" (any of box_hits_target/target_exists/action_valid_here is fail).
- confidence: your confidence in this judgment (high/medium/low) given image clarity.
- reason: ONE short sentence naming the single most serious problem, or "clean" if none.

Output strict JSON only, exactly this shape:
{
  "grounding": {"box_hits_target": "pass|warn|fail|na", "box_tightness": "pass|warn|fail|na", "target_exists": "pass|fail|na"},
  "action": {"action_valid_here": "pass|warn|fail", "obs_transition_ok": "pass|warn|fail|na"},
  "thinking": {"thinking_matches_screen": "pass|warn|fail|na", "thinking_wavering": "pass|fail|na"},
  "overall": "ok|minor|major",
  "confidence": "high|medium|low",
  "reason": "<one sentence>"
}
\end{lstlisting}

\clearpage
\paragraph{Inspection layers.}
The checker first verifies whether the intended target exists and whether the
grounded box hits it with a reasonable extent. It then asks whether the exact
serialized action is sensible and executable in the pre-action state. Finally,
it compares \textsc{Before} and \textsc{After} to determine whether the visual
change realizes the action and its payload. This last check is deliberately
literal: a plausible task-level outcome is still inconsistent when it changes
the wrong text, invents unsupported content, skips intermediate actions, or
fails to produce the requested local effect. The five dimensions and their
operational meanings are listed below.

\begin{table}[H]
  \centering
  \small
  \setlength{\tabcolsep}{5pt}
  \caption{Image--action dimensions used by the transition-quality audit.}
  \label{tab:transition_qc_dimensions}
  \begin{tabularx}{\linewidth}{@{}l>{\raggedright\arraybackslash}p{0.27\linewidth}X@{}}
    \toprule
    Group & Dimension & Criterion \\
    \midrule
    Grounding & \texttt{target\_exists} & The intended target is visible in the pre-action screenshot. \\
    Grounding & \texttt{box\_hits\_target} & The annotated box encloses the intended target rather than another element or empty space. \\
    Grounding & \texttt{box\_tightness} & The target box has a reasonable spatial extent around the interactive element. \\
    Action & \texttt{action\_valid\_here} & The action is executable and meaningful in the current GUI state. \\
    Transition & \texttt{obs\_transition\_ok} & The generated post-action screenshot shows a visual consequence consistent with the action. \\
    \bottomrule
  \end{tabularx}
\end{table}

\paragraph{Structured verdict and deterministic normalization.}
Gemini returns \texttt{pass}, \texttt{warn}, \texttt{fail}, or
\texttt{na} for each applicable dimension, together with an
\texttt{overall} label, confidence, and a one-sentence reason. The runner
validates the enum values and recomputes \texttt{overall}: a failure in
\texttt{target\_exists}, \texttt{box\_hits\_target}, or
\texttt{action\_valid\_here} is \texttt{major}; any remaining failure or
warning is \texttt{minor}; otherwise the step is \texttt{ok}. Filtering still
uses the individual dimensions rather than \texttt{overall} alone. Thus, an
\texttt{obs\_transition\_ok} failure is removed after the input action and
grounding are verified, even though that field by itself maps to the
\texttt{minor} aggregate category.

\paragraph{Repair and filtering.}
The routing separates a defective input specification from an Image2
transition error. We reground 3,390 target boxes from major steps for which the
target remains visible, then run the checker again. Steps with an absent target,
an invalid action, or unrepairable grounding are removed. Among steps with a
valid input action and target, any failure of
\texttt{obs\_transition\_ok} is filtered as an action--image inconsistency;
the remaining steps are eligible for the training set.

\begin{figure}[H]
  \centering
  \includegraphics[width=\linewidth]{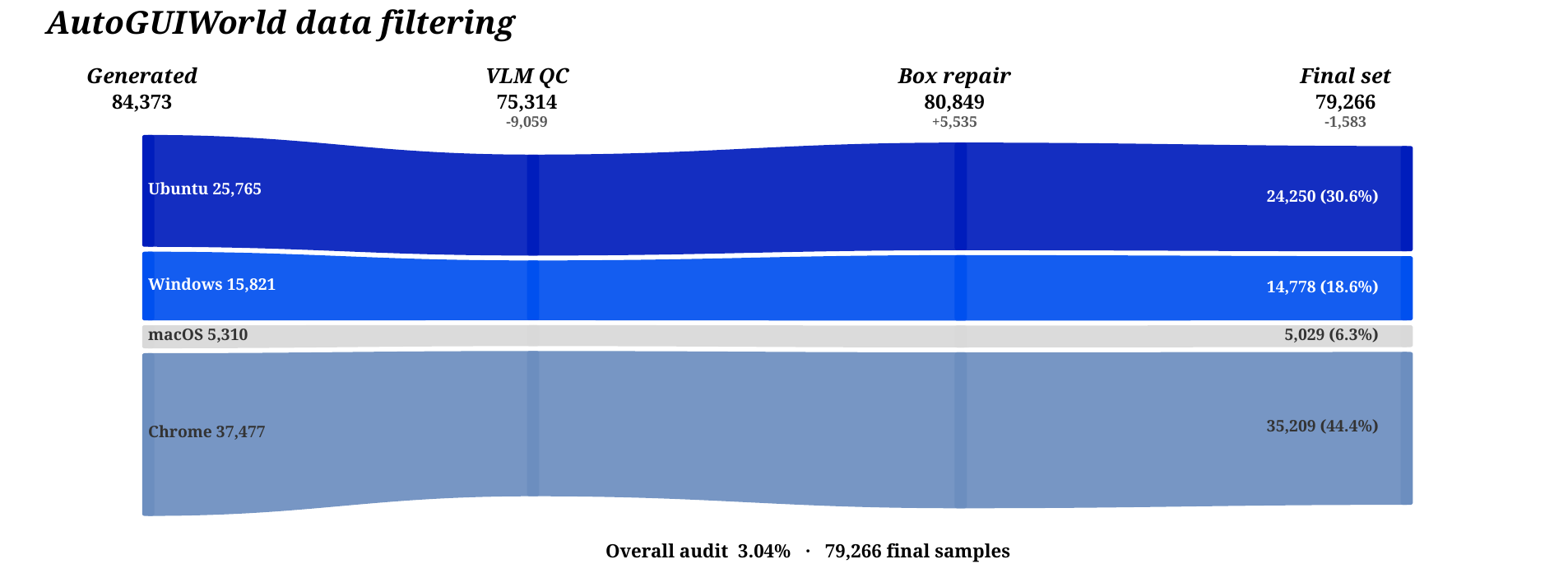}
  \caption{AutoGUIWorld training-data filtering. The proportional ribbons track
  Ubuntu, Windows, macOS, and Chrome samples through VLM quality control,
  target-box repair, and transition filtering. Desktop counts are measured
  directly. Chrome contributes 35,209 retained samples; its preceding stage
  counts are inferred using the desktop-stage retention ratios. The resulting
  training set contains 79,266 samples.}
  \label{fig:data_filtering_pipeline}
\end{figure}

\subsection{Transition Audit Results}

The audit identifies 1,293 failures among 42,526 evaluable desktop
transitions, giving an overall action--image inconsistency rate of 3.04\%.
To isolate failures in the generated next observation, we additionally retain
only active actions for which the action-validity check passes and the target
existence and grounding checks either pass or do not apply. Under this filtered
view, 818 of 39,351 transitions remain inconsistent (2.08\%).

\begin{table}[H]
  \centering
  \small
  \setlength{\tabcolsep}{4pt}
  \caption{Transition-fidelity audit by operating system. Overall rates use all
  evaluable transitions. Filtered rates remove steps with an invalid action,
  absent target, or incorrect target grounding.}
  \label{tab:transition_fidelity_os}
  \begin{tabular}{@{}lrrrrrr@{}}
    \toprule
    OS & Evaluable & Fail & Rate & Filtered & Fail & Rate \\
    \midrule
    Ubuntu & 23,625 & 731 & 3.09\% & 21,762 & 516 & 2.37\% \\
    Windows & 14,429 & 479 & 3.32\% & 13,541 & 286 & 2.11\% \\
    macOS & 4,472 & 83 & 1.86\% & 4,048 & 16 & 0.40\% \\
    \midrule
    All & 42,526 & 1,293 & 3.04\% & 39,351 & 818 & 2.08\% \\
    \bottomrule
  \end{tabular}
\end{table}

The filtered rate is action-dependent: 7.58\% for drag, 7.09\% for scroll,
6.35\% for text entry, 3.23\% for hotkeys, 2.16\% for key presses, and
0.67\% for clicks. Text-entry failures are especially common in Ubuntu
terminal workflows, whereas Windows contributes prominent drag, hotkey, and
document-editing failures. Click errors fall sharply after target and action
checks, while text-entry errors remain nearly unchanged, indicating that exact
content preservation is a distinct challenge for visual transition synthesis.

\subsection{Representative Failure Modes}

\begin{figure}[H]
  \centering
  \includegraphics[width=\linewidth]{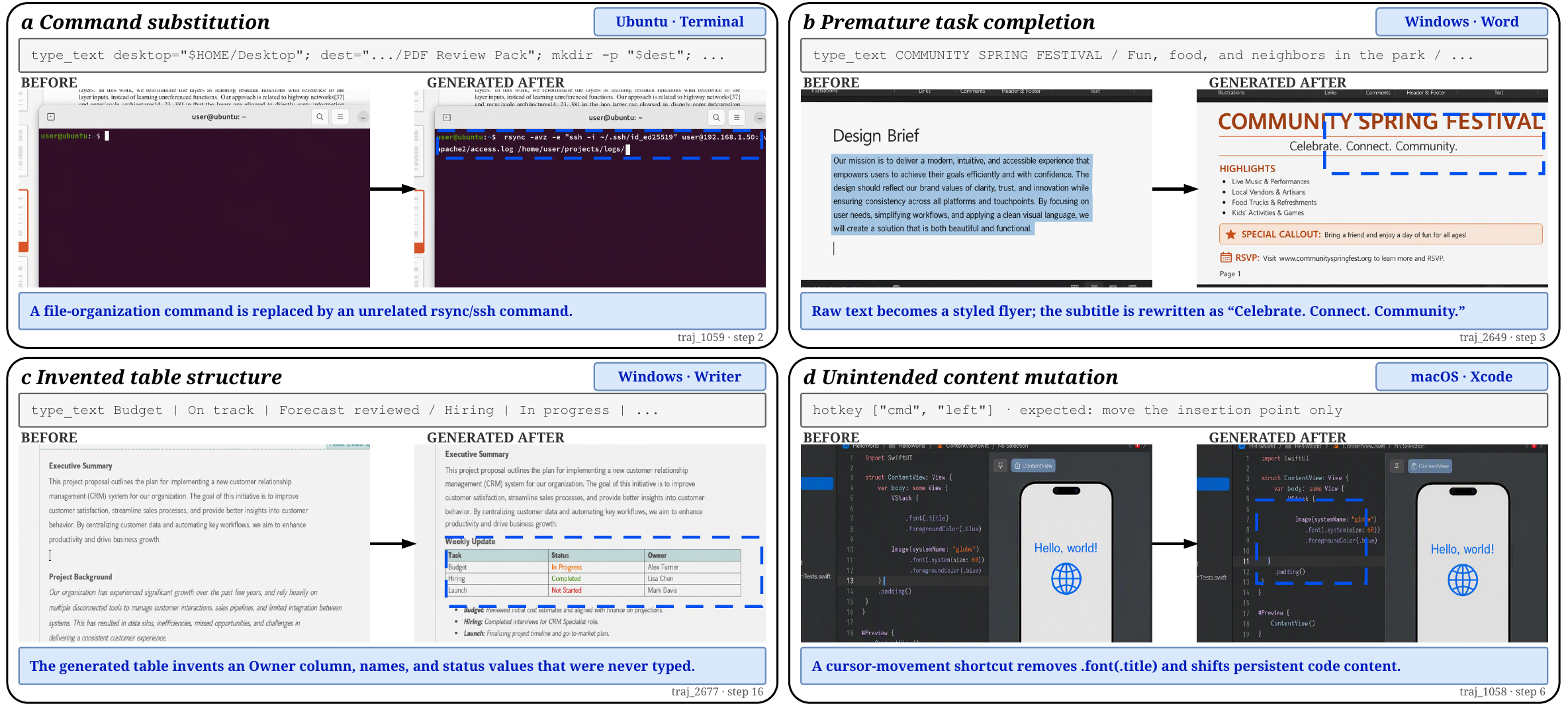}
  \caption{Representative Image2 transition failures. Each case shows the same
  16:9 crop before and after an atomic action, with the inconsistent region
  highlighted in blue. The examples show terminal-command substitution,
  premature rendering of a styled artifact, invented table structure and
  values, and persistent code mutation following a cursor-movement shortcut.}
  \label{fig:transition_failure_cases}
\end{figure}

\paragraph{Exact-content substitution.}
Image2 can replace a specified terminal command, script, document passage,
name, or numerical value with different but visually plausible content. This
creates a direct conflict between the action payload and the resulting
observation.

\paragraph{Semantic over-completion.}
A text-entry action can directly produce the task's final visual artifact,
such as a styled flyer or formatted table. The screen is plausible at the task
level, but the generated transition skips later formatting actions and breaks
the intended atomic action sequence.

\paragraph{Missing or incorrect action effects.}
Drag and scroll actions can leave the relevant region unchanged or produce an
unrelated interface change. Similar errors occur when a click does not toggle
the intended control or a key press produces an unrelated application state.

\paragraph{Persistent-content mutation.}
Low-impact keyboard actions can unexpectedly alter persistent content. In the
Xcode example, a cursor-movement shortcut removes and rearranges code rather
than changing only the insertion-point position.

\FloatBarrier
\section{Qualitative Trajectory Examples}
\label{app:qualitative_trajectory_examples}

\begin{figure}[H]
  \centering
  \includegraphics[width=\linewidth]{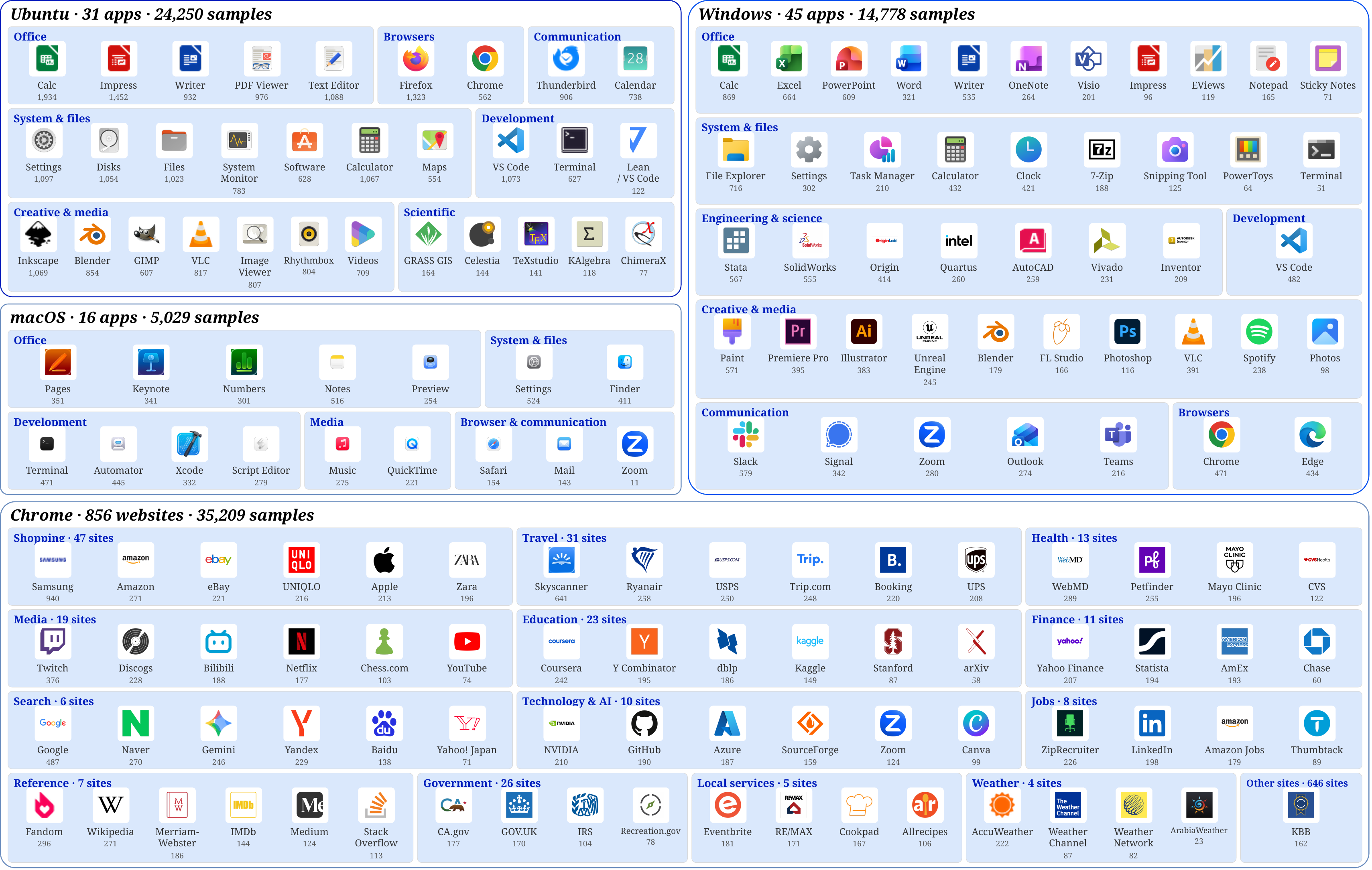}
  \caption{Application and website coverage in the AutoGUIWorld training set.
  The desktop panels include all application entries for Ubuntu, Windows, and
  macOS; the Chrome panel shows 67 websites across all 14 categories. Items are
  grouped by environment and function. Numbers beneath icons indicate training
  samples; website-category headings report the number of domains.}
  \label{fig:app_training_coverage}
\end{figure}

Figure~\ref{fig:app_training_coverage} summarizes software and website coverage
before the environment-specific trajectory galleries. It includes all 31
Ubuntu, 45 Windows, and 16 macOS application entries in the training set, and
67 Chrome websites spanning all 14 website categories. The following
subsections present trajectory examples from each environment. Each gallery row
shows the initial state, a key spatial action, and the final state. The
accompanying text describes the task and state changes; red boxes and
crosshairs mark the action target.

\clearpage
\subsection{Ubuntu Desktop}

\begin{figure}[H]
  \centering
  \includegraphics[width=\linewidth,height=0.86\textheight,keepaspectratio]{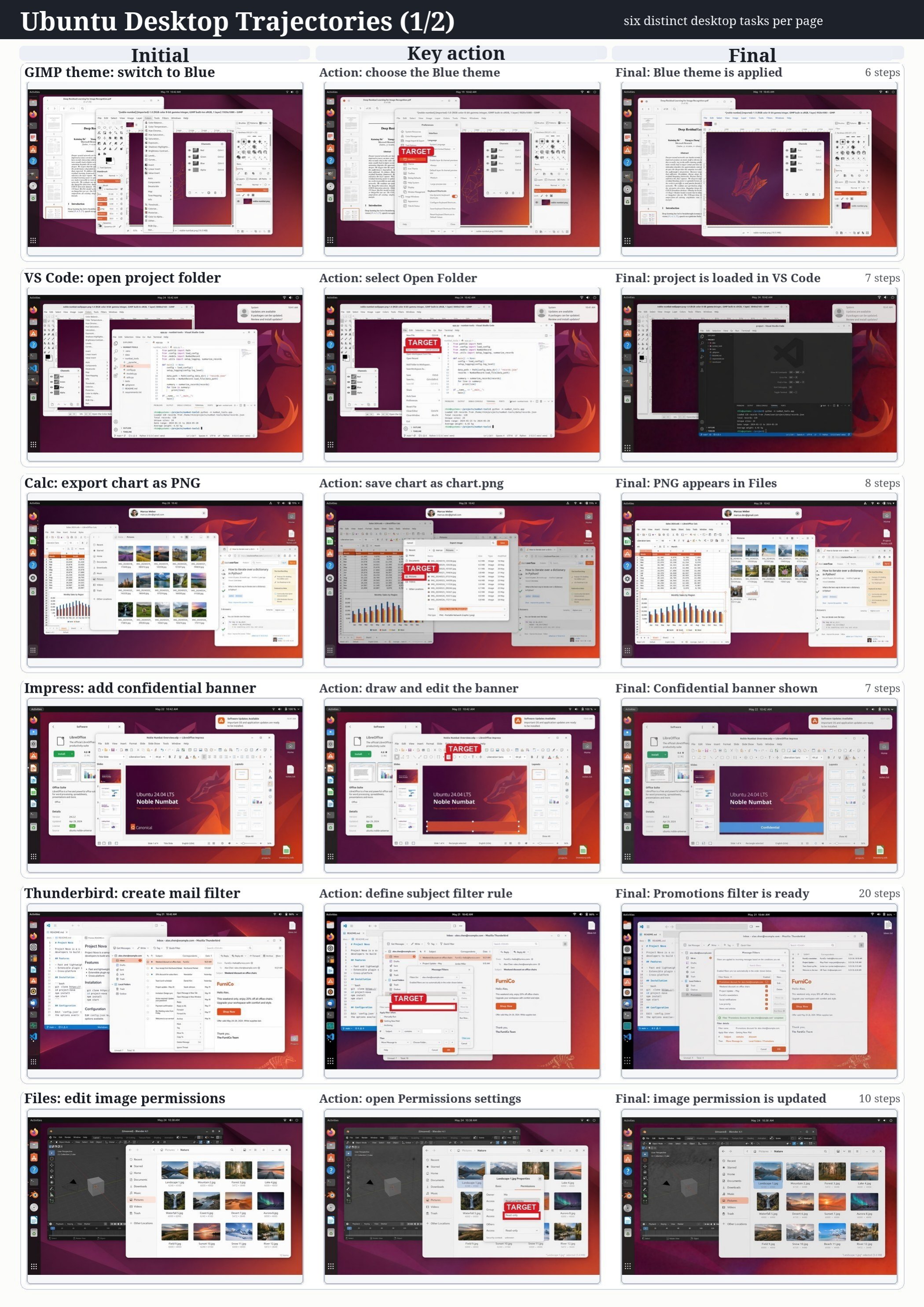}
  \caption{Ubuntu desktop trajectories (1/2).}
  \label{fig:app_autogui_ubuntu_gallery_p1}
\end{figure}

\clearpage
\begin{figure}[H]
  \centering
  \includegraphics[width=\linewidth,height=0.86\textheight,keepaspectratio]{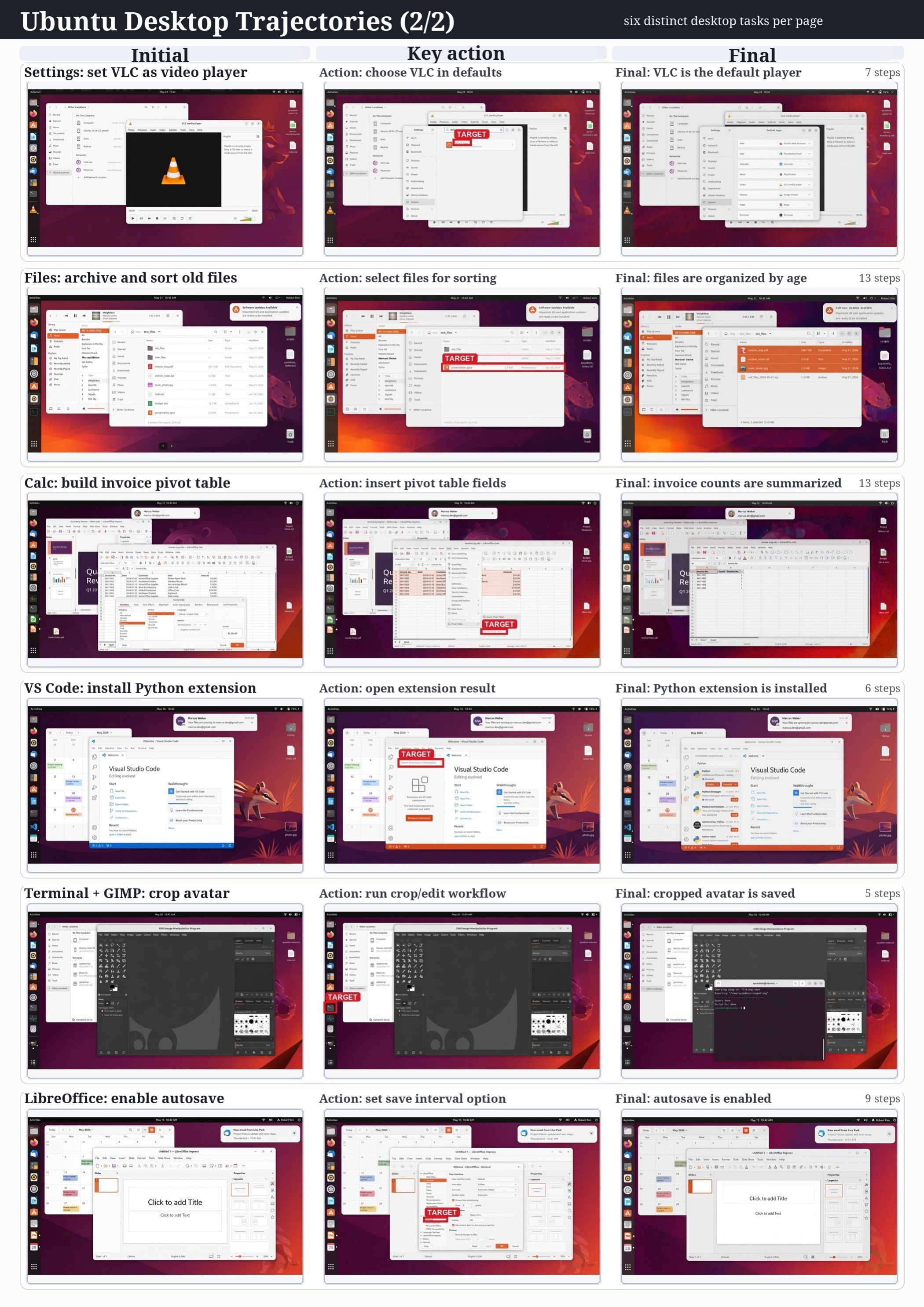}
  \caption{Ubuntu desktop trajectories (2/2).}
  \label{fig:app_autogui_ubuntu_gallery_p2}
\end{figure}

\clearpage
\subsection{Windows 11}

\begin{figure}[H]
  \centering
  \includegraphics[width=\linewidth,height=0.86\textheight,keepaspectratio]{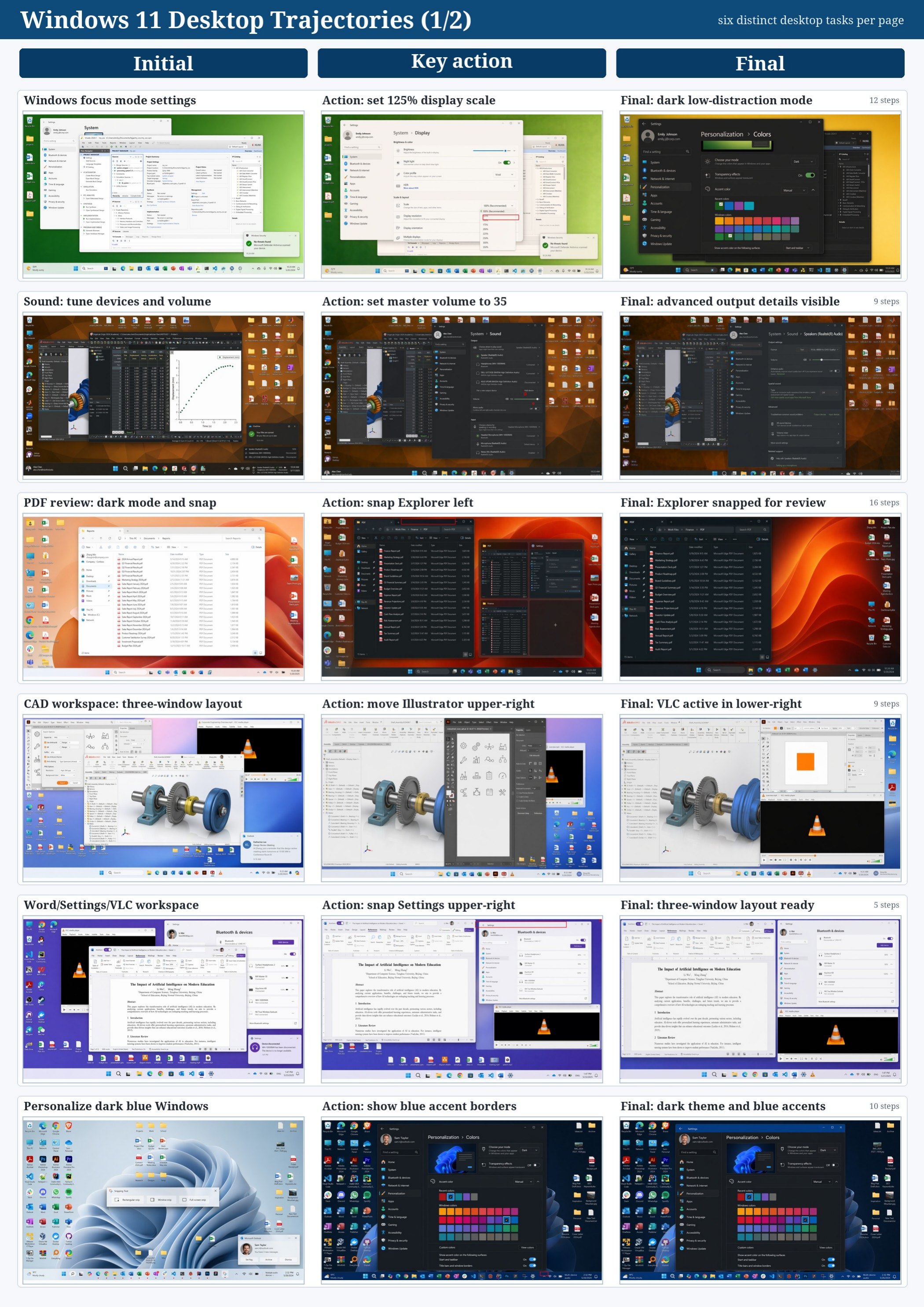}
  \caption{Windows 11 trajectories (1/2).}
  \label{fig:app_autogui_windows11_gallery_p1}
\end{figure}

\clearpage
\begin{figure}[H]
  \centering
  \includegraphics[width=\linewidth,height=0.86\textheight,keepaspectratio]{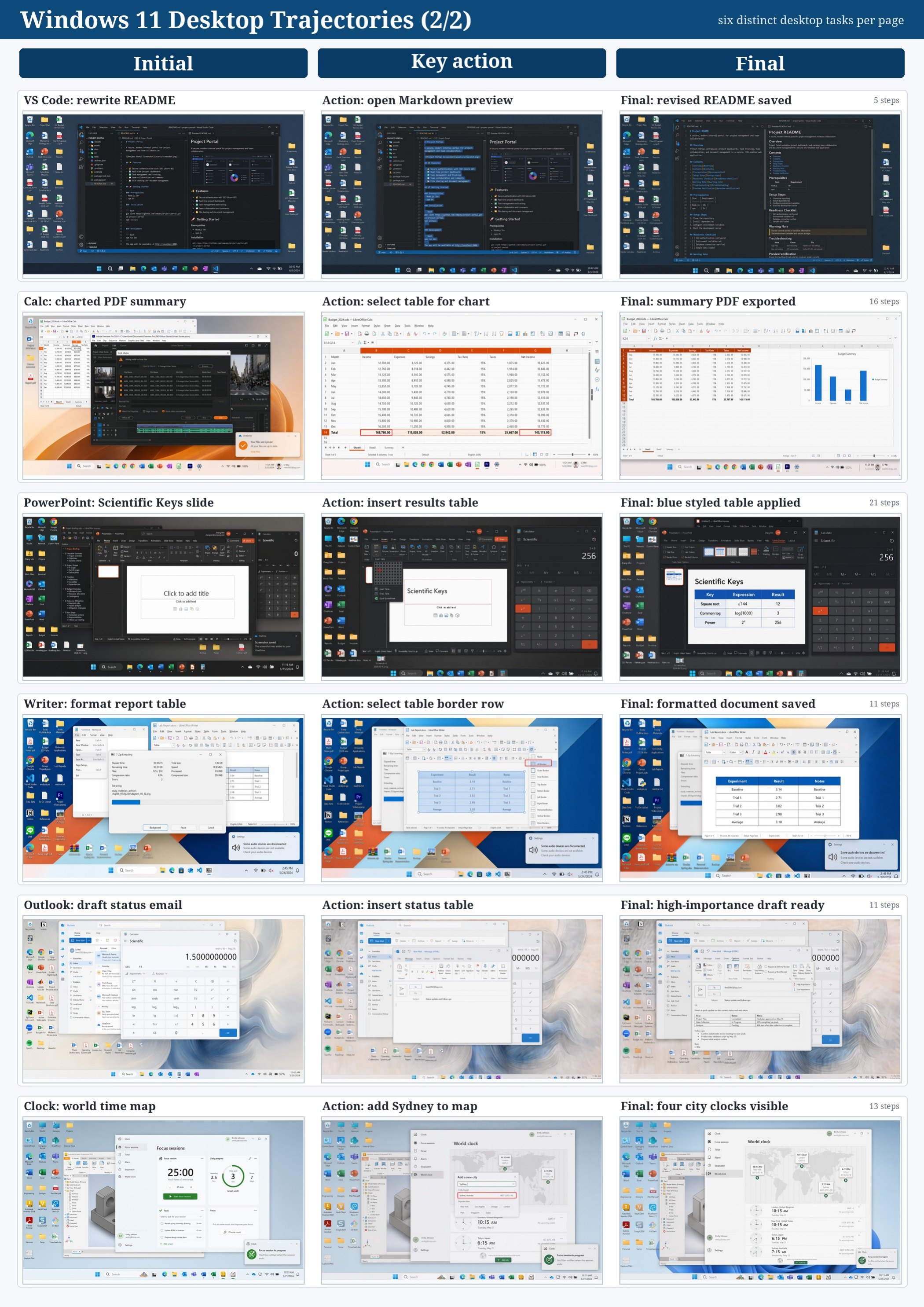}
  \caption{Windows 11 trajectories (2/2).}
  \label{fig:app_autogui_windows11_gallery_p2}
\end{figure}

\clearpage
\subsection{macOS}

\begin{figure}[H]
  \centering
  \includegraphics[width=\linewidth,height=0.86\textheight,keepaspectratio]{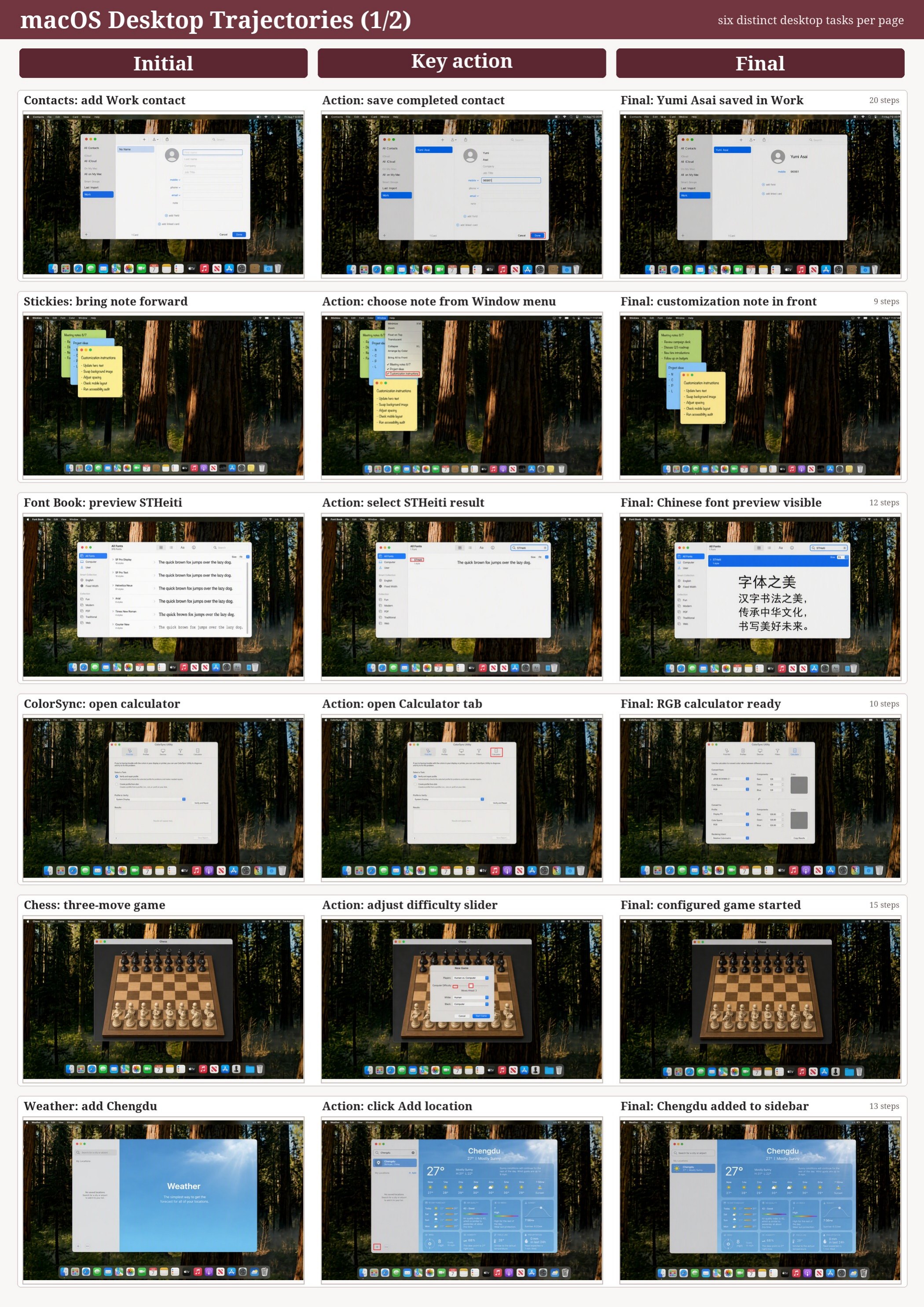}
  \caption{macOS trajectories (1/2).}
  \label{fig:app_autogui_macos_gallery_p1}
\end{figure}

\clearpage
\begin{figure}[H]
  \centering
  \includegraphics[width=\linewidth,height=0.86\textheight,keepaspectratio]{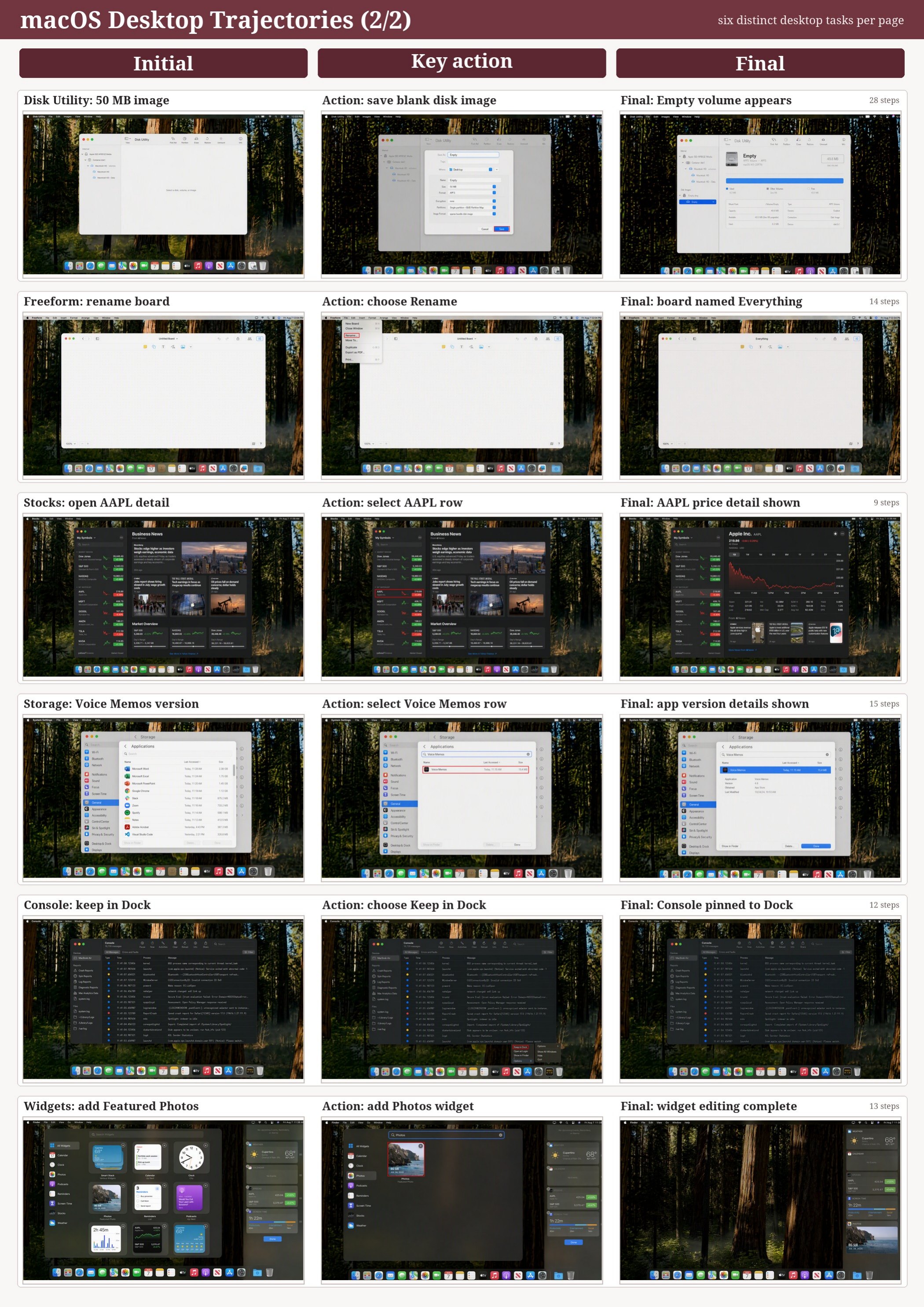}
  \caption{macOS trajectories (2/2).}
  \label{fig:app_autogui_macos_gallery_p2}
\end{figure}

\clearpage
\subsection{Chrome}

\begin{figure}[H]
  \centering
  \includegraphics[width=\linewidth,height=0.86\textheight,keepaspectratio]{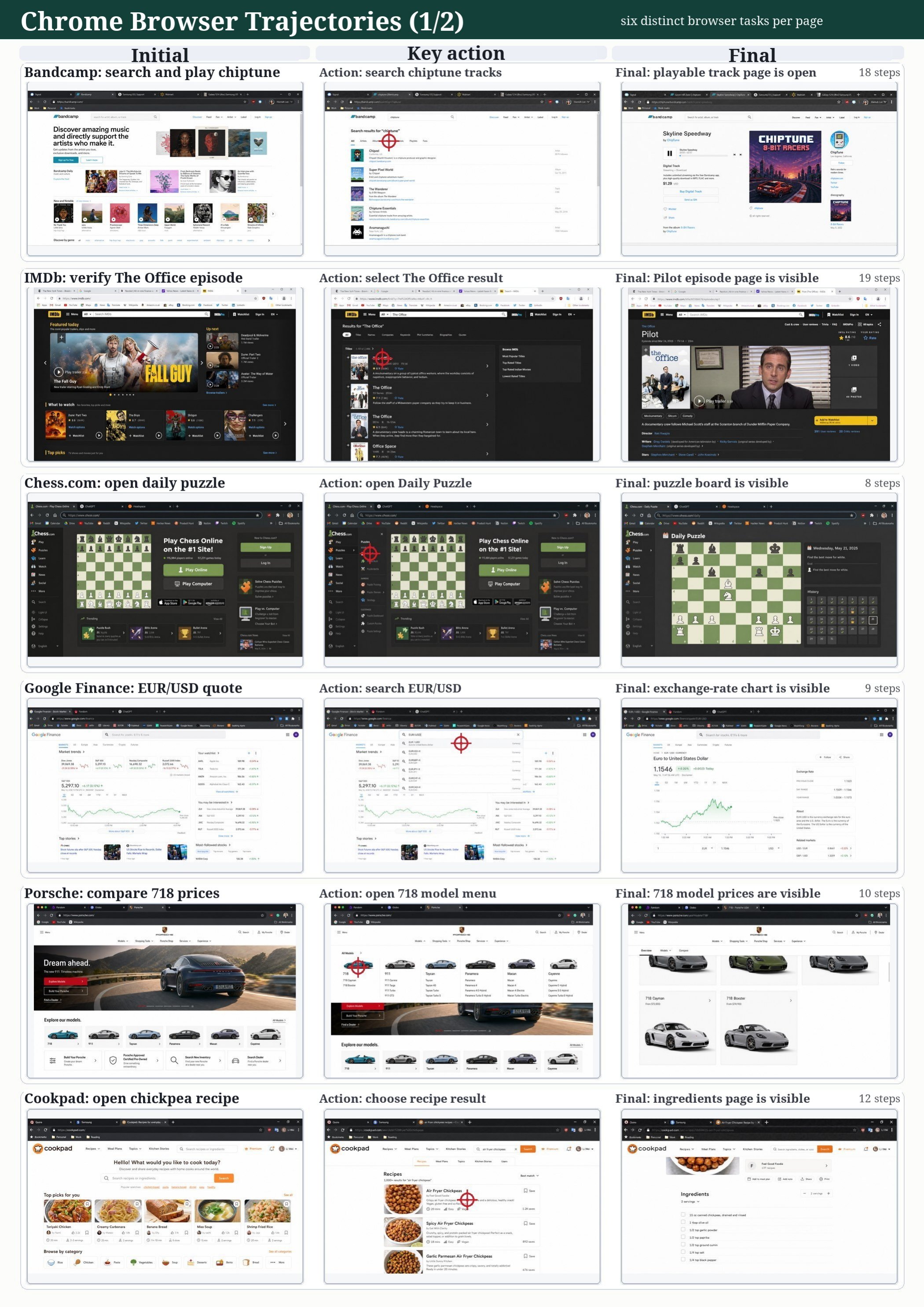}
  \caption{Chrome trajectories (1/2).}
  \label{fig:app_image2gui_web_gallery_p1}
\end{figure}

\clearpage
\begin{figure}[H]
  \centering
  \includegraphics[width=\linewidth,height=0.86\textheight,keepaspectratio]{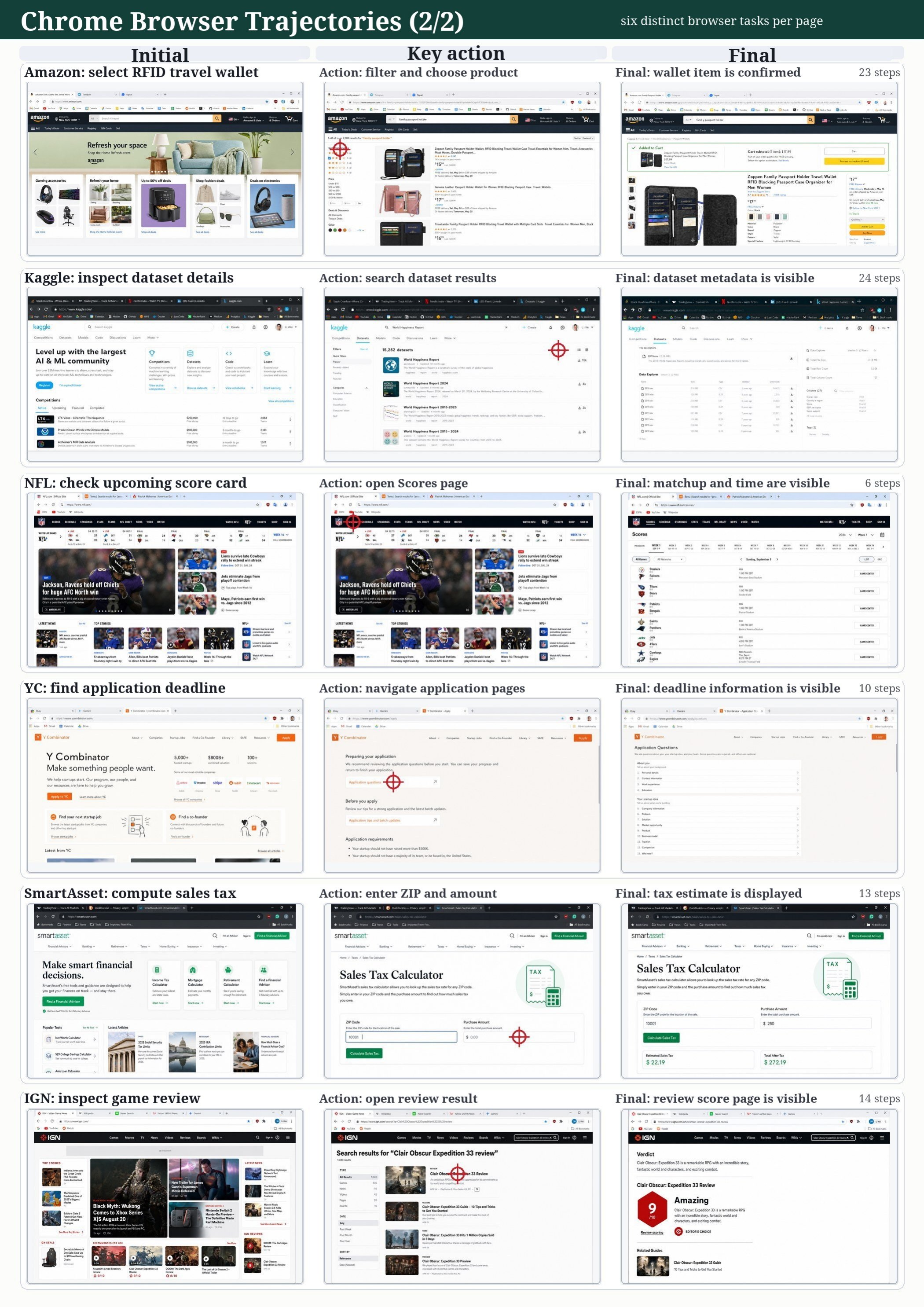}
  \caption{Chrome trajectories (2/2).}
  \label{fig:app_image2gui_web_gallery_p2}
\end{figure}

\clearpage
\section{Professional Application Workflows}
\label{app:professional_application_workflows}

We further examine whether AutoGUIWorld can produce coherent workflows in
dense professional interfaces spanning seven complementary domains. The
coverage includes engineering authoring, visual design, quantitative analysis,
electronic design, audio production, scientific research, and development
automation. The following pages present representative professional outcomes
within each branch of this hierarchy.

\begin{figure}[H]
  \centering
  \includegraphics[width=\linewidth,height=0.78\textheight,keepaspectratio]{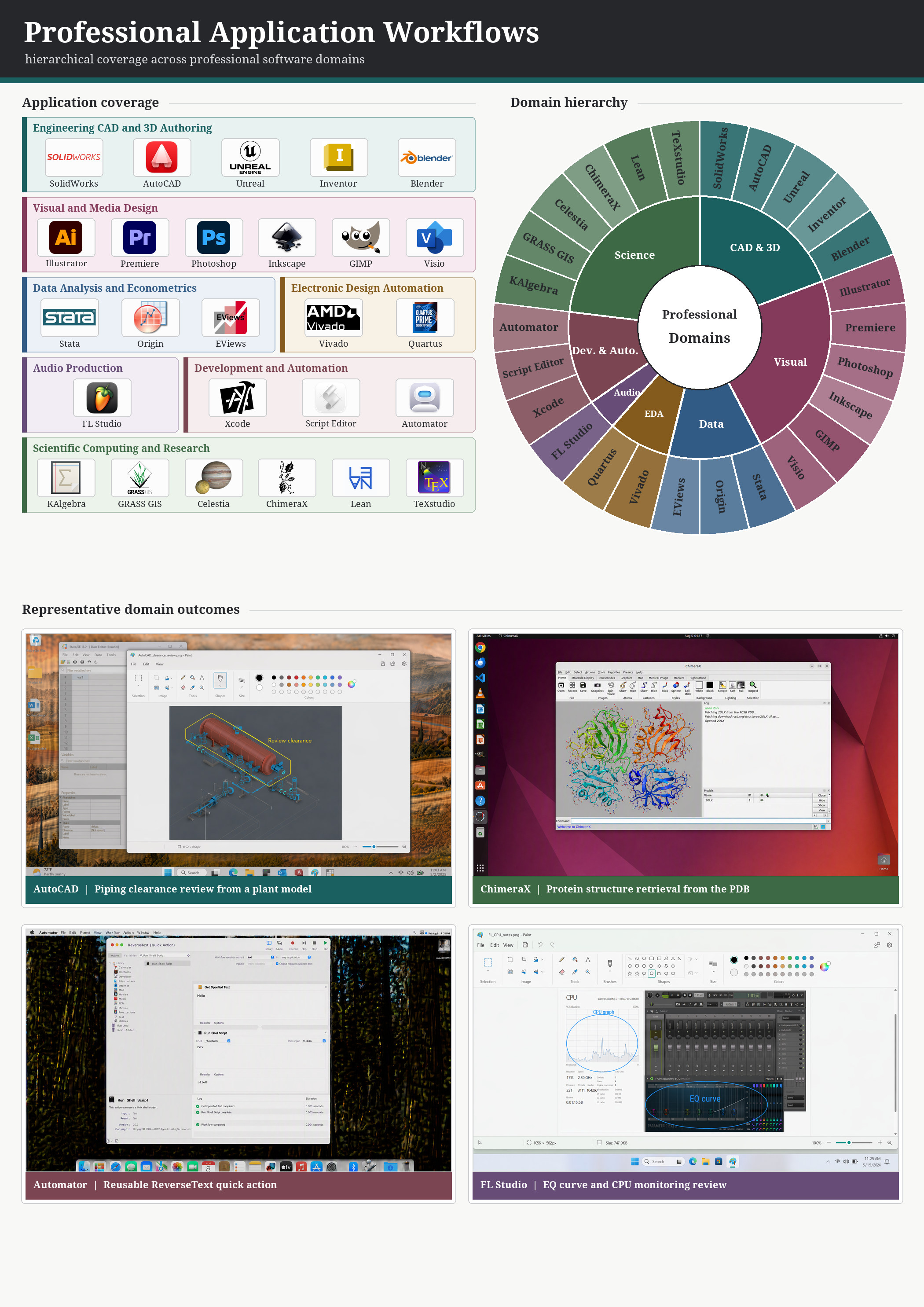}
  \caption{Hierarchical coverage of professional application domains.}
  \label{fig:app_professional_workflows_overview}
\end{figure}

\clearpage
\subsection{Engineering CAD and 3D Authoring}

\begin{figure}[H]
  \centering
  \includegraphics[width=\linewidth,height=0.86\textheight,keepaspectratio]{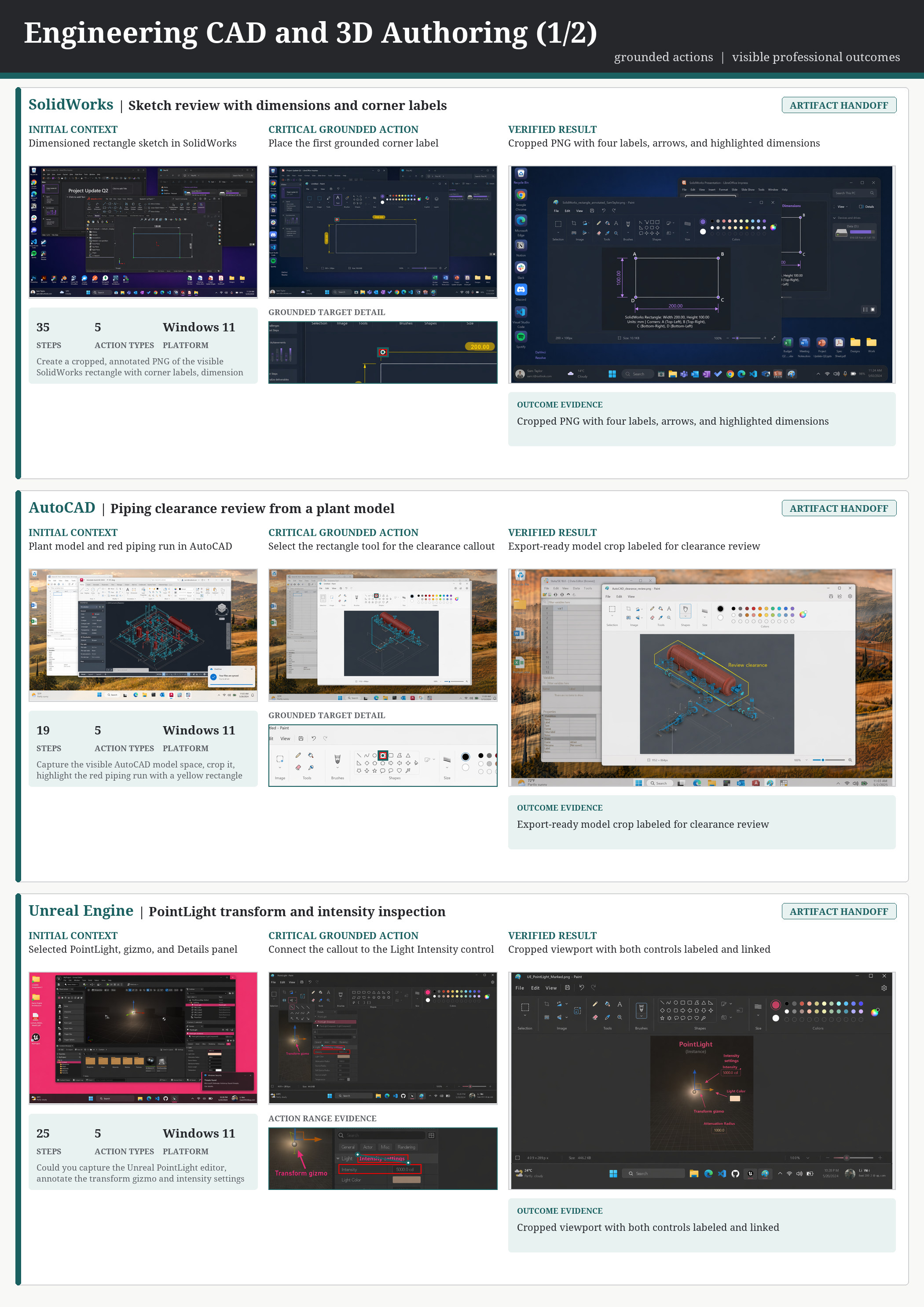}
  \caption{Engineering CAD and 3D authoring workflows (1/2).}
  \label{fig:app_professional_cad_gallery_p1}
\end{figure}

\clearpage
\begin{figure}[H]
  \centering
  \includegraphics[width=\linewidth,height=0.86\textheight,keepaspectratio]{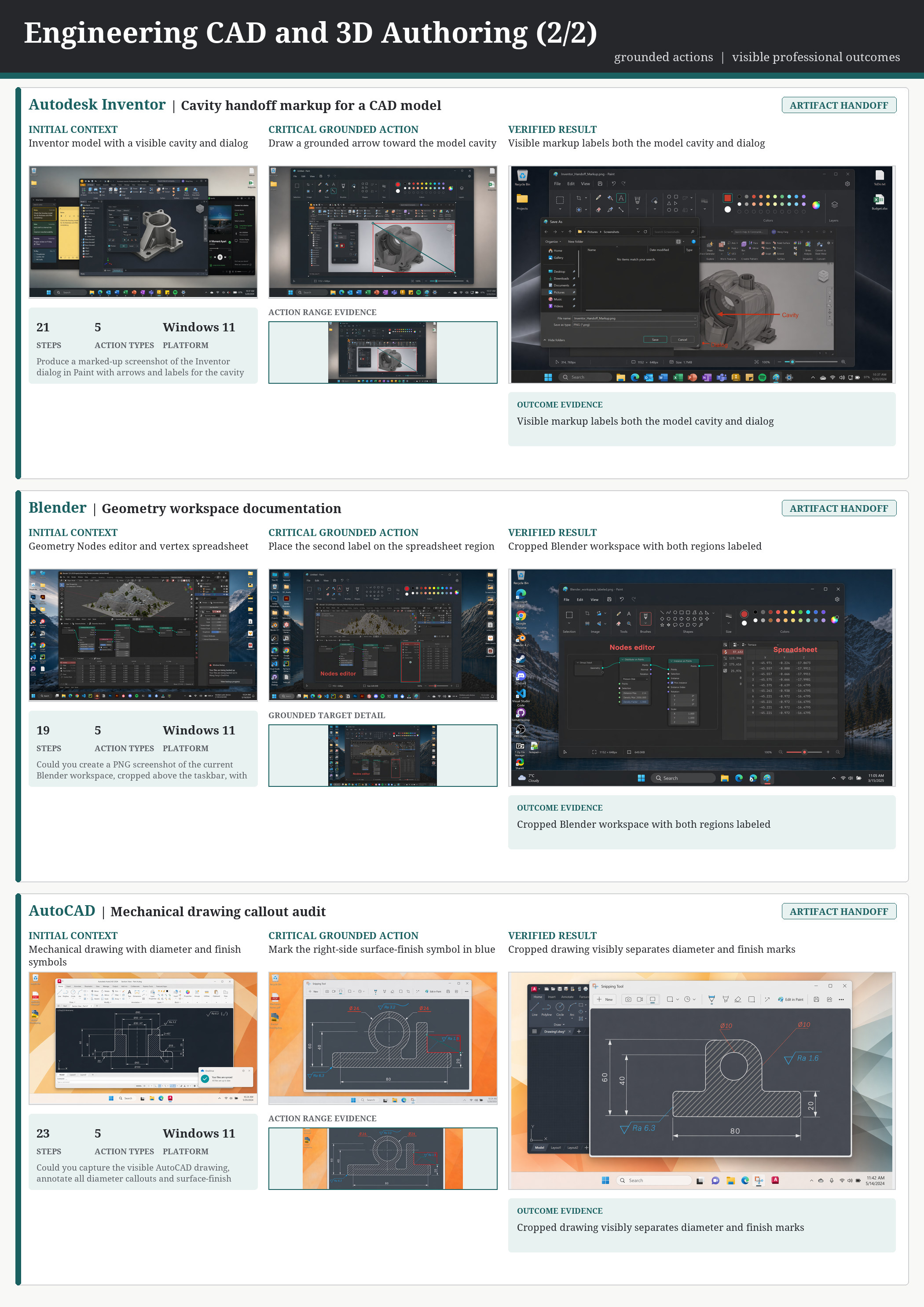}
  \caption{Engineering CAD and 3D authoring workflows (2/2).}
  \label{fig:app_professional_cad_gallery_p2}
\end{figure}

\clearpage
\subsection{Visual and Media Design}

\begin{figure}[H]
  \centering
  \includegraphics[width=\linewidth,height=0.86\textheight,keepaspectratio]{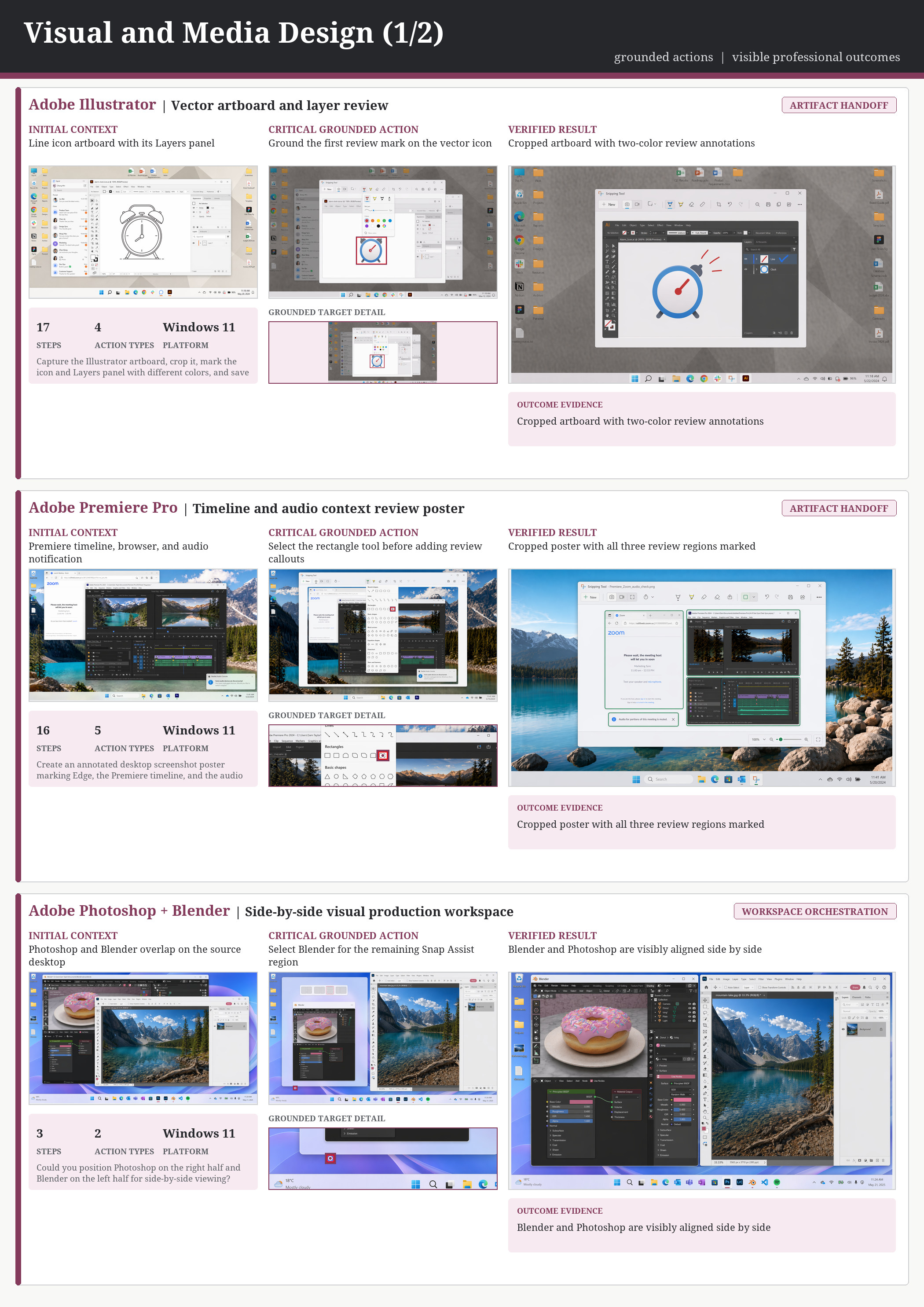}
  \caption{Visual and media design workflows (1/2).}
  \label{fig:app_professional_visual_gallery_p1}
\end{figure}

\clearpage
\begin{figure}[H]
  \centering
  \includegraphics[width=\linewidth,height=0.86\textheight,keepaspectratio]{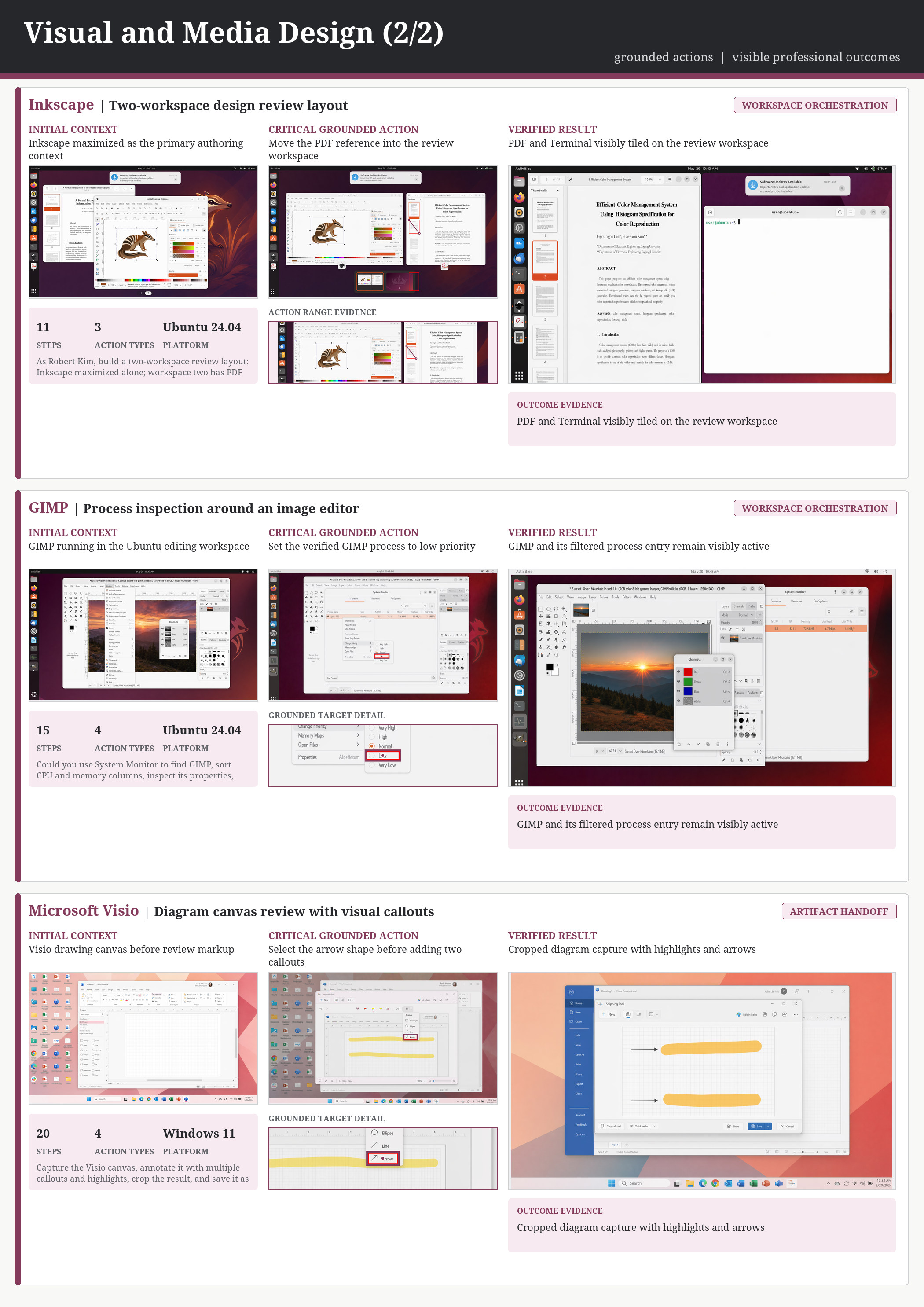}
  \caption{Visual and media design workflows (2/2).}
  \label{fig:app_professional_visual_gallery_p2}
\end{figure}

\clearpage
\subsection{Data Analysis and Econometrics}

\begin{figure}[H]
  \centering
  \includegraphics[width=\linewidth,height=0.86\textheight,keepaspectratio]{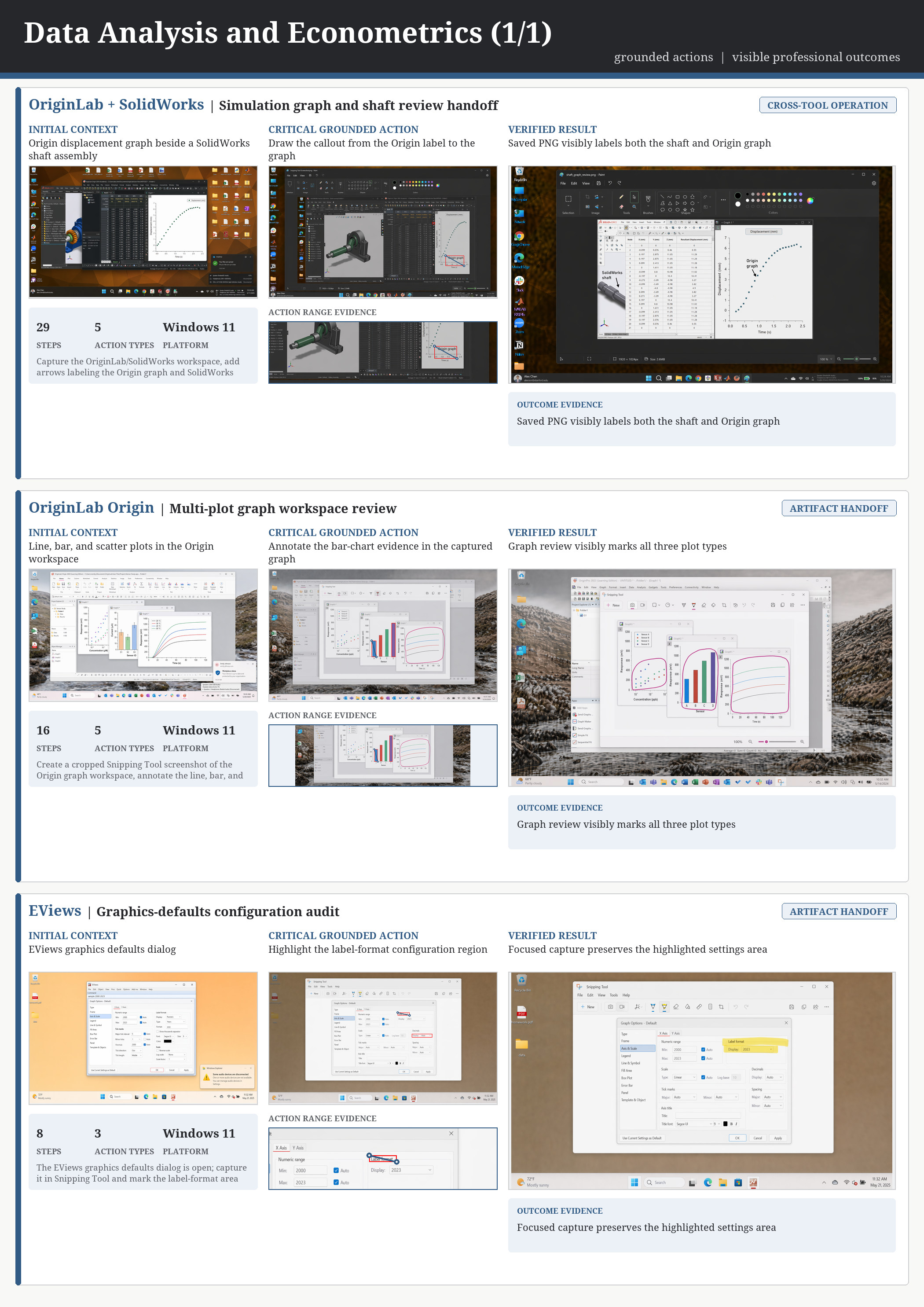}
  \caption{Data analysis and econometrics workflows.}
  \label{fig:app_professional_data_gallery_p1}
\end{figure}

\clearpage
\subsection{Electronic Design Automation}

\begin{figure}[H]
  \centering
  \includegraphics[width=\linewidth,height=0.86\textheight,keepaspectratio]{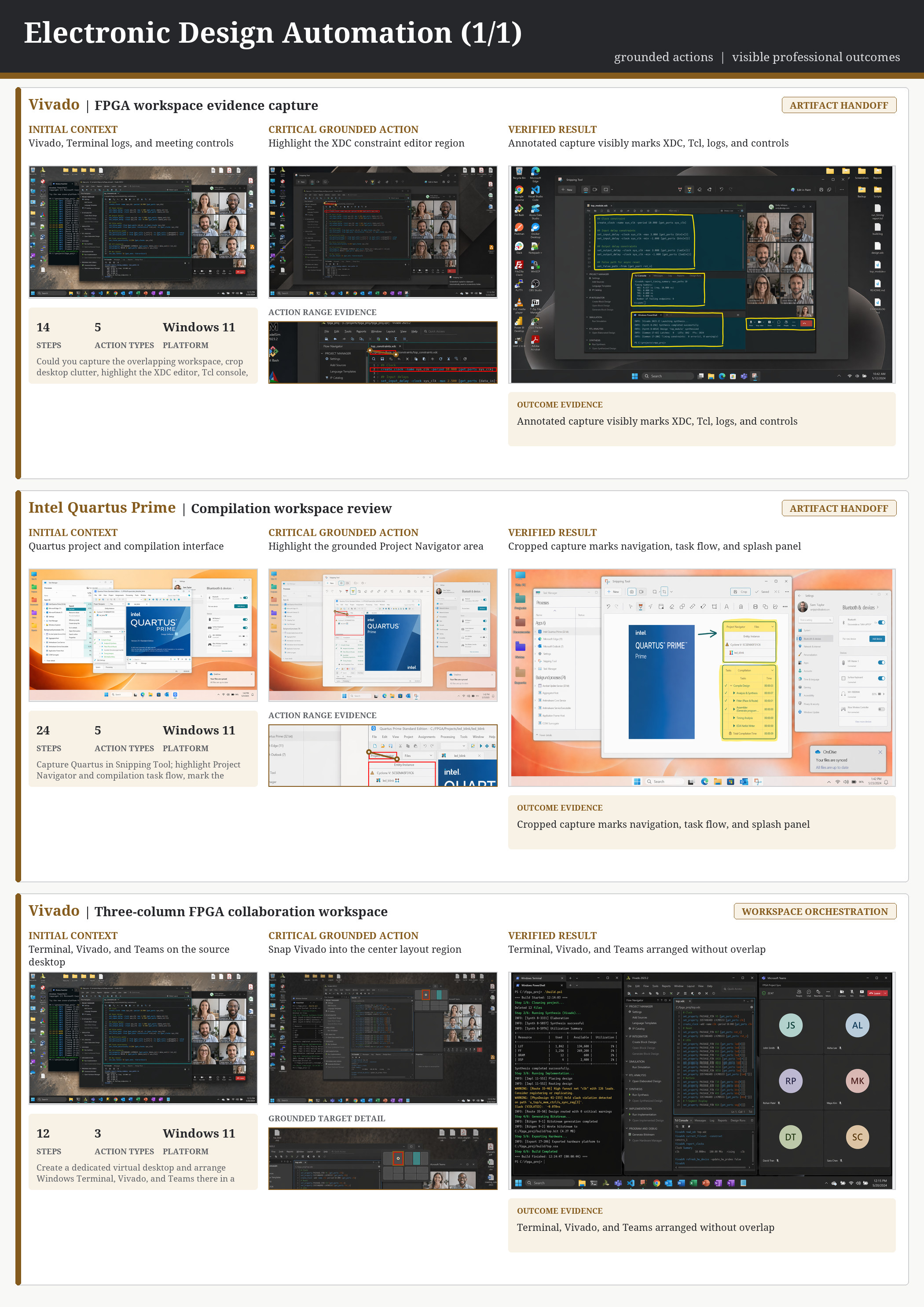}
  \caption{Electronic design automation workflows.}
  \label{fig:app_professional_eda_gallery_p1}
\end{figure}

\clearpage
\subsection{Audio Production}

\begin{figure}[H]
  \centering
  \includegraphics[width=\linewidth,height=0.86\textheight,keepaspectratio]{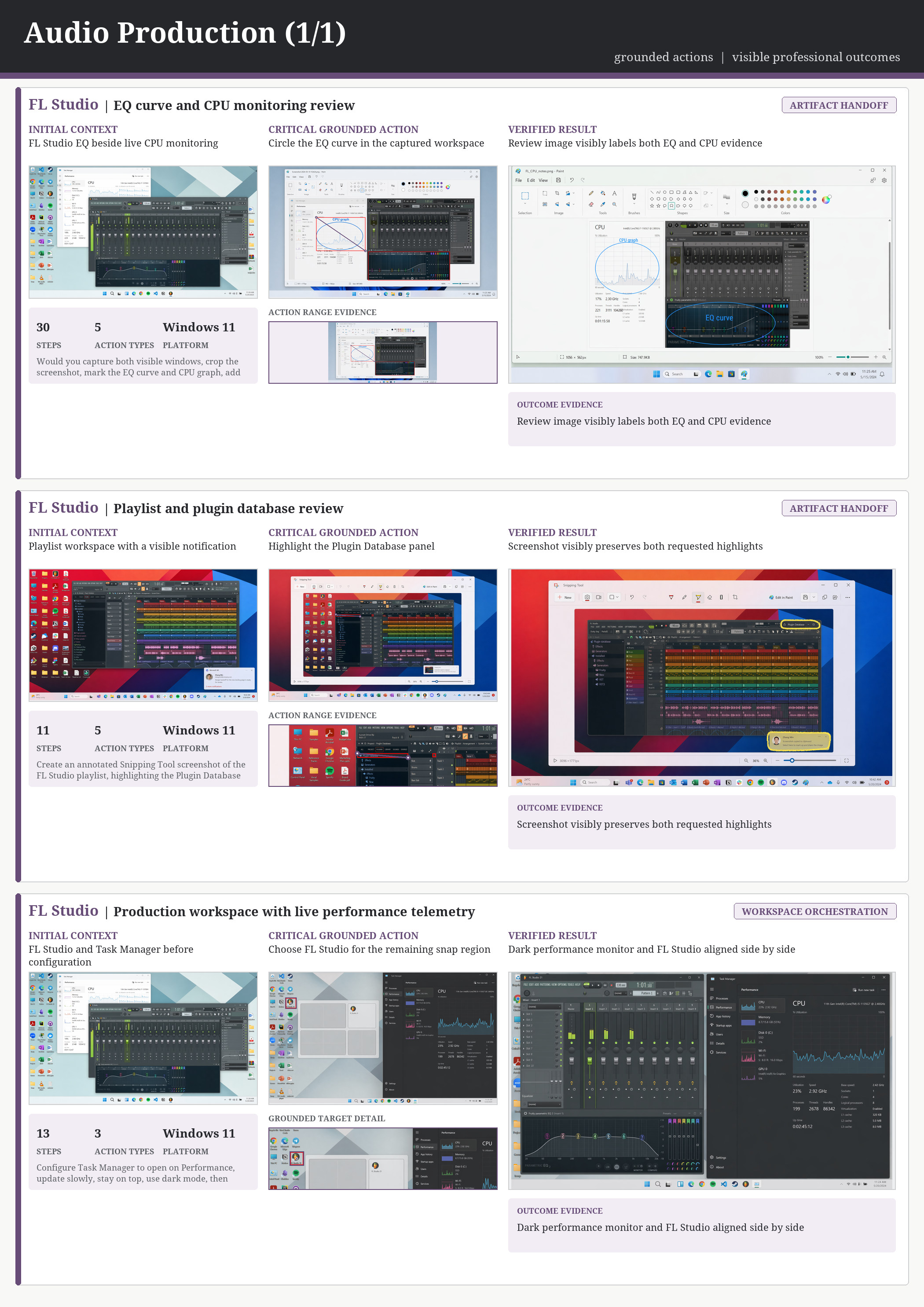}
  \caption{Audio production workflows.}
  \label{fig:app_professional_audio_gallery_p1}
\end{figure}

\clearpage
\subsection{Scientific Computing and Research}

\begin{figure}[H]
  \centering
  \includegraphics[width=\linewidth,height=0.86\textheight,keepaspectratio]{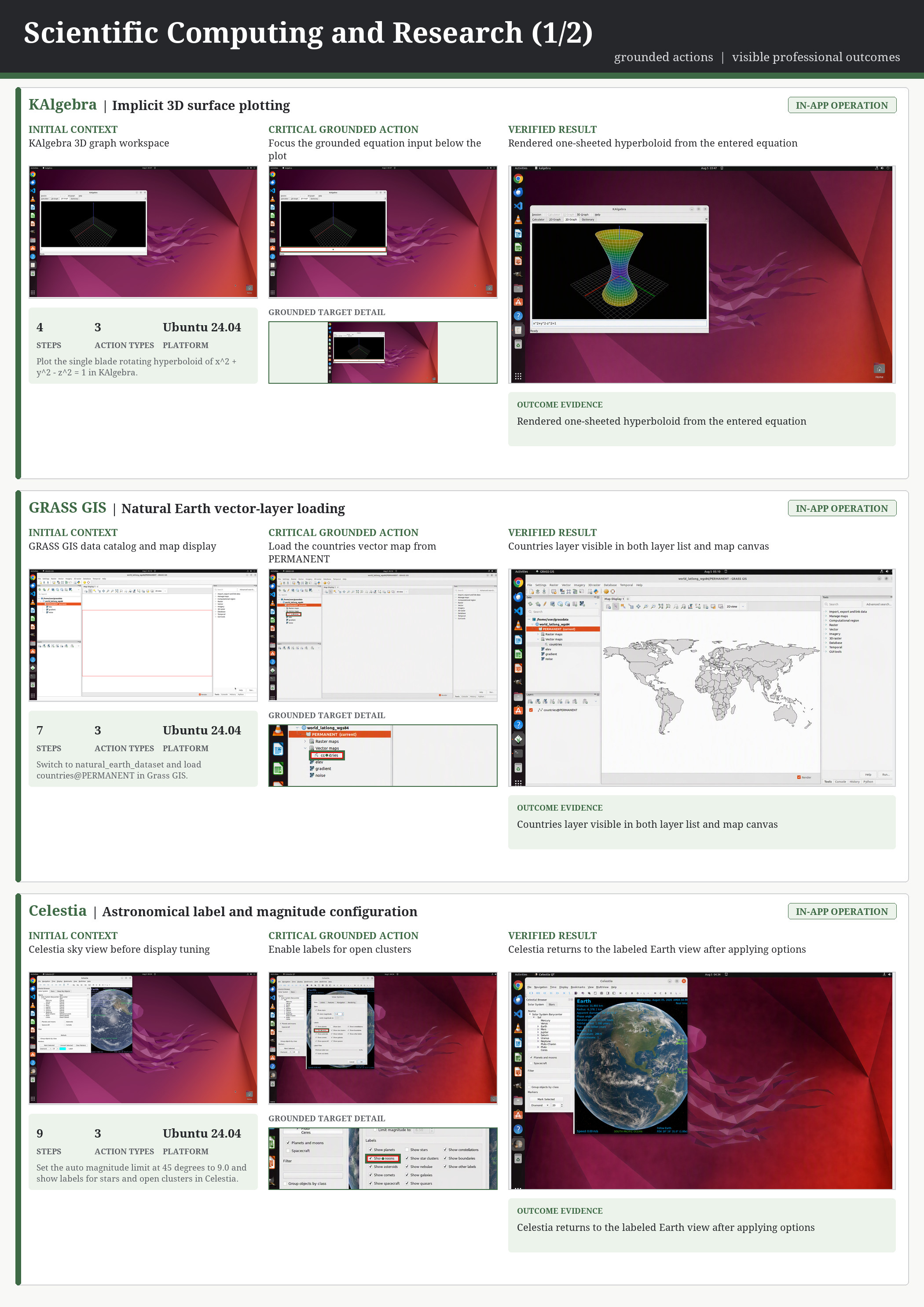}
  \caption{Scientific computing and research workflows (1/2).}
  \label{fig:app_professional_science_gallery_p1}
\end{figure}

\clearpage
\begin{figure}[H]
  \centering
  \includegraphics[width=\linewidth,height=0.86\textheight,keepaspectratio]{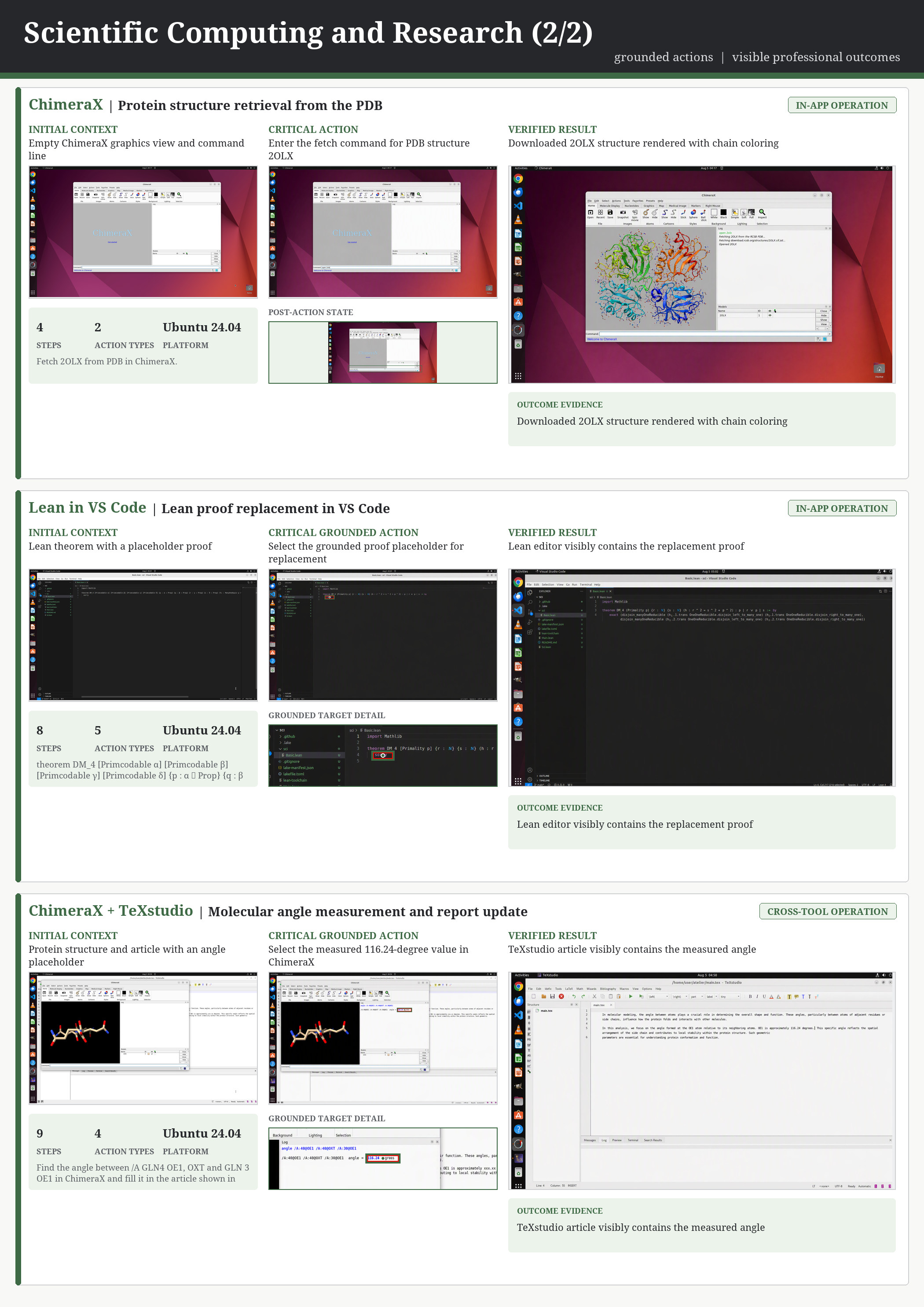}
  \caption{Scientific computing and research workflows (2/2).}
  \label{fig:app_professional_science_gallery_p2}
\end{figure}

\clearpage
\subsection{Development and Automation}

\begin{figure}[H]
  \centering
  \includegraphics[width=\linewidth,height=0.86\textheight,keepaspectratio]{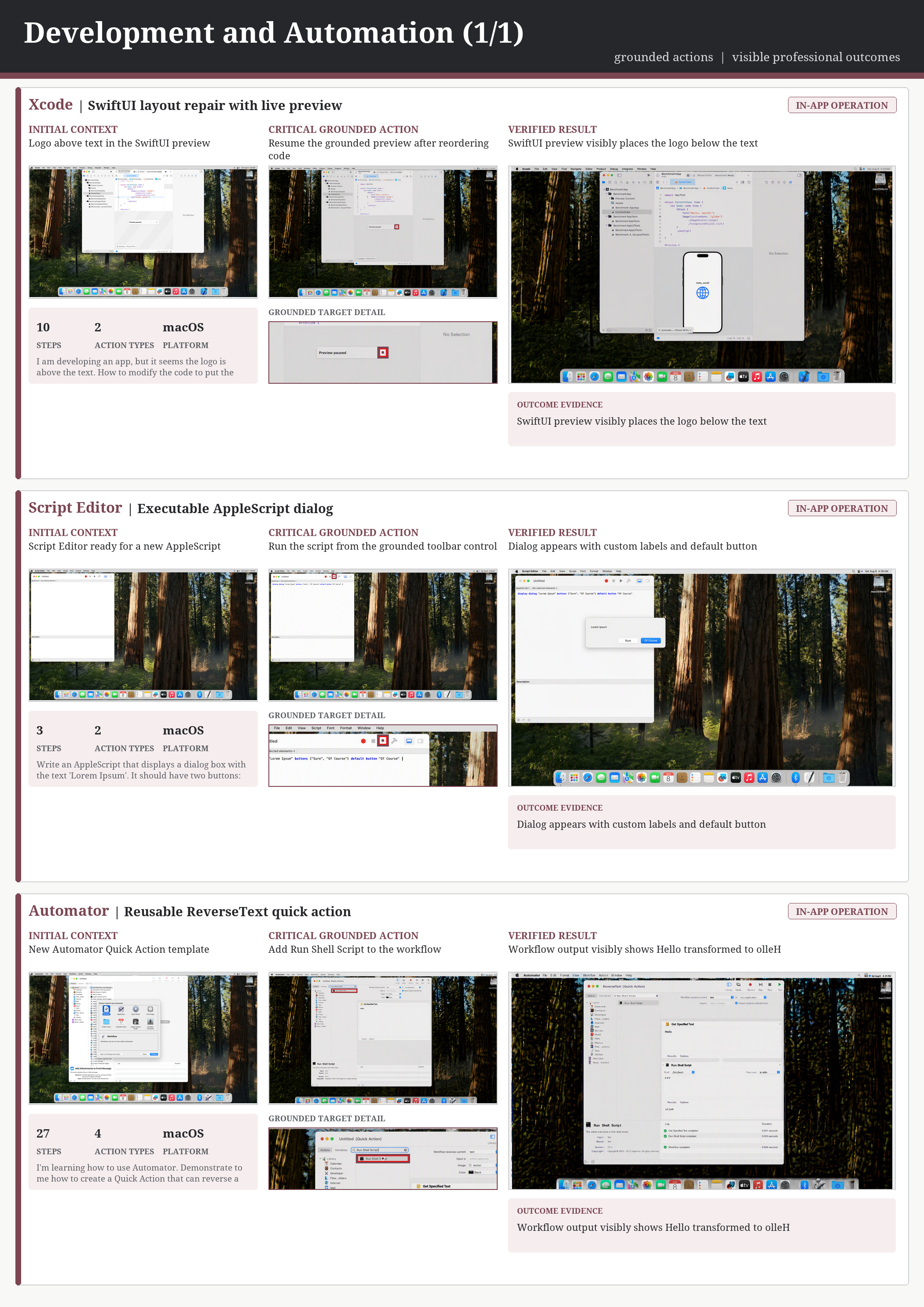}
  \caption{Development and automation workflows.}
  \label{fig:app_professional_dev_gallery_p1}
\end{figure}

\clearpage
\clearpage
\section{Prompt Catalog}
\label{app:prompt_catalog}

AutoGUIWorld constructs each trajectory through seed realization, task
generation, action planning, visual transition synthesis, and target annotation.
The templates below specify the inputs and outputs of these stages in the
public implementation. Braced names denote fields populated at runtime.
The accompanying source manifest records the code revision and source symbol
for every prompt.

\begin{table}[H]
  \centering
  \small
  \setlength{\tabcolsep}{4pt}
  \begin{tabularx}{\linewidth}{@{}>{\raggedright\arraybackslash}p{0.22\linewidth}>{\raggedright\arraybackslash}X>{\raggedright\arraybackslash}X@{}}
    \toprule
    Stage & Input & Output \\
    \midrule
    Seed realization & Platform, appearance, sampled environment & Initial-screen description and visible elements \\
    Task generation & Seed context, task history, sampling directives & Task instruction and task attributes \\
    Meta Planner & Task, fixed seed, action vocabulary & High-level plan and atomic action sequence \\
    Voyager and Image2 & Current screenshot, action, rollout progress & Step text, rendering instruction, next screenshot \\
    Target grounding & Clean screenshot and element description & Target region or point \\
    Quality checks & Seed, task, or transition evidence & Structured defect and consistency judgments \\
    \bottomrule
  \end{tabularx}
  \caption{Prompt inputs and outputs in the AutoGUIWorld generation pipeline.}
  \label{tab:prompt_catalog}
\end{table}

\begingroup
\newcommand{\catalogblock}[2]{%
  \par\Needspace{5\baselineskip}%
  \noindent\textit{#1}\par\nopagebreak
  \lstinputlisting[style=catalogprompt]{assets/prompts/catalog/#2}%
}
\newcommand{\catalogshortblock}[2]{%
  \par\noindent\begin{minipage}{\linewidth}%
  \textit{#1}\par
  \lstinputlisting[style=catalogprompt]{assets/prompts/catalog/#2}%
  \end{minipage}\par
}

\subsection{Environment Description and Seed Realization}
\label{app:prompt_seed}

The seed describer converts the sampled platform, appearance, and environment
into a structured initial-screen description. Its output records the visible
elements, foreground surface, and rendering prompt. Image2 receives
\texttt{global\_state.prompt} to generate the initial observation.

\catalogblock{System template.}{seed_system.txt}
\catalogblock{User template.}{seed_user.txt}

The builder fills \texttt{os\_json} from the platform registry and selects the
layout instructions by interface category. Windows, macOS, and Ubuntu share
the desktop block. Chrome uses the browser block, while Android and iOS use
the mobile block. The following pairs fill \texttt{layout\_directive} and
\texttt{describe\_items}, respectively.

\catalogblock{Desktop layout instructions.}{seed_desktop_layout.txt}
\catalogblock{Desktop description checklist.}{seed_desktop_items.txt}
\catalogblock{Chrome layout instructions.}{seed_browser_layout.txt}
\catalogblock{Chrome description checklist.}{seed_browser_items.txt}
\catalogblock{Mobile layout instructions.}{seed_mobile_layout.txt}
\catalogblock{Mobile description checklist.}{seed_mobile_items.txt}

\subsection{Seed-Conditioned Task Generation}
\label{app:prompt_task}

The task generator receives the platform's application and scenario lists,
the fixed seed context, and recent task instructions. It returns one task
with application, category, complexity, and feasibility fields. The prompt
ties the requested goal to the seed's visible interface and uses task history
to encourage semantic diversity.

\catalogblock{System message.}{task_system.txt}
\catalogblock{User template.}{task_user.txt}

The \texttt{seed\_section} lists open windows and their positions, browser
tabs, or home-screen applications, followed by visible elements and the
foreground target. The \texttt{directive\_section} specifies feature coverage,
target difficulty, instruction style, recent verbs to avoid, and task focus.
Each request includes the applicable fields. Task history contains up to
15 recent instructions; a rejected duplicate is added before another request.

\catalogblock{Task-generation directives.}{task_directives.txt}

Feasibility directives select a completable task, an immediately identifiable
infeasible request, or a request whose infeasibility requires interface
inspection. The task metadata carries this choice to the planner.

\catalogblock{Feasible-task directive.}{task_feasible.txt}
\catalogblock{Infeasible-task directive: upfront judgment.}{task_infeasible_upfront.txt}
\catalogblock{Infeasible-task directive: interface exploration.}{task_infeasible_explore.txt}

\subsection{Meta Planner}
\label{app:prompt_planner}

The Meta Planner receives the task as its user message. Its system prompt
combines the fixed seed, unresolved preconditions, and a platform-specific
action vocabulary. The output contains a high-level plan and an ordered list
of atomic actions. Each action records its intent, target, and expected
before and after states.

\catalogblock{System template.}{planner_system.txt}

The builder inserts the following action records into
\texttt{action\_space\_str}. These records express semantic actions and
element names during generation. Appendix~\ref{app:action_space} specifies
the coordinate-based \texttt{computer\_use} interface for agent training
and evaluation.

\catalogblock{Desktop generation actions.}{planner_actions_pc.txt}
\catalogblock{Chrome generation actions.}{planner_actions_web.txt}
\catalogblock{Mobile generation actions.}{planner_actions_mobile.txt}

The following blocks fill \texttt{precondition\_rules} and
\texttt{grounding\_rules}. They define how each interface category handles
blocked states and selects visible action targets. The separate
\texttt{blocker\_section} enumerates the blockers recorded in the seed.
With no blockers, actions enter the main phase directly.

\catalogblock{Desktop preconditions and target rules.}{planner_desktop_rules.txt}
\catalogblock{Chrome preconditions and target rules.}{planner_browser_rules.txt}
\catalogblock{Mobile preconditions and target rules.}{planner_mobile_rules.txt}

For an infeasible task, the builder adds a directive specifying when and why
the task cannot be completed. The exploration hint identifies the interface
to inspect. An upfront judgment produces a terminal failure
response. An exploration judgment plans the necessary inspection before
reporting failure. The \texttt{must\_not\_render} field carries absent-option
constraints into visual synthesis.

\catalogblock{Planner directive: upfront judgment.}{planner_infeasible_upfront.txt}
\catalogblock{Planner directive: interface exploration.}{planner_infeasible_explore.txt}

\subsection{Voyager and Next-State Rendering}
\label{app:prompt_voyager}

Voyager observes the clean pre-action screenshot and receives the complete
action record. It produces the agent's step reasoning, an imperative action
summary, and a description of the next screen. The prompt separates the
agent's before-action perspective from the renderer's after-action view.
The high-level plan and completed action summaries provide rollout context.

\catalogblock{Voyager system message.}{voyager_system.txt}
\catalogshortblock{Voyager user template.}{voyager_user.txt}

The \texttt{action\_json} field contains the full action object, including
text, keys, direction, or element arguments. The progress fields identify
the current step and preceding action summaries. Target hints and
\texttt{must\_not\_render} appear when provided by the planner.
The current clean observation accompanies the text as the image input.

Image2 receives the same observation as its image-edit reference. The
following instruction wraps Voyager's \texttt{voyager\_prompt} in
\texttt{desc}, yielding the next clean observation for the following step.

\catalogblock{Image2 transition instruction.}{render_edit.txt}

\subsection{Action Target Grounding}
\label{app:prompt_grounding}

LocateAnything receives the clean pre-action screenshot and a semantic target
description. The region query requests the target's bounding box, while the
point query requests its location. The generation pipeline uses the region
query and converts the returned 0--1000 coordinates to image pixels for
action annotation.

\catalogblock{LocateAnything region query.}{ground_locateanything_box.txt}
\catalogblock{LocateAnything point query.}{ground_locateanything_point.txt}

\subsection{Quality Checks}
\label{app:prompt_quality}

Quality checks evaluate initial-screen fidelity, task--seed compatibility,
and individual trajectory transitions. Each checker returns structured
fields that can be aggregated or used for filtering.

\subsubsection{Seed Image Quality}

The seed checker receives the initial screenshot and platform name. It
reports concrete defects from nine categories, with a severity and screen
region for each. Its system message is
\texttt{Return the requested assessment as JSON.}

\catalogblock{Seed-quality user template.}{qc_seed_user.txt}

\subsubsection{Task--Seed Compatibility}

The compatibility checker receives the task, expected foreground application,
and initial screenshot. It checks both the application and the task-specific
content, returning \texttt{match}, \texttt{partial}, or \texttt{mismatch}.
It uses the same JSON-only system message as the seed checker.

\catalogblock{Task--seed compatibility user template.}{qc_relevance_user.txt}

\subsubsection{Trajectory Step Quality}

The trajectory checker receives the task and action context, followed by
the clean \textsc{Before} image, the annotated \textsc{Target} image for
pointing actions, and the generated \textsc{After} image. Its judgments
cover target existence, box placement, action validity, visual transition,
and step reasoning.

\catalogblock{Trajectory-QC system message.}{qc_trajectory_system.txt}
\catalogblock{Trajectory-QC user template in the public implementation.}{qc_trajectory_user.txt}

Appendix~\ref{app:transition_fidelity} gives the trajectory-audit prompt,
dimension definitions, and filtering procedure for the reported transition
analysis. In the public implementation, a failed
\texttt{obs\_transition\_ok} check receives the aggregate label \texttt{major}.
The audit in
Appendix~\ref{app:transition_fidelity} filters that field directly after
checking the action and target.

\endgroup

\end{document}